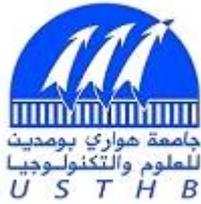 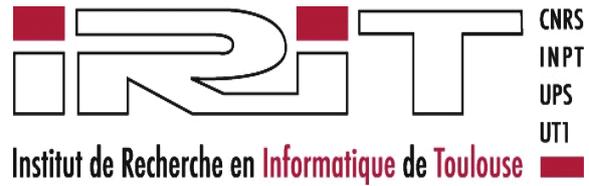

Université des Sciences et Technologie Houari Boumediene

# THESE

Spécialité : Informatique

Option : Recherche

Présentée pour obtenir le titre :

## D'ingénieur en informatique

Par

## Djallel Bouneffouf

Rapport du stage effectué au laboratoire de recherche en

Informatique de Toulouse (IRIT)

## Rôle de l'inférence temporelle dans la reconnaissance de l'inférence textuelle

Soutenu le 18 juin 2008 devant le jury composé de :

Madame A.Aissani Présidente

Monsieur H.Azzoune Examinateur

Madame F.Khellaf Directrice de thèse

# Résumé du projet

Ce projet s'insère dans le cadre du traitement du langage nature. Il a pour objectif le développement d'un système de reconnaissance d'inférence textuelle, nommé TIMINF. Ce type de système permet de détecter, étant donné deux portions de textes, si un des textes est sémantiquement déduit de l'autre.
Nous nous sommes focalisés sur l'apport de l'inférence temporelle dans ce type de système. Pour cela, nous avons constitué et analysé un corpus construit à partir de questions collectées à travers le web.
Cette étude, nous a permis de classer différents types d'inférences temporelles et de concevoir l'architecture informatique de TIMINF qui a pour but l'intégration d'un module d'inférence temporelle dans un système de détection d'inférence textuelle.
Nous proposons, également d'évaluer les performances des sorties du système TIMINF sur un corpus de test avec la même stratégie adopté dans le challenge RTE.

**Mot clef :** Traitement du langage naturel, reconnaissance d'inférence textuelle, inférence temporelle, système question réponse, Recherche d'information.

# Project summary

This project is a part of nature language processing and its aims to develop a system of recognition inference text-appointed TIMINF. This type of system can detect, given two portions of text, if a text is semantically deducted from the other. We focused on making the inference time in this type of system. For that we have built and analyzed a body built from questions collected through the web. This study has enabled us to classify different types of times inferences and for designing the architecture of TIMINF which seeks to integrate a module inference time in a detection system inference text.
We also assess the performance of sorties TIMINF system on a test corpus with the same strategy adopted in the challenge RTE.

# Table des matières

# Table des illustrations

## Les figures

## Les tableaux

# Introduction générale

Nous regroupons sous le vocable de traitement automatique du langage naturel (TALN) l'ensemble des recherches et développements visant à modéliser et à reproduire, à l'aide de machines, la capacité humaine à produire et à comprendre des énoncés linguistiques dans le but de communication (Yvon, 2007).

Les deux sources principales de motivation à l'étude du TALN sont d'une part; la volonté de modéliser une compétence fascinante (le langage), afin de tester des hypothèses sur les mécanismes de la communication humaine, ou plus généralement sur la nature de la cognition humaine et d'autre part le besoin de disposer d'applications capables de traiter efficacement les morceaux d'informations « naturelles» (documents écrits ou sonores) aujourd'hui disponibles sous forme électronique (mails, pages HTML, documents hypermédias, etc).

Le TALN est un champ de savoir et de techniques élaborés autour de problématiques diverses. Les concepts et techniques qu'il utilise se trouvent à la croisée de multiples champs disciplinaires : l'Intelligence Artificielle «traditionnelle», l'informatique théorique, la logique, la linguistique, mais aussi les neurosciences, les statistiques, etc.

Une des principales problématiques du TALN est que dans une langue en général, nous pouvons toujours exprimer la même idée avec plusieurs phrases différentes, ce qui pose un vrai problème d'ambiguïté, que les chercheurs, dans tous les domaines du traitement du langage, veulent résoudre.

Extraction d'information (EI), question réponse (QR), recherche d'information (RI), résumé automatique et traduction automatique sont des exemples d'applications qui ont besoin d'évaluer la relation sémantique entre des segments de textes, c'est-à-dire, si un segment de texte peut être sémantiquement déduit d'un autre.

Au début du traitement du langage naturel, le problème d'ambiguïté était dispersé dans ses différentes applications et chaque groupe de recherche traite le problème à sa façon, mais cela a produit une grande perte de temps. Pour cela, les chercheurs ont choisi d'unifier leurs forces pour créer un domaine qui a pour but de centraliser le problème d'ambiguïté et de proposer des méthodes de traitement du langage au niveau lexical, syntaxique et sémantique indépendamment d'une application donnée. La reconnaissance de l'inférence textuelle (RTE) est née.

Ainsi: on dira qu'un texte, noté T (texte), infère un texte, noté H (hypothése), si et seulement si H peut être inféré à partir de T (Dagan et al, 05).

Exemple d'inférence dite TRUE
T: Since its formation in 1948, Israel was involved in many wars with neighboring Arab countries.
H: Israel was established in 1948.

Exemple d'inférence dite FALSE
T: Since its formation in 1948, Israel was involved in many wars with neighboring Arab countries.
H: Israel was established before 1948.

Le Pascal RTE est un concoure qui à débuter en 2005 et son objectif et de comparer les réalisations des différents groupes de recherches travaillant sur le RTE.
Il y a eu trois compétitions Pascal RTE (2005, 2006 et 2007) et dans ces trois compétitions, les principales méthodes utilisées sont basées sur:

- le word matching (contage de mot) : l'inférence entre le texte T et H est vrai si le nombre de mot similaire entre les deux segments de textes est élevé.

Exemple:
T: Amine eats chocolates in the kitchen.
H : Amine eats chocolates.

Dans l'exemple l'inférence est considéré comme vrai par l'algorithme puisqu'il a 100 % des mots du texte H qui existe dans le texte T. nous appelons cette méthode le comptage de mots ou en anglais « le word matching ».

- l'inférence lexicale : T infère H si les mots contenus dans la phrase H peuvent être déduits de T après des transformations lexicales.

- les relations de dépendances syntaxiques (telles que les relations entre un verbe et ses arguments). Un matching entre les graphes de dépendances de T et H est alors effectué.

- l'inférence logique: transformer T et H en une représentation logique (souvent du premier ordre) puis vérifier si H est une déduction logique de T.

Pour le moment, les aspects temporels ne sont pas du tout abordés (reconnaissances des dates, expressions temporelles, événements, ordonnancement d'événements dans le temps, etc.) dans le RTE. Pour cela, notre projet, nommé TIMINF, pour « Time-inference », vise à modéliser, à développer et à évaluer l'apport de l'inférence temporelle dans le domaine de la reconnaissance de l'inférence textuelle (RTE).

# Motivation

Notre approche est motivée par les constatations suivantes :
La plupart des systèmes de détection d'inférence textuelle évalués au Pascal RTE, se sont focalisés sur les principales inférences (lexical, syntaxique et logique) et pour le moment, les aspects temporels ne sont pas du tout abordés.
Aussi les groupes travaillant sur les inférences temporells ne se basent que sur l'amélioration des détéctions des relations temporelles existentes entre évenements et expressions temporelles et n'essayent en aucun cas d'intégrer leurs travaux a un système d'inférence textuelle.

# Méthodologie de travail

Pour parvenir à la réalisation du système d'inférence textuelle intégrant l'inférence temporelle. Nous avons en premier lieu étudié les différents méthodes existantes dans la reconnaissance de l'inférence textuelle pour cela nous nous sommes basés sur les trois challenges qui se sont déroulés pour avoir un état des lieux sur les différentes méthodes existant.

Ensuite nous avons étudié la logique temporelle et son application sur le langage naturel, pour pouvoir avoir une idée de l'intégration du temps dans la langue.
Apres avoir étudié les différentes inférences textuelles et temporelles nous avons entamé l'étude des relations temporelles qui peuvent exister entre deux ségments de textes à travers un corpus que nous avons élaboré. La suite logique à notre projet est de concevoir notre systéme d'inférnece textuelle intégrant les différentes régles d'inférences temporelles découvertes au paravant.

Nous terminons notre travail avec l'évaluation de notre système et l'étude des différentes failles existentes en proposant quelques perspectives de recherche future.

## Plan du mémoire

Le plan que nous adoptons dans ce manuscrit reflète les différentes évolutions de notre projet. Ce document comporte cinq chapitres. Après avoir étudié les différentes approches adoptées pour traiter l'inférence textuelle dans le premier chapitre, le deuxième chapitre présente le temps dans la langue et aussi une étude sur l'inférence temporelle.

Dans le chapitre trois nous avons entrepris une démarche expérimentale à base de corpus afin de dégager différentes classes d'inférence temporelle. A partir de cette analyse, la seconde étape a été de concevoir l'architecture d'un système de reconnaissance d'inférence textuelle présenté dans le chapitre 4.

Enfin une fois le système conçu, nous nous sommes intéressés dans le dernier chapitre à l'évaluation des sorties de notre système en le confrontant à un corpus de test adapté.
Nous résumons, en conclusion de ce manuscrit, les différentes contributions de ce projet et nous donnons plusieurs pistes de recherches futures.

# Partie 1

# L'état de l'art

## Résumé

Avant d'entamer la conception de notre système d'inférence, nous avons besoin d'explorer les deux notions d'inférences textuelles et temporelles. Pour cela la partie état de l'art de notre mémoire est constituée de deux chapitres contenants successivement un large tour d'horizon sur l'inférence textuelle et ses différents niveaux de traitements. Le deuxième chapitre va contenir l'étude de la logique temporelle sous ses différentes facettes et les différentes techniques d'inférences temporelles existantes à nos jours.

# -Chapitre 1-

# LE TALN ET LE RTE

# Chapitre 1

# Le TALN et Le RTE

## 1) Introduction

Dans ce chapitre, nous commencerons par clarifier quelques concepts linguistiques, en étudiant les différents niveaux de représentation et de traitement des énoncés linguistiques. La section suivante est consacrée à l'étude de l'inférence textuelle où nous présentons les différentes applications du RTE et les principaux niveaux d'inférences textuelles nous détaillons les étapes de développement du challenge Pascale RTE qui a été mis en œuvre pour évaluer les avances des groupes de recherches dans ce domaine.
Nous terminons ce chapitre par la présentation de quelques méthodes d'inférences utilisées par des groupes de recherches évaluées dans le challenge Pascal RTE.

## 2) Brève historique du traitement automatique du langage naturel

Historiquement, les premiers travaux importants dans le domaine du TALN ont porté sur la traduction automatique, avec, dès 1954, la mise au point du premier traducteur automatique (très rudimentaire). Quelques phrases russes, sélectionnées à l'avance, furent traduites automatiquement en anglais.

Depuis 1954, de lourds financements ont été investis et de nombreuses recherches ont été lancées. Les principaux travaux présentés concernent alors la fabrication et la manipulation de dictionnaires électroniques, car les techniques de traduction consistent essentiellement à traduire mot à mot, avec ensuite un éventuel réarrangement de l'ordre des mots.

Cette conception simpliste de la traduction a conduit à l'exemple célèbre suivant : la phrase *The spirit is willing but the flesh is weak* (l'esprit est fort mais la chair est faible) fut traduite en russe puis retraduite en anglais.
Cela donna quelque chose comme : *The vodka is strong but the meat is rotten* (la vodka est forte mais la viande est pourrie) !

Ce qui ressort de cet exemple, c'est que de nombreuses connaissances contextuelles (i.e. portant sur la situation décrite) et encyclopédiques (i.e. portant sur le monde en général) sont nécessaires pour trouver la traduction correcte d'un mot (par exemple ici spirit, qui, suivant les contextes peut se traduire comme esprit ou comme alcool).

Posant comme conjecture que tout aspect de l'intelligence humaine peut être décrit de façon suffisamment précise pour qu'une machine le simule, les figures les plus marquantes de l'époque (John Mc Carthy, Marvin Minsky, Allan Newell, Herbert Simon) y discutent des possibilités de créer des programmes d'ordinateurs qui se comportent intelligemment, et en particulier qui soient capables d'utiliser le langage.

Aujourd'hui, le champ du traitement du langage naturel est un champ de recherche très actif. De nombreuses applications industrielles (traduction automatique, recherche documentaire, interfaces en langage naturel), qui commencent à atteindre le grand public, sont là pour témoigner de l'importance des avancées accomplies mais également des progrès qu'il reste encore à accomplir.

# 3) Les niveaux de traitement

Nous introduisons dans cette section les différents niveaux de traitements nécessaires pour parvenir à une compréhension complète d'un énoncé en langage naturel. Ces niveaux correspondent à des modules qu'il faudrait développer et faire coopérer dans le cadre d'une application complète de traitement de la langue.

Nous considérons à titre d'exemple l'énoncé suivant :

*(1) Le président des antialcooliques mangeait une pomme avec un couteau,*

Nous envisageons les traitements successifs qu'il convient d'appliquer à cet énoncé pour parvenir automatiquement à sa compréhension la plus complète. Il nous faudra successivement :
– identifier les composants lexicaux, et leurs propriétés : c'est l'étape de traitement **lexical** ;
– identifier des constituants (groupe) de plus haut niveau, et les relations (de dominance) qu'ils entretiennent entre eux : c'est l'étape de traitement **syntaxique** ;
– construire une représentation du sens de cet énoncé, en associant à chaque concept évoqué un objet ou une action dans un monde de référence (réel ou imaginaire) : c'est l'étape de traitement **sémantique**.
– identifier enfin la fonction de l'énoncé dans le contexte particulier de la situation dans lequel il a été produit : c'est l'étape de traitement **pragmatique.**

## 3.1) Le niveau lexical

Le but de cette étape de traitement est de passer des formes atomiques (tokens) identifiées par le segmenteur de mots (Nugues, 2006), c'est-à-dire de reconnaître dans chaque chaîne de caractères une (ou plusieurs) unité(s) linguistique(s), dotée(s) de caractéristiques propres (son sens, sa prononciation, ses propriétés syntaxiques, etc).

Selon l'exemple *(1)*, l'étape d'identification lexicale devrait conduire à un résultat voisin de celui donné ci-dessous, dans lequel on peut constater en particulier l'ambiguïté d'une forme telle que président: cette chaîne correspond à deux formes du verbe présider (indicatif et subjonctif), ainsi à une forme nominale, et sa prononciation diffère selon qu'elle représente un nom ou un verbe.

On conçoit aisément que pour les mots les plus fréquents, comme « le », la solution la plus simple est de rechercher la forme dans (un lexique)[1] précompilé. Dans les faits, c'est effectivement ce qui se passe, y compris pour des formes plus rares, dans la mesure où l'utilisation des formalismes de représentations compacts permettant un accès optimisé (par exemple sous la forme d'automates d'états finis), et l'augmentation de la taille des mémoires rend possible la manipulation de vastes lexiques (de l'ordre de centaines de milliers de formes).

Pour autant, cette solution ne résout pas tous les problèmes. Le langage est création, et de nouvelles formes surgissent tous les jours, que ce soit par emprunt à d'autres langues (il n'y a qu'a écouté parler les enseignants des autres modules de la dominante informatique !), ou, plus fréquemment, par l'application de procédés réguliers de créations de mots, qui nous permettent de composer pratiquement à volonté de nouvelles formes immédiatement compréhensibles par tous les locuteurs de notre langue : si j'aime lire Proust, ne peut-on pas dire que je m'emproustise, que de proustien je deviens proustiste, voire proustophile, puis que, lassé, je me désemproustise... Ce phénomène n'a rien de marginal, puisqu'il est admis que, même si l'on dispose d'un lexique complet du français, environ 5 à 10 % des mots d'un article de journal pris au hasard ne figureront pas dans ce lexique. La solution purement lexicale atteint là ses limites, et il faut donc mettre en œuvre d'autres approches, de manière à traiter aussi les formes hors-lexiques.

## 3.2) Le niveau syntaxique

La syntaxe est l'étude des contraintes portant sur les successions licites de formes qui doivent être prises en compte lorsque l'on cherche à décrire les séquences constituant des phrases grammaticalement correctes: toutes les suites de mots ne forment pas des phrases acceptables (Ligauzat, 1994).
La description des contraintes caractéristiques d'une langue donnée se fait par le biais d'une grammaire.

Les modèles et les formalismes grammaticaux proposés dans le cadre du traitement automatique du langage sont particulièrement nombreux et variés.
Le niveau syntaxique est donc le niveau conceptuel concerné par le calcul de la validité de certaines séquences de mots, les séquences grammaticales ou bien-formées. On conçoit bien l'importance d'un tel traitement dans une application de génération, pour laquelle il est essentiel que la machine engendre des énoncés corrects. Dans une application de compréhension, la machine analyse des textes qui lui sont fournis, et dont on peut supposer qu'ils sont grammaticaux. Pourquoi donc, dans ce cas, mettre en œuvre des connaissances syntaxiques ?

Une première motivation provient du fait que les textes ne sont pas toujours grammaticaux, par exemple à cause des fautes d'orthographes. Une analyse syntaxique peut donc permettre de choisir entre plusieurs corrections à apporter à une phrase incorrecte, mais également se révéler bien utile pour améliorer les sorties d'un système de reconnaissance optique de caractère ou d'encore un système de reconnaissance de la parole.

---

[1] En linguistique, le lexique d'une langue constitue l'ensemble de ses lemmes ou, d'une manière plus courante mais moins précise, « l'ensemble de ses mots ». Toujours dans les usages courants, on utilise, plus facilement le terme *vocabulaire*.

Une seconde raison est que l'entrée du module syntaxique est une série de formes étiquetées morpho syntaxiquement, une forme pouvant avoir plusieurs étiquettes différentes. Une première fonction du module syntaxique consiste donc à désambiguïser la suite d'étiquettes, en éliminant les séquences qui correspondent à des énoncés grammaticalement invalides.

## 3.3) Le niveau sémantique

Intuitivement, la sémantique se préoccupe du sens des énoncés (yvon, 2007). Une phrase comme *Le jardin de la porte mange le ciel*, bien que grammaticalement parfaitement correcte, n'a pas de sens dans la plupart des contextes. Mais qu'est ce que le sens ? Pour une expression comme *la bouteille de droite* dans la phrase :

*Sers-toi du vin. Non, pas celui-là, prends la bouteille de droite.*

Le sens correspond à l'objet (au concept) désigné. Dans cet exemple, le sens dépend étroitement du contexte : il faut une représentation de la scène pour savoir de quelle bouteille, et donc de quel vin, il s'agit.

Pour une expression prédicative, comme *Il commande un Margaux 1982*, le sens peut être représenté par un prédicat logique comme <demander(paul,chateau_margaux_82)>. L'identification d'un tel prédicat dépend encore une fois du contexte. Le verbe commander aurait en effet renvoyé à un autre prédicat s'il s'agissait de commander un navire.

## 3.4) Le niveau pragmatique

Le niveau pragmatique est parfaitement dissociable du niveau sémantique. Alors que la sémantique se préoccupe du sens des énoncés, la pragmatique porte sur les attitudes (vérité, désirabilité, probabilité) que les locuteurs adoptent vis à vis des énoncés et sur les opérations logiques que ces attitudes déclenchent (yvon, 2007).

Historiquement, certains linguistes ont appelé pragmatique tout traitement du langage faisant intervenir le contexte d'énonciation. Ce critère présente fort peu d'intérêt, dans la mesure où les processus sémantiques sont les mêmes, que le contexte intervienne ou non. En revanche, il existe une distinction très importante, basée sur la notion d'inférence logique. Considérons l'exemple suivant :

(a) Pierre : viendras-tu au bal ce soir ?

(b) Marie : j'ai entendu que Paul y sera !

La seconde phrase sera interprétée comme une réponse négative si l'on sait que *Marie* n'aime pas *Paul*.

Cette interprétation n'est pas de nature sémantique. À partir de la compréhension du sens de l'intervention de *Marie*, *Pierre* réalise une inférence logique en utilisant une connaissance contextuelle, l'inimitié entre Paul et Marie. Pierre conclut que Marie ne veut pas aller au bal, autrement dit il reconstruit l'attitude de Marie par rapport à son propre énoncé. Cette opération n'est pas une construction conceptuelle, c'est une opération logique. Elle appartient donc à la pragmatique.

Les techniques correspondant à ce niveau de traitement sont encore très mal maîtrisées. Le niveau pragmatique, même si les techniques qui lui correspondent ne sont pas encore stabilisées, apparaît moins difficile à aborder que le niveau sémantique. Il semble en effet qu'il repose sur un ensemble de principes fixes, comme le principe de pertinence, qu'il s'agit de modéliser correctement. La détermination de l'intention argumentative de l'auteur ou du locuteur est essentielle dans bon nombre d'applications, notamment la gestion de dialogue, le résumé de texte, la traduction automatique, les systèmes d'aide contextuelle ou d'enseignement, etc. On attend donc des progrès significatifs à ce niveau dans les années qui viennent.

# 4) Les difficultés du TALN : ambiguïté

Le langage naturel est ambigu, et cette ambiguïté se manifeste par la multitude d'interprétations possibles pour chacune des entités linguistiques pertinentes pour un niveau de traitement, comme en témoignent les exemples suivants :

## 4.1) Ambiguïté des graphèmes (lettres)

Cette ambiguïté existe dans le processus d'encodage orthographique en comparant la prononciation du i dans lit, poire et maison.

## 4.2) Ambiguïté dans les propriétés grammaticales et sémantiques

Ainsi *mange* est ambigu à la fois morpho-syntaxiquement, puisqu'il correspond aux formes indicatives et subjonctives du verbe manger), mais aussi sémantiquement. En effet, cette forme peut aussi bien référer (dans un style familier) à un ensemble d'actions conventionnelles (comme de s'assoir à une table, mettre une serviette, utiliser divers ustensiles, ceci éventuellement en maintenant une interaction avec un autre humain) avec pour vision finale d'ingérer de la nourriture (auquel il ne requière pas de complément d'objet direct); et à l'action consistant à effectivement ingérer un type particulier de nourriture (auquel cas il requiert un complément d'objet direct), etc. Comparez en effet :

*(a) Demain, Paul mange avec ma sœur.*
*(b) Paul mange son pain au chocolat.*

Ainsi que les déductions que l'on peut faire à partir de ces deux énoncés : de (a), on peut raisonnablement conclure que Paul sera assis à une table, disposera de couverts,... ; tout ceci n'est pas nécessairement vrai dans le cas de l'énoncé (b).

## 4.3) Ambiguïté de la fonction grammaticale des groupes de mots

L'ambiguïté est illustrée par la phrase :
*il poursuit la jeune fille à vélo*.
Dans cet exemple *à vélo* est soit un complément de manière de poursuivre (et c'est *il* qui pédale), soit un complément de nom de fille (et c'est *elle* qui mouline) ;

## 4.4) Ambiguïté de la portée des quantificateurs, des conjonctions et des prépositions

Ainsi, dans *Tous mes amis ont pris un verre,* nous pouvons supposer que chacun avait un verre différent, mais dans *Tous les témoins ont entendu un cri,* il est probable que c'était le même cri pour tous les témoins. De même, lorsque l'on évoque les chiens et les chats de Paul, l'interprétation la plus naturelle consiste à comprendre de Paul comme le complément de nom du groupe les chats et les chiens ; cette lecture est beaucoup moins naturelle dans les chiens de race et les chats de Paul ;

## 4.5) Ambiguïté sur l'interprétation à donner en contexte à un énoncé

Nous comparons ainsi la « signification » de non, dans les deux échanges suivants :

*(a) Si je vais en cours demain ? Non (négation)*
*(b) Tu vas en cours demain ! Non ! (j'y crois pas).*

En effet, l'ambiguïté est un problème majeur du TALN. Pour y pallier les chercheurs ont crée un domaine qui a pour but de centraliser ce problème et de proposer des méthodes de traitement du langage au niveau lexical, syntaxique et sémantique indépendamment d'une application donnée. Dans ce qui suit nous allons explorer ce domaine ainsi que ces différentes applications.

# 5) La reconnaissance de l'inférence textuelle (RTE)

## 5.1) Introduction

Le RTE est un domaine de recherche assez récent en traitement du langage (2005) qui a pour but de fédérer les recherches en TALN afin de proposer des méthodes de traitement du langage au niveau lexical, syntaxique et sémantique indépendamment d'une application donnée (résumé automatique, système de question réponse ou encore la recherche d'information).
Le RTE vise à déterminer automatiquement si un segment de texte (H) est déduit d'un autre segment de texte (T) (Dagan et al, 05).

*Exemple :*

*T : « Amine a 40 degrés de **fièvre**, sa mère l'a pris immédiatement à l'**hôpital** ».*

*H : « Amine est **malade** ».*

Dans l'exemple ci dessus, comprendre que le segment H est déduit du segment T, est une déduction simple pour l'être humain, mais pour la machine c'est tout autre. Pour cela, les chercheurs ont proposé plusieurs approches pour résoudre le problème.

Dans l'exemple, pour dire que H est inféré de T le système doit lier le fait d'être malade (texte H) avec le mot hôpital et fièvre (texte T) pour déduire qu'il y a inférence.

Dans cette section, nous présentons les différentes applications du RTE, puis nous détaillons les étapes de développement du challenge Pascale RTE qui a été mis en œuvre pour évaluer les avances des groupes de recherches dans ce domaine.

Nous développons dans la section 2, les principaux niveaux d'inférences textuelles et nous terminons ce chapitre par la présentation de quelques méthodes d'inférences utilisées par des groupes de recherches évaluées dans le challenge pascal RTE.

## 5.2) Les applications du RTE

L'inférence entre des segments de textes est au cœur de plusieurs applications du traitement automatique du langage naturel (TALN). Nous décrivons dans ce qui suit comment le RTE contribue dans ces différents domaines :

### 5.2.1) La recherche d'information

La recherche d'information est la science qui consiste à rechercher l'information dans des documents, des bases de données, qu'elles soient relationnelles ou mises en réseau par des liens hypertextes (Joachims, 2003).

La recherche d'information est un domaine historiquement lié aux sciences de l'information et à la bibliothéconomie qui ont toujours eu le souci d'établir des représentations des documents dans le but d'en récupérer des informations, à travers la construction d'index. L'informatique a permis le développement d'outils pour traiter l'information et à établir la représentation des documents au moment de leur indexation, ainsi que pour rechercher l'information.
Les approches qui étaient utilisées auparavant se basaient sur la recherche de mots clés dans les textes. Le problème dans ces systèmes c'est qu'ils ne prennent en compte ni les relations entre les mots clés ni leurs sens.

Exemple 1 :

**Figure 1.1 : Exemple de moteur de recherche a base de mot clé**

Dans cet exemple (Figure 1.1) nous remarquons qu'un moteur de recherche fonctionnant à base de mot clé comme Google fait bien ce type de recherche et répond bien à la question simple comme « the first president Algerie » puisque la simple recherche des mots clés dans les différents documents permet de donner une bonne réponse à l'utilisateur.

Exemple 2 :

**Figure 1.2 : Exemple où le moteur de recherche à base de mot clé ne marche pas**

Dans cet exemple (Figure 1.2) nous remarquons que l'utilisation des mots clés seuls peut nous mener à un document qui n'a aucune relation avec notre requête et qui montre que l'inférence sémantique est indispensable à la recherche d'information.

## 5.2.2) L'extraction d'information

L'extraction d'information consiste à identifier l'information bien précise d'un texte en langue naturelle et à la représenter sous forme structurée. Par exemple, à partir d'un rapport sur un accident d'automobile, un système d'extraction d'information sera capable d'identifier la date et le lieu de l'accident, le type d'incident, ainsi que les victimes. Ces informations pourront ensuite être stockées dans une base de données pour y effectuer des recherches ultérieures ou être utilisées comme base à la génération automatique de résumés (Kosseim., 2005).

L'extraction d'information s'avère très pratique dans l'industrie où des opérations d'extractions y sont quotidiennement effectuées à la main. Nous pensons, par exemple, au traitement de rapports de filature d'une agence de surveillance, à la gestion de dépêches d'une agence de presse, à la manipulation de rapports d'incidents d'une compagnie d'assurances, etc.
Un système d'extraction d'information permet de traiter automatiquement et plus rapidement de grandes quantités de documents.

Dans ce cas de figure le RTE donne son apport dans la détection de l'information.

## 5.2.3) Le système question- réponse

Les systèmes Questions/Réponses sont capables de répondre à des questions écrites en langage naturel en cherchant la réponse dans un corpus de textes. Ils sont classiquement constitués d'un ensemble de modules réalisant respectivement : une analyse de la question, une recherche de portions de documents pertinents et une extraction de la réponse à l'aide de motifs d'extractions, ou patterns en anglais (Nyberg et al, 2002).
Le système doit identifier le segment de texte qui contient la réponse. L'inférence entre le texte T et le segment H peut aider à détecter le segment qui contient la réponse.

Exemple :
H : « who is Ariel Sharon ? ».
T : « Israel's Prime Minister, Ariel Sharon, visited Prague ».
Le système effectue d'abord une transformation à l'affirmatif de la question « Ariel Sharon is Isreal's Prime Minister » puis une comparaison entre le segment de texte T et le segment H.
Si H est inféré de T comme dans l'exemple alors T est accepté comme un segment contenant la réponse à la question H.

## 5.2.4) La traduction automatique

La traduction automatique désigne, au sens strict, le fait de traduire entièrement un texte grâce à un ou plusieurs programmes informatiques, sans qu'un traducteur humain n'ait à intervenir (Laurian et Marie, 1996). La traduction automatique est encore très imparfaite, et la génération de traduction d'une qualité comparable à celle de traducteurs humains relève encore de l'utopie.

Pour évaluer les performances de la machine, le RTE permet de comparer la traduction faite par la machine avec celle faite par l'humain.

## 5.2.5) Le résumé automatique

Le résumé automatique se propose de faire une extraction de l'information jugée importante d'un texte d'entré pour construire, à partir de cette information, un nouveau texte de sortie, condensé. Ce nouveau texte permet d'éviter la lecture en entier du document source.

Le RTE est utilisé pour trouver les redondances d'informations.
Si un segment de texte infère un autre, un des deux va être supprimé.
En particulier c'est intéressant dans les applications qui font le résumé de plusieurs documents. S'il y a plusieurs documents qui relatent le même fait, un seul doit être pris.

## 5.2.6) L'acquisition des Paraphrases (AP)

Une paraphrase, c'est le fait de dire avec d'autres mots, d'autres termes ce qui est dit dans un texte, un paragraphe.

Dans ce cas de figure le RTE est utilisé pour détecter l'inférence entre le texte paraphrasé et le texte d'origine. Comme dans l'exemple suivant où les deux phrases ont le même sens avec juste une autre disposition des mots dans la phrase.
Exemple :
T : « Ce médicament est commercialisé au Canada seulement ».
H **:** « La commercialisation de ce médicament s'est effectuée au Canada seulement ».

# 5.3) Le challenge "PASCAL Recognizing of Textual Entailment"

Le Pascal recognition of Textual Entailment est un concours qui a débuté en 2005. Il se déroule chaque année et son objectif, est de fournir à la communauté du TAL un nouveau point de repère pour vérifier les progrès dans la reconnaissance l'inférence textuelle, et de comparer les réalisations des différents groupes de recherches travaillant dans ce domaine ( http://www.pascal-network.org/Challenges/RTE/ ).

Suite au succès du premier RTE un nouveau RTE a été organisé, avec 23 groupes venus du monde entier (par rapport à 17 pour le premier défi) qui ont présenté les résultats de leurs systèmes. Les représentants des groupes participants ont présenté leurs travaux au **PASCAL Challenges atelier** en avril 2006 à Venise, Italie.

L'événement a été un succès et le nombre de participants et leurs contributions à la discussion ont démontré que le Textual Entailment est un domaine en expansion rapide. Déjà, les ateliers ont donné naissance à un nombre impressionnant de publications dans les grandes conférences, en plus des travaux en cours.
Les démarches entreprises pour réaliser le concours sont :
- Préparation du corpus.
- Etablissement des mesures d'évaluations.

Dans ce qui suit les démarches citées sont détaillées.

## 5.3.1) La préparation du corpus

La première étape à entreprendre consiste à créer le corpus de texte-hypothèse (T-H) pair de petit segment de texte, qui correspond à des informations collectées à travers le web dans des domaines différents.
Les exemples ont été collectés manuellement pour l'inférence par des annotateurs humains.
Les exemples ont été divisés en deux types de corpus (**Corpus de développement** et **Corpus de test**).

**Le corpus de développement** est utilisé au début de challenge pour donner aux utilisateurs la possibilité de tester leurs systèmes et de faire des petites mises au point pour se préparer au test.

**Le corpus de test** est utilisé pour l'évaluation finale.
1. Pour le RTE 1 Le corpus était composé de 567 paires de (H-T) pour le développement et 800 pairs pour le test.

Le choix d'un large corpus est justifié par la nécessité d'avoir des résultats statistiques significatifs.

Le corpus est collecté en respectant les différentes applications du traitement de langage naturel (QR, RI, IE., PP…) et la collecte des exemples est faite par niveau d'inférence :
L'analyse lexique, syntaxique, logique et connaissance du monde, et les différents niveaux de difficultés.

```
<pair id="754" value="TRUE" task="CD">
    <t>
Mexico City has a very bad pollution problem because the mountains
    around the city act as walls and block in dust and smog.
</t>
    <h> Poor air circulation out of the mountain-walled Mexico
    City aggravates pollution.</h>
</pair>
Id : représente le numéro de la pair.
Value : représente la décision de l'annotateur (vrai ou faux).
Task : représente le type de l'application ou l'inférence existe.
```

**Figure 1.3 : Exemple du corpus annoté**

Le corpus doit inclure 50% d'un exemple de T-H correspondant à de vraies inférences et 50% de fausses inférences. Pour cela, chaque exemple (T-H) est jugé vrai ou faux par l'annotateur qui crée l'exemple.
Puis l'exemple est évalué par un second juge qui évalue les paires de textes et d'hypothèses, sans avoir pris conscience de leurs contextes.

Les annotateurs étaient d'accord avec le jugement dans 80% des exemples, ce qui correspond à 0.6 Kappa[2], les 20% du corpus où il n'y a pas eu d'accord ont été supprimés). Le reste du corpus est considéré comme un «gold standard» ou « BASELINE » pour l'évaluation.

Le but de cette manœuvre est de créer un corpus où il n'y aura pas de jugements controverses. Pour effectuer leurs jugements et annoter le corpus les annotateurs suivent des directives. Dans ce qui suit, nous allons citer les différentes directives qui étaient prises en considération.

## 5.3.2) Les directives de jugements

- L'inférence est une relation à un seul sens.
- L'hypothèse doit être inférée d'un texte, mais le texte ne doit pas forcement être inféré de l'hypothèse.

- L'hypothèse doit être inférée entièrement du texte. L'inférence est fausse s'il reste une partie de l'hypothèse qui ne peut être inférée par le texte.

---

[2] Kappa (J.Cohen, 1960) :c'est une mesure statistique pour calculer a quel point deux personnes (ou groupes de personnes) A et B sont d'accord pour classer N éléments dans K catégories mutuellement exclusives.

- les cas où l'inférence est probable doit être jugé comme vrai.

- il est autorisé d'utiliser les connaissances du monde comme dans l'exemple *le chiffre d'affaire de Google est de 50 millions de dollars*. On doit savoir que Google est une entreprise donc on peut lui attribuer la possibilité d'avoir un chiffre d'affaire.

### 5.3.3) Les mesures d'évaluation

Le système d'annotation du corpus adopté dans les deux challenges précédant est binaire, c'est-à-dire que le système donne deux résultats possibles soit l'inférence entre les deux textes est vrai ou fausse}.

Le résultat est comparé au '**GOLD standard'**, et le pourcentage donnant le nombre de fois où il y a similitude entre le système et le '**gold standard'** donne **'l'accuracy'** du système.

L'accuracy est une mesure standard dans les systèmes de traitement du langage naturel. Elle est fréquemment utilisée pour évaluer les performances des applications, (Beyer et al. 2005). Elle est calculée comme ceci.

**Accuracy** = **X** / **Y**.

Où :

**X** : représente le nombre de fois où les résultats du système sont similaires au gold standard.

**Y** : représente le nombre de paires contenu dans le corpus de test.

Par exemple Le nombre de résultats similaires est de 500 paires et le corpus est de 800 paires, l'accuracy est de 500/800 qui est égale à 62,5%.

## 5.4) L'analyse des principales méthodes utilisées

Dans ce qui suit, nous allons présenter les différentes étapes de traitements effectuées pour détecter l'inférence textuelle.

### 5.4.1) Les prétraitements

Quelque soit la technique adoptée pour effectuer l'inférence textuelle, il est nécessaire de pré traiter les données brutes avant d'appliquer les techniques d'inférences.

Dans le RTE trois niveaux de prétraitements ont été utilisés:
- Niveau lexical pour éviter les problèmes liés à la morphologie de mots.
- Niveau syntaxique pour pouvoir donner une structure préalable au texte.
- Niveau sémantique pour analyser les sens des mots.

Ci-dessous nous allons présenter les différents niveaux de prétraitements existants et utiliser pour l'inférence textuelle.

### 5.4.1.1) Le Niveau lexical

L'objectif du prétraitement au niveau du "mot" est de réduire les variations dues à la morphologie et d'éviter que des petites erreurs initiales se propagent dans toutes les étapes du traitement. Pour cela, différentes transformations ont été introduites :

#### A) La tokenisation

L'objectif de la tokenisation est de trouver les unités de base du "sens " dans les textes. Pour cela, les systèmes doivent résoudre différents problèmes comme la gestion des blancs, de la ponctuation, des retours lignes et des fins de paragraphes.

### B) La lemmatisation

La lemmatisation d'une forme d'un mot consiste à en prendre sa forme canonique. Celle-ci est définie comme ceci :
Quand c'est un verbe on doit le mètre à l'infinitif :

Exemple :
Parti (verbe) -> partir

Pour les autres mots, ils doivent être mis au masculin singulier.

Exemple :
Parti (nom) -> parti

Pour effectuer l'analyse lexicale, différents outil qui ont été mis en point. Le TreeTagger est un des outils le plus utilisés pour la langue anglaise.

**Le TreeTagger** effectue une tokinisation, une lemmatisation et un étiquetage comme le montre l'exemple suivant :

Exemple d'entrée dans le **TreeTagger** : « Le TreeTagger est facile à utiliser ».

La figure suivante reprend la sortie du logiciel.

| Tokenisation | Étiquetage | Lemmatisation |
|---|---|---|
| Le | DT | La |
| TreeTagger | NP | TreeTagger |
| Est | VBZ | Être |
| Facile | JJ | Facile |
| À | D' | À |
| Utiliser | VB | Utiliser |

**Figure 1.4 : Sortie du TreeTagger**

### 5.4.1.2) Le niveau syntaxique

L'objectif de cette étape est de décrire les structures de phrases possibles et d'analyser les phrases en structures.
La structure révélée par l'analyse donne alors précisément la façon dont les règles syntaxiques sont combinées dans le texte. Cette structure est souvent une hiérarchie de syntagmes, représentée par un arbre syntaxique dont les nœuds peuvent être décorés (dotés d'informations complémentaires).

Nous illustrons cette analyse avec la sortie d'un des outils utilisés dans l'annotation syntaxique (SYNTEX)[3].

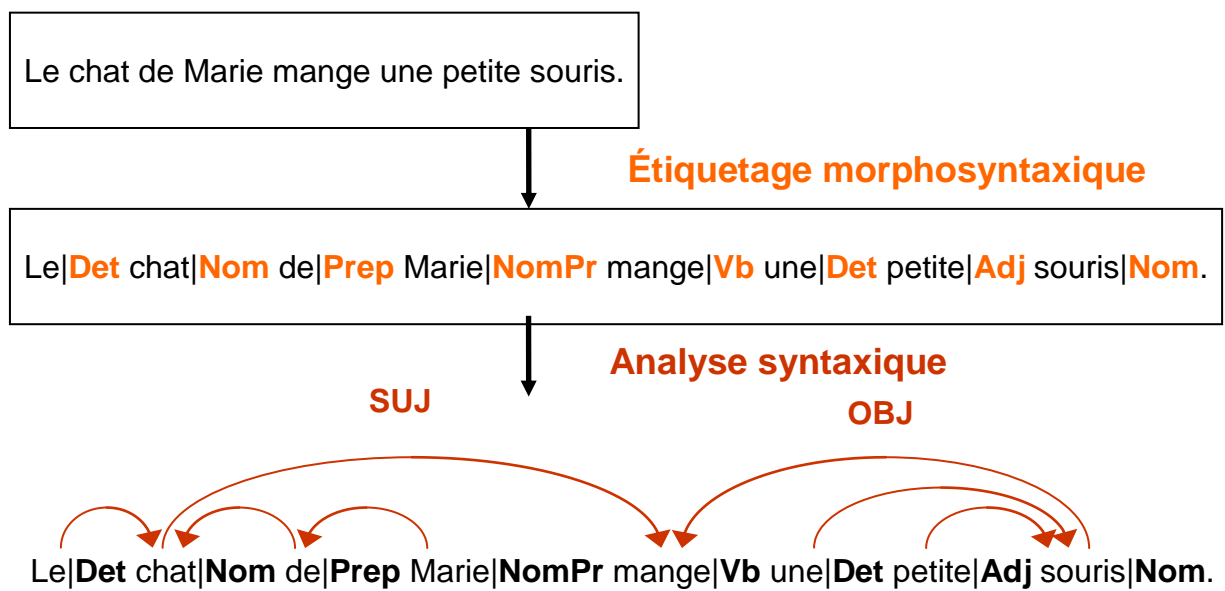

**Figure 1.5 : Exemple d'annotation syntaxique**

Nous remarquons dans l'exemple ci-dessus que l'analyse morphosyntaxique permet d'étiqueter les mots et l'analyse syntaxique permet de les relier entre eux.

### 5.4.1.3) Le niveau sémantique

Pour simplifier, nous pouvons dire que l'analyse sémantique s'appuie, entre autres, sur la compréhension du sens des mots des textes, contrairement aux analyses lexicales ou grammaticales, qui analysent les mots à partir du lexique ou de la grammaire. Dans le cadre de l'analyse sémantique, il est donc fondamental d'analyser le sens des mots pour comprendre ce qu'on dit. Pour cela plusieurs approches ont été adoptées pour annoter les relations entre les mots pour mieux cerner leur sens. Une de ces approches est la structure prédicat argument qui est expliquée ci-dessous.

La structure que nous appelons prédicative est un graphe de relation prédicat-argument, où les prédicats représentent l'action.
Une relation prédicative correspond à une relation de dépendance syntaxique. Le prédicat peut avoir plusieurs types d'arguments (sujet, complément d'objet direct et complément d'objet indirect).

---

[3] La fonction de cet analyseur est d'identifier des relations de dépendances entre mots et d'extraire d'un corpus des syntagmes (verbaux, nominaux, adjectivaux) (Bourigault, 2000).

Exemple :

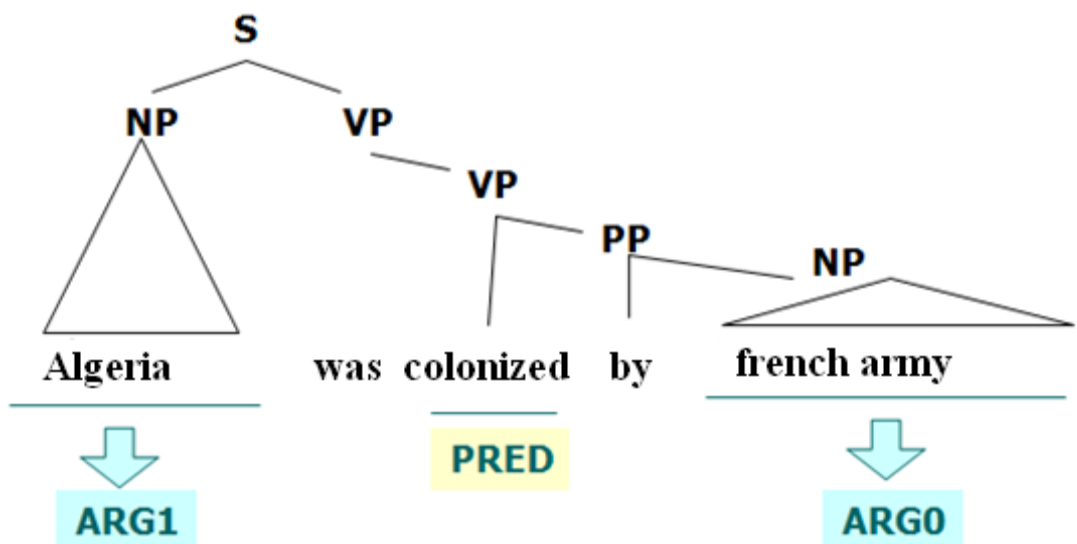

**Figure 1.6 : Exemple de structure prédicat argument**

## 5.4.2) Les différents niveaux d'inférence textuelle

Dans cette section nous allons présenter les différents niveaux d'inférences (Lexical, lexico syntaxique, sémantique (logique) et connaissance du monde) utilisées pour la détection de l'inférence textuelle.

### 5.4.2.1) L'inférence au niveau lexical

A ce niveau, l'inférence entre deux segments de textes est accepté s'il existe des mots semblables entre T et H, où les mots contenus dans la phrase H peuvent être inférés de T après des transformations lexicales (vanderwede et al., 2005). Les trois techniques d'inférence sont ci-dessous :

#### A) Les dérivations morphologiques

Ce mécanisme d'inférence considère que deux des termes sont équivalents si l'un peut être obtenu de l'autre après une dérivation morphologique. Il existe trois type de dérivations morphologiques :

- **La normalisation**

Exemple :
T : « l'**acquisition** d'un AIRBUS A380 par le roi FAHD ».
H : « le roi FAHD **a acquis** un AIRBUS A380 ».

La transformation <*d'acquisition*> en <*a acquis* > a permis de faire la déduction de l'inférence entre les deux textes.

- **La dérivation nominale**

Exemple

T : Le GIA a donne de la **terreur** au peuple algérien.

H : Le GIA est un groupe **terroriste.**

La transformation de **terreur** en **terroriste** a permis de faire la déduction de l'inférence entre les deux textes.

- **Les relations entre noms et verbes**

Exemple

T : Mark gagne à tous les coups.

H : Mark est un gagnant.

La transformation de **Mark est un gagnant** en **Mark gagne** a permis de faire la déduction de l'inférence entre les deux textes.

## B) Les relations ontologiques

Une ontologie est un ensemble structuré de concepts permettant de donner un sens aux informations. Elle est aussi un modèle de données qui représente un ensemble de concepts dans un domaine et les rapports entre ces concepts (Bourigault, 2004). Elle est employée pour raisonner au sujet des objets dans ce domaine.
Les concepts sont organisés dans un graphe dont les relations peuvent être : des relations sémantiques et des relations de subsomption.
L'objectif premier d'une ontologie est de modéliser un ensemble de connaissances dans un domaine donné.
Ce mécanisme d'inférence se réfère à la relation **ontologique** qui existe entre deux termes. Ces différentes relations sont citées ci dessous.

- **La synonymie**

Représente un ensemble de mots interchangeables dans un contexte donné. Elle est souvent utilisée pour reconnaître l'inférence.

Exemple

T : « Jane a **abattue** Mark ».

H : « Jane a **tué** Mark ».

Autre exemple comme (''commencer''/''démarrer''), (''enlever ''/'' retirer'').

- **La généralisation (hypernymie)**

La relation d'Hypernymie est le terme générique utilisé pour désigner une classe englobant des instances de classes plus spécifiques. Y est un hypernyme de X si X est un type de Y.

Exemple

T : « On a coupé le **sapin** »**.**

H : « On a coupé **l'arbre** ».

La relation entre l'arbre et le sapin (l'arbre est une généralisation sapin) a permis l'inférence entre les deux textes.

- **L'hyponymie**

La relation Hyponymie est le terme spécifique utilisé pour désigner un membre d'une classe (relation inverse de Hypernymie). X est un hyponyme de Y si X est un type de Y.

Exemple

T : John a pris un **moyen de transport pour terrestre** pour faire le trajet Toulouse paris.

H : John a fait Toulouse Paris en **TGV**.

La relation entre **moyen de transport pour terrestre** et **TGV** qui a permis l'inférence entre les deux textes.

- **La relation de Méronymie**

X est un méronyme de Y si X est une partie de Y.

Exemple :

{Avion} a comme méronyme {{porte}, {moteur}} ;

## C) La connaissance du monde dans l'analyse lexique

Ce mécanisme d'inférence se réfère à la connaissance du monde pour détecter l'inférence au niveau lexical (Len Schubert, 2002).

Exemple :

‘'Taliban ➔ organisation ''et ''yahoo ➔ moteur de recherche ''

## 5.4.2.2) L'inférence au niveau lexico syntaxique

Au niveau lexico syntaxique l'hypothèse est représentée par des relations de dépendances syntaxiques.
La relation d'inférence entre T et H est définit comme un recouvrement des relations de H par les relations de T, ou le recouvrement est obtenu après une séquence de transformation appliquée à la relation de T. Les différents s types de transformations sont spécifies par :

### A) Les transformations syntaxiques

Dans ce mécanisme d'inférence, la transformation se fait entre les structures syntaxiques qui ont les mêmes éléments lexicaux et préservent le sens de la relation entre elles (Vanderwende et al..,2005).
Ce genre de mécanisme inclut la transformation passive active et l'apposition[4].
Exemple :
« Mon chat, ce gentil petit siamois, est assis sur cette table ». « Il peut devenir :   Mon chat est assis sur cette table, ce gentil petit siamois ! ».

### B) L'inférence basée sur les paraphrases

Dans ce mécanisme d'inférence, la transformation modifie la structure syntaxique du segment du texte et quelques éléments lexicaux, mais elle garde la relation d'inférence entre le segment de texte original et celui qui est transformé.
Ce type de relation entre les deux segments est appelé dans la littérature « Paraphrase ». Des méthodes pour effectuer la transformation sont proposées dans (Lin et Pantel, 2001).
Exemple :
T : « Ce médicament est commercialisé au Canada seulement ».
H **:** « La commercialisation de ce médicament s'est effectuée au Canada seulement ».

### C) La coréférence

La relation de coréférence met en relation un pronom et un antécédent éloigné l'un de l'autre dans la phrase. Par exemple :
« **L'Italie et l'Allemagne** ont tous deux joué deux matchs, **ils** n'ont perdu aucun match encore ».
**Infère à**
 « Ni l'Italie ni l'Allemagne n'a encore perdu un match », cela inclut la transformation de coréférence « **ils  ➔ l'Italie et l'Allemagne ».**

---

[4] L'apposition est une construction grammaticale dans laquelle deux éléments, normalement substantif expressions, sont placés à côté de l'autre, avec un élément servant à définir ou modifier les autres.. Lorsque ce dispositif est utilisé, les deux éléments sont censés être à l'apposition. Par exemple, dans l'expression "mon ami Alice" le nom "Alice" est à l'apposition de "mon ami".

### 5.4.2.3) L'inférence sémantique (logique)

A ce niveau, l'inférence entre deux segments de textes est acceptée si le sens des deux phrases se concorde. En d'autre termes, l'inférence textuelle est considérée comme un problème d'implication logique entre les sens des deux phrases (Tatu et al., 2006).
Pour cela, la structure prédicat argument est souvent utilisée, c'est-à-dire que, les segments de textes T et H sont transformés en prédicat et à travers des déductions logiques comme par exemple l'utilisation de la (preuve par réfutation[5]) on arrive à déduire l'inférence.
Un exemple des systèmes utilisant cette méthode d'inférence est décrit dans la section (**5.5.4.2**).

## 5.4.3) Les ressources utilisées

Dans les différents techniques d'inférence textuelle plusieurs ressources sont utilisées (WordNet, framnet, Cyc…). L'ensemble constitue un « écosystème » complet couvrant des aspects lexicaux, syntaxiques et sémantiques. Combinées, ces ressources fournissent un point de départ intéressant pour des développements sémantiques en TAL ou dans le cadre du Web sémantique, telle que la recherche d'information, l'inférence pour la compréhension automatique de textes, la désambiguïsation lexicale, la résolution d'anaphore et aussi l'inférence textuelle. Dans ce qui suit, nous allons définir les différentes ressources existantes et utilisées pour détecter l'inférence textuelle.

### 5.4.3.1) Le WordNet

WordNet (Miller, 1995) est une base de données lexicale développée depuis 1985 par des linguistes du laboratoire des sciences cognitives de l'université de Princeton. C'est un réseau sémantique de la langue anglaise, qui est fondé sur une théorie psychologique du langage. La première version diffusée remonte à juin 1991. Son but est de répertorier de classifier et de mettre en relation de diverses manières le contenu sémantique et lexical de la langue anglaise. Le système se présente sous la forme d'une base de données électronique (Chaumartin, 2007).

Le synset (ensemble de synonymes) est la composante atomique sur laquelle repose WordNet. Un synset correspond à un groupe de mots, dénotant un sens ou un usage particulier. Un synset est défini par les relations qu'il entretient avec les sens voisins. Les noms et verbes sont organisés en hiérarchies. Des relations d'hyperonymie et d'hyponymie relient les « ancêtres » des noms et des verbes avec leurs «spécialisations». Au niveau racine, ces hiérarchies sont organisées en types de base.

À l'instar d'un dictionnaire traditionnel, WordNet offre ainsi, pour chaque mot, une liste de synsets correspondant à toutes ses acceptions répertoriées. Mais les synsets ont également d'autres usages : ils peuvent représenter des concepts plus abstraits, de plus haut niveau que les mots et leurs sens, qu'on peut organiser sous forme d'ontologie. Nous pouvons ainsi interroger le système quant aux hyperonymes d'un mot particulier. À partir par exemple du sens le plus commun du nom "car" (correspondant au synset "1. car, auto..."), la relation d'hyperonymie définit un arbre de concepts de plus en plus généraux:

```
1. car, auto, automobile, machine, motorcar
    => motor vehicle, automotive vehicle
        => vehicle
            => conveyance, transport
                => instrumentality, instrumentation
                    => artifact, artefact
                        => object, physical object
                            => entity, something
```

---

[5] La réfutation est un procédé logique consistant à prouver la fausseté ou l'insuffisance d'une proposition ou d'un argument.

Dans cet exemple, il est clair que le dernier concept, "entity, something", est le plus général, le plus abstrait (il pourrait ainsi être le super-concept d'une multitude de concepts plus spécialisés).

Nous pouvons également interroger le système quant à la relation inverse de l'hypernymie, l'hyponymie. WordNet offre en fait une multitude d'autres ontologies, faisant usage de relations sémantiques plus spécialisées et restrictives. Nous pouvons ainsi interroger le système quant aux méronymes d'un mot ou d'un concept, les parties constitutives d'un objet ("HAS-PART"). Les méronymes associés au sens "car, auto..." du mot "car" sont :

```
1. car, auto, automobile, machine, motorcar
    HAS PART: accelerator, accelerator pedal, gas pedal, gas,
              throttle, gun
    HAS PART: air bag
    HAS PART: auto accessory
    HAS PART: automobile engine
    HAS PART: automobile horn, car horn, motor horn, horn
    (...)
```

### 5.4.3.2) Le FrameNet

FrameNet (Baker, Fillmore et Lowe, 1998), projet mené à Berkeley à l'initiative de Charles Fillmore, est fondé sur la sémantique des cadres (frame semantics). FrameNet a pour objectif de documenter la combinatoire syntaxique et sémantique pour chacun des sens d'une entrée lexicale à travers une annotation manuelle d'exemples choisis dans des corpus sur des critères de représentativité lexicographique. Les annotations sont ensuite synthétisées dans des tables, qui résument pour chaque mot les cadres avec leurs arguments syntaxiques.

### 5.4.3.3) Le Cyc

Cyc est un projet d'Intelligence Artificielle lancé en 1984 par Doug Lenat. Cyc vise à regrouper une ontologie et une base de données complètes sur le sens commun, pour permettre à des applications d'intelligence artificielle. D'effectuer des raisonnements similaires à ceux des humains. Des fragments de connaissances typiques sont par exemple : « les chats ont quatre pattes » ; « *Paris est la capitale de la France* ». Elles contiennent des termes (PARIS, FRANCE, CHAT?) et des assertions (« *Paris est la capitale de la France* ») qui relient ces termes entre eux. Grâce au moteur d'inférence fourni avec la base Cyc, il est possible d'obtenir une réponse à une question comme « Quelle est la capitale de la France ? » La base Cyc contient des millions d'assertions (faits et règles) rentrées à la main.

## 5.5) L'analyse des systèmes participant au RTE 2

Nous avons marqués pour chaque groupe de recherche participant au RTE2 les types d'inférences utilisés. Les résultats sont affiches dans le tableau 1.6.

| Type d'analyse / Groupes de recherches | lexicale | syntaxique | lexico-sémantique | Logique | numérique | Temporelle |
|---|---|---|---|---|---|---|
| **UNED** | + | + | + | | + | |
| **UMESS** | + | | | | | |
| **MITRE** | + | | | + | | |
| **IRST** | + | + | | | | |
| **GOGEX** | + | | + | + | | |
| **LCC'S** | + | | + | | | |
| **C&C** | + | + | | | | |

**Tableau 1.1 Représentation des différents types d'inférences entrepris par les groupes de recherches**

## 5.5.4) Quelques exemples d'inférence utilisés par des groupes de recherches

Dans le RTE 2 nous avons remarqué que tous les groupes de recherches n'ont pas utilisé d'inférence temporelle dans leurs systèmes et à l'heure actuelle, les résultats du RTE 3 ne sont pas encore publiés officiellement mais d'après notre lecture des différentes publications des groupes de recherches participant au RTE3, il y a deux groupes qui ont fait allusion à l'inférence temporelle. Pour cela, nous avons choisi de décrire leurs systèmes.

## 5.5.4.1) La reconnaissance de l'inférence textuelle basée sur l'analyse de dépendance et WordNet (Université nationale de l'éducation a distance de Madrid)

Le système présenté montre comment des informations sémantiques peuvent être extraites du texte en utilisant les structurations syntaxiques données par l'analyse de dépendance, et des ressources lexico- sémantiques comme Word Net peuvent développer le RTE.

Les techniques utilisées par ce système sont les suivantes :
- l'analyse dépendance du texte et de l'hypothèse.
- l'inférence lexicale entre les nœuds des arbres en utilisant Word Net.
- la concordance entre les arbres de dépendance basée sur la notion de l'inclusion.

### A) L'architecture du système

L'architecture du système est montrée dans la figure suivante (**Figure 1.7**) :

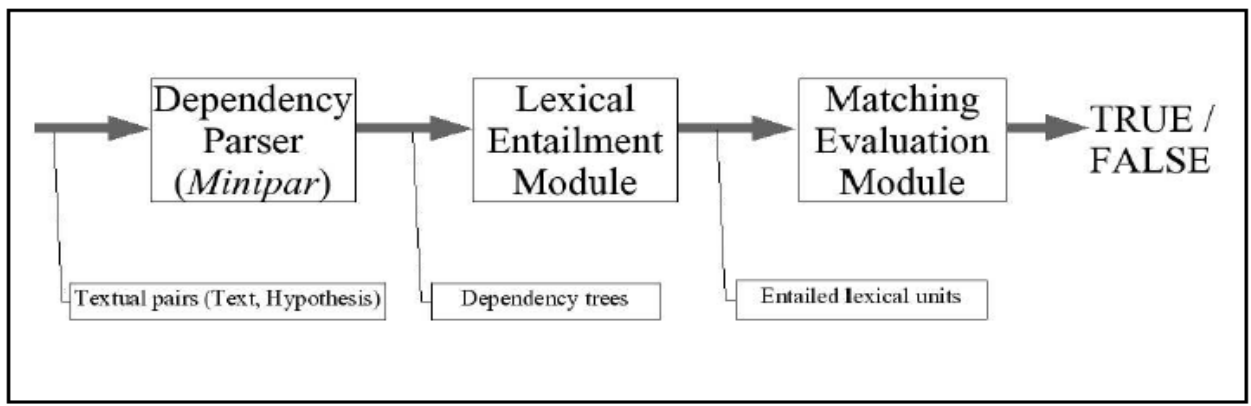

**Figure 1.7 : L'architecture du système**

Cette architecture est composée de Trois modules :

- **L'analyse de dépendance :** Elle consiste à normaliser les informations du dataset, de générer les dépendances existantes entre les mots et de donner à la sortie un arbre de dépendance constitué de nœuds qui représentent les mots de la phrase et d'arcs qui représentent les dépendances entre les nœuds. Ce travail est réalisé par un logiciel nommé « Lin's Minipar ».

- **L'analyse lexicale :** prend les informations données par l'analyse de dépendance et retourne les mots de l'hypothèse H qui sont infères du texte T. Ce module utilise WordNet pour détecter les relations de (synonymie, hyponymie, meronymie ) entre les unîtes lexicales.

- **Les relations entre les arbres de dépendance** : le but est de déduire si l'arbre de l'hypothèse est recouvert par l'arbre de dépendance du texte, Pour cela, la règle établie est qu'un arc est dit recouvert s'il est dans le même emplacement que dans l'arbre représentant le texte et il y a une inférence entre ces nœuds et celle du texte. La figure ci-dessous (figure 1.8) reprend ce genre de recouvrement.

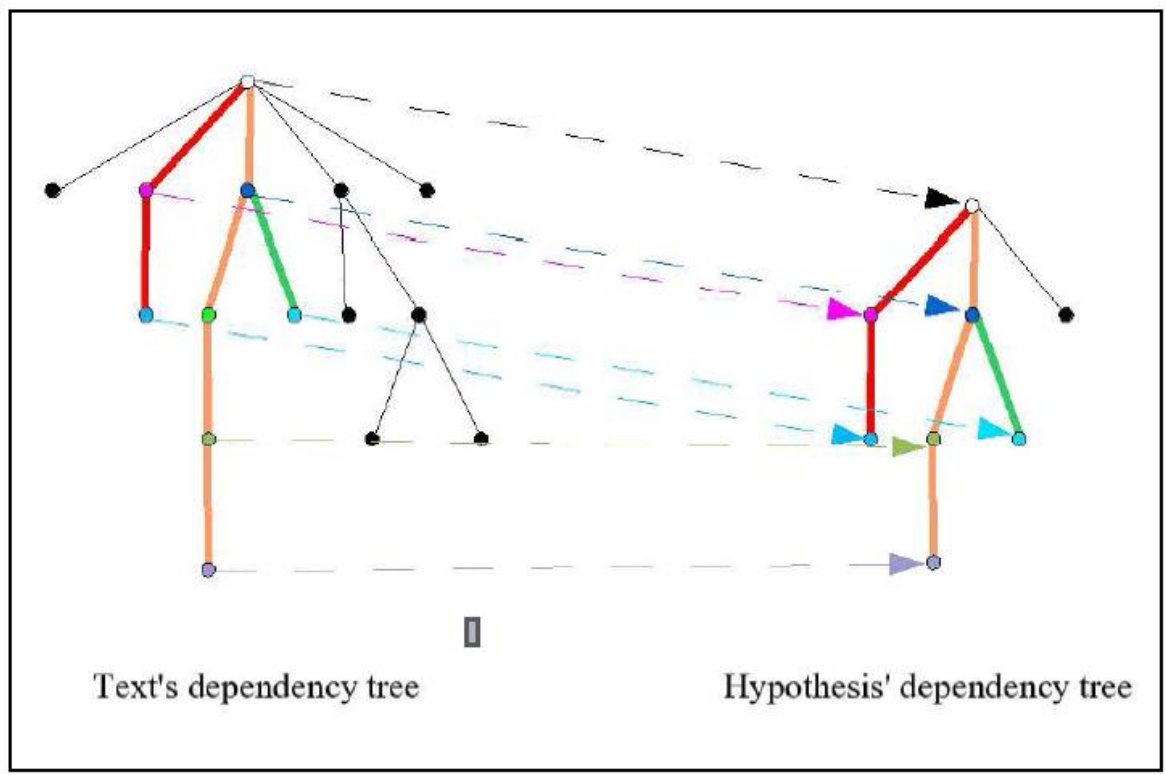

**Figure 1.8: Exemple de recouvrement entre arbre de dépendance**

### B) L'expérimentation du système

Le groupe a soumis deux systèmes au challenge.
- **Système 1**

Le systeme1 n'utilise que les deux premiers modules, et la décision de l'existence d'inférence est prise par rapport au nombre de nœuds de l'hypothèse infère de l'arbre de dépendance du texte.
- **Système 2**

Le système 2 utilise les 3 modules et la décision est prise par rapport au nombre d'arc recouverts.

Les résultats sont affiches dans le **tableau 1.2**. L'utilisation de WordNet seule a donné de bons résultats, mais en ajoutant le module de recouvrement il décroît les performances du système.

| Les systèmes | Précision |
|---|---|
| Système 1 : | 56,37 % |
| Système 2 : | 54,75 % |

**Tableau 1.2: Les valeurs de précision des systèmes**

La notion de recouvrement n'est pas appropriée pour le RTE, car un large recouvrement n'implique pas une inférence sémantique, et un faible recouvrement n'implique pas une différence sémantique. L'utilisation de Word Net a contribué à l'inférence au niveau lexical et a augmenté les performances du système. Dans cette direction, les prochaines étapes seront de reconnaître et d'évaluer les inférences entre les expressions numériques, **les entités nommées** [6] et les expressions temporelles.

### C) L'évolution du système

Ce qui a été développé pour le RTE2 est un module pour la détection des expressions numériques, ce qui a permis d'augmenter fortement **la précision** (harrera et al.,2006). La figure suivante montre comment le module est introduit dans leur système.

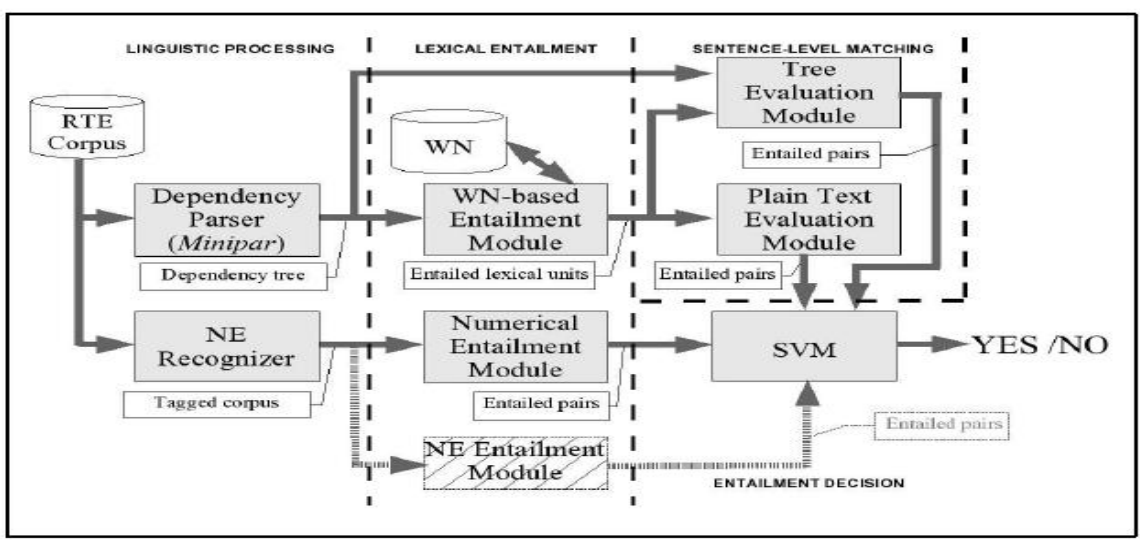

**Figure 1.9: Architecture du système UNED**

Dans le RTE 3, le groupe s'est focalisé sur l'inférence entre les entités nommées. Il a défini les relations d'inférences entre les entités nommées (Rodrigo et al., 2007). Exemple :

- Nom propre E1 infère nom propre E2 si une chaîne E1 contient la chaîne E2.
- une expression du temps t1 infère une expression du temps T2 si l'intervalle de temps exprimée dans t1 est inclus dans l'intervalle T2.

Ce module de d'inférence a lui aussi contribué à augmenter la précision (Rodrigo et al, 2007).

## 5.5.4.2) COGEX (université du Texas, USA)

Le système utilise une approche logique pour résoudre l'inférence textuelle. En d'autres termes, l'inférence textuelle est considérée comme un problème d'implication logique entre les sens des deux phrases (Tatu et al., 2006).
La description du système et l'évolution qui s'est produite dans chaque challenge est décrite dans ce qui suit.

---

[6] Les entités nommées désignent l'ensemble des noms de personnes, de lieux, d'entreprise contenues dans un texte.

## A) La description du système

La première étape consiste à transformer le texte et l'hypothèse en forme logique (Moldovan and Rus, 2001).

Pour cela il faut d'abord transformer du langage nature a un format prédicat argument, pour cella le groupe utilise WordNet pour lier le prédicat avec ses argument. Concrètement WordNet produit des relations entre les synsets, et chaque synset lui correspond un prédicat.

Le prédicat peut avoir un ou plusieurs arguments et le prédicat qui correspond au nom a un seul argument en général, et le prédicat qui correspond à un verbe a trois arguments : l'événement, le sujet et le complément d'objet.

Pour chaque relation dans la chaîne lexicale[7], le système génère un axiome utilisant les prédicats qui correspondent au synset de la relation.

Par exemple : il y a une relation d'inférence entre le verbe **vendre** et le verbe **payer.**
Le système génère l'axiome suivant pour cette relation :
Vendre_VB_1(e1,x1,x2) ➔ payer_VB_1(e1,x1,x3)
Ce type d'axiome contribue à l'inférence quand une chaîne lexicale est trouvée.

Apres la transformation des deux paires de texte en format logique le groupe utilise la preuve par « l'absurde » ou ''preuve par contradiction'' (Wos, 1998). La négation de l'hypothèse H est réalisée s'il y a une contradiction ou une déduction de contradiction par rapport au texte T, nous concluons que l'hyponyme est dérivable du texte.

## B) L'évolution du système

Il a été développé pour le RTE 2 un module qui traite la négation dans la transformation du texte en prédicat et un autre module qui fait une analyse sémantique en tant que pré traitement pour donner les relations existantes entre le verbe et ses arguments et aussi entre les arguments eux-mêmes (Tatu et al.,2006).

Pour le RTE3 le groupe a développé et intégrer a leur système plusieurs outils.
Dans ce qui suit nous allons présenter l'architecture du système et les nouveaux outils conçus et utilises pour améliorer l'inférence.
Le schéma du dernier système conçu pour le RTE 3 par le groupe est donné par la figure ci-dessous.

---

[7] Une chaîne lexicale est une chaîne où il y a une relation entre deux synsets.

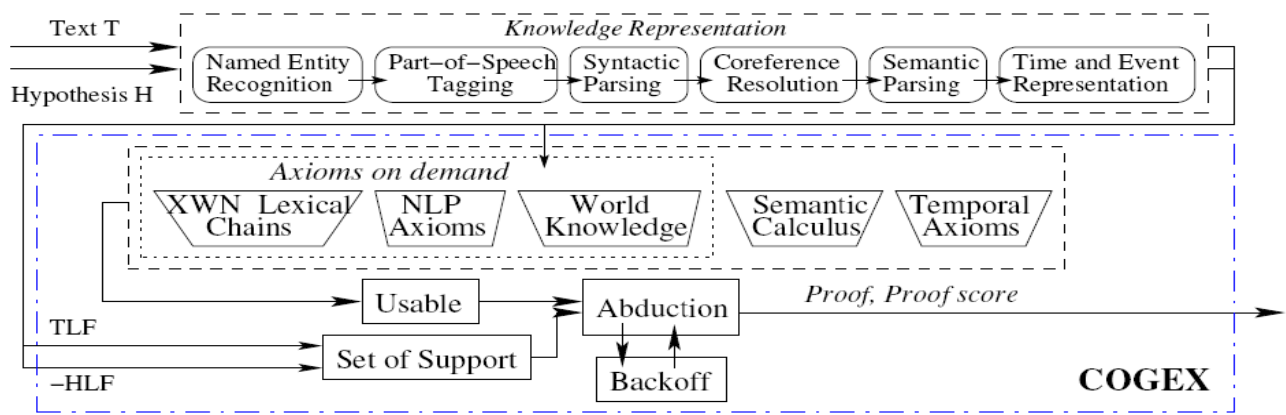

**Figure 1.10 : Architecture du système**

- **EXtended WordNet**

XWN (eXtended WordNet) est un projet qui a pour but d'enrichir les relations du dictionnaire WordNet avec des relations sémantique entre les synsets et les transforment en format logique (Tatu et Moldovan, 2007).

- **TARSQI**

C'est un système modulaire pour l'annotation automatique temporelle qui ajoute les expressions du temps, des événements et des relations temporelles de l'actualité des textes (Venhaguane et al. ,2005).

- **Outil pour la gestion des coréférences**

Pour relier les phrases dans les textes longs et, résoudre le problème qui est apporté par les coréférences dans l'inférence textuelle, l'outil développé combine l'algorithme Hobbs (Hobbs, 1978) et l'algorithme de résolution d'anaphore (Lappin and Leass, 1994).
Pour le RTE, il est important d'avoir les relations entre les prédicats d'un long texte.

Exemple 1 : *George Bush grandit à Greenwich au Connecticut, Il est à l'époque membre d'une confrérie étudiante secrète devenue célèbre.*
*Lier George Bush* et *il,* est une des taches que l'outil doit résoudre.

Le développement du XWN-KB a eu un impact considérable sur le RTE, mais l'utilisation du TARSQI n'a donné aucun impact sur le résultat car l'utilisation des expressions temporelles dans ce corpus est inexistante.

## 5.5.5) Conclusion

Dans les travaux entamés par UNED sur les entités nommées, le groupe a établi plusieurs règles d'inférence entre les entités nommées, parmi lesquelles se trouve une règle d'inférence entre les expressions temporelles. Celle-ci peut être considérer comme une contribution implicite à l'inférence temporelle. Mais concrètement l'inférence temporelle est considérée comme une perspective pour leurs prochaines recherches.

## 5.6) Conclusion

Dans ce chapitre nous avons explorés l'apport du RTE dans les différentes applications du TALN (RI, QR, EI et RA) et nous avons exploré les différentes approches utilisées pour détecter l'inférence (lexical, lexico syntaxique, sémantique et logique). Aussi nous avons analysé les approches des différents groupes de recherches qui ont participe au challenge Pascal RTE. Cette étape nous a permis de découvrir les chemins qui n'ont pas encore été pris pour détecter l'inférence textuelle.

Enfin nous nous sommes focalisés à décrire les systèmes qui ont mentionné l'aspect temporel dans leurs recherches. Nous avons remarqué que dans les trois RTE qui se sont déroulés, l'inférence temporelle est une perspective qui n'est pas encore entamée. Nous allons justement décrire dans le prochain chapitre l'aspect temporel dans le RTE.

# - Chapitre 2 -

# Le temps et la langue

# Chapitre 2

# Le temps et la langue

## 1) Introduction

Avant d'entamer notre travail sur l'inférence textuelle, nous avons besoin d'explorer la notion du temps dans ses différentes bannières, d'abord par rapport à la logique modale et aussi par rapport à la langue.

Le mot « temps » recouvre plusieurs significations en français, et il est nécessaire pour la compréhension de distinguer le temps grammatical du temps notionnel. Le second est représenté en logique par une ligne droite et infinie, avec un point marquant le présent et séparant le passé du futur. Le temps grammatical désigne les marques linguistiques utilisées pour exprimer le temps notionnel dans le langage (l'imparfait, le présent de l'indicatif, etc.…).

Dans ce qui suit nous allons explorer ces deux notions du temps, du point de vu logique avec ces différentes représentations (structure de points, structure d'intervalles, événement et Allen) ensuite au point de vu langage.
Nous terminons avec une étude sur l'inférence temporelle élaborée par l'un des groupes les plus abouti dans le domaine.

## 2) La structure de points

La conception du temps est couramment reliée à la notion de point ou d'instant sur un axe temporel. Les points permettent en effet d'utiliser les structures de nombres (entiers, rationnels ou réels). Cette conception est largement utilisée dans la modélisation de phénomènes évoluant dans le temps.
Cette structure temporelle doit être manipulée avec un langage logique ; la logique du temps, historiquement très liée au développement des logiques modales. Elle est basée sur les connecteurs logiques habituels ($\wedge, \vee, \neg, \Rightarrow, \Leftrightarrow$) et les opérateurs temporels P (passé) et F (futur). Ainsi, si l'action de chanter effectuée par John est notée p, on aura les représentations suivantes :
- John chante : p
- John chanta : Pp
- John chantera : Fp
- John avait chanté : PPp (on se place dans le passé d'un point situé au passé lui-même)
- John aura chanté : FPp

Ces formules seront enrichies avec de nouveaux opérateurs similaires à ceux utilisés en logique modale (Bras, 1990).
Toutes les logiques dérivées de la logique du temps sont basées sur une ontologie de points. Nous allons maintenant nous intéresser à des ontologies d'intervalles.

# 3) La structure d'intervalles

Du point de vue philosophique, il semble que le concept de point dépourvu de durée ne correspond pas à la réalité :
Du point de vue linguistique, il est encore plus évident qu'une entité ponctuelle est mal adaptée pour l'expression de la référence temporelle. Même les expressions dites ponctuelles se réfèrent à des périodes étendus, comme dans les exemples suivants :
*A six heures précises, Harry quitta son bureau.*
Une structure d'intervalle est définie par $< I, <, \subseteq >$, avec *I* un ensemble non vide d'entités temporelles, des relations de précédence ($<$) et d'inclusion ($\subseteq$). Voici quelques propriétés de cette structure :

$\subseteq$ est un ordre partiel, elle est en effet :
Réflexive : $(\forall x \in I) \ x \subseteq x$.
Antisymétrique : $(\forall x \in I)(\forall y \in I)(x \subseteq y \land y \subseteq x \Rightarrow x=y)$.
Transitive : $(\forall x \in I)(\forall x \in I)(x \subseteq y \land y \subseteq z \Rightarrow x \subseteq z)$.

Nous pouvons également remplacer la relation (par la relation O (overlap) qui exprime que deux événements ont une partie commune, et définie par rapport à l'inclusion:
$xOy \Leftrightarrow (\exists z)(z \subseteq x \land z \subseteq y)$

La mise en place des logiques temporelles basées sur les sémantiques d'intervalles amènent à des résultats relativement complexes, qu'il n'est pas nécessaire d'exposer ici.
Des critiques ont été adressées aux sémantiques d'intervalles, notamment en ce qui concerne la difficulté de définir la vérité d'une proposition (vraie sur toutes intervalles ? sur au moins l'un deux ?). Ces problèmes ont provoqué la nécessité de concevoir une entité plus globale et plus complète.

# 4) La structure d'événements

L'événement est une nouvelle entité primitive, de durée non nulle et fini, correspondant intuitivement à des fragments de notre perception du monde. Pour les linguistes comme pour les philosophes, les logiciens et les spécialistes de l'intelligence artificielle, la tendance est de préférer les événements aux intervalles car les événements ont une structure à portée non seulement temporelle, mais aussi spatiale.
Davidson a proposé de traiter les événements comme des objets, ajoutant à l'ensemble des individus d'un modèle, un ensemble d'événements, par exemple, la phrase *Marie aime Paul* n'est plus représentée par *aimer (Paul, Marie)*, mais par :
*Aimer(e, Paul, Marie)*

Une structure d'événement est définie par Kamp par le triplet $<E, \alpha, O>$, où E est un ensemble d'entités de base non nulle, $\alpha$ est la relation de précédence, et O la relation de recouvrement si e1Oe2 alors une partie de e1 au moins a eu lieu en même temps que e2.
$\alpha$ est asymétrique : $(e1 \ \alpha \ e2) \Rightarrow \neg (e2 \ \alpha \ e1)$
$\alpha$ est transitive : $(e1 \ \alpha \ e2) \land (e2 \ \alpha \ e3) \Rightarrow (e1 \ \alpha \ e3)$
O est symétrique : $(e1 \ O \ e2) \Rightarrow (e2 \ O \ e1)$
O est réflexive : $(e1 \ O \ e2)$
Principe de séparation : $(e1 \ \alpha \ e2) \Rightarrow \neg (e1 \ O \ e2)$

Transitive mixte : (e1 α e2)∧ (e2 O e3) ∧ (e3 α e4) ⇒ (e1 α e4)
Principe de linéarité : (e1 α e2) (e1 O e2) (e2 α e1)
Ces conditions minimales sont dictées par l'intuition lorsque nous avons des événements et des relations qui les lient.

Nous avons présenté trois ontologies (structures de points, structures d'intervalles et structures d'événements). Il est fondamental de séparer le niveau temporel (points et intervalles) du niveau relatif à l'expérience du monde (événements). En effet, si les relations définies dans les structures d'événements sont des relations temporelles, les événements sont également des expériences, des « faits » qui ont lieu et qui déterminent la structure du temps. C'est pourquoi nous pouvons dire que la logique d'Allen, que nous allons présenter, permet de rattacher les deux notions.

# 5) La théorie d'Allen

Selon Allen, deux intervalles peuvent être liés entre eux par les 13 relations primitives suivantes (Bras, 1990). Où X et Y sont des termes de types intervalles de temps (on appelle « relation inverse » la relation correspondante entre Y et X) :

| Relation | Symbole | Symbole relation inverse |
|---|---|---|
| X beforeY | < | > |
| X equalsY | = | = |
| X meetsY | M | Mi |
| X overlapsY | O | Oi |
| X duringY | D | Di |
| X startsY | S | Si |
| X finishes Y | F | Fi |

**Tableau 2.1: Les relations d'Allen**

Les relations sont mutuellement exclusives : une seule relation est possible entre deux intervalles.

Il est possible de composer les relations. Ainsi, la transitivité des relations entre intervalles est définie par:
∀ I ∀ j∀ k (( I < j) ∧(j < k) ⇒I < k)

$\forall$ I $\forall$ j$\forall$ k ((I mj) $\wedge$(j d k)$\Rightarrow$ (I o k) $\vee$ (I d k) $\vee$ (I s k))

Il existe 169 relations de transitivité de ce type.

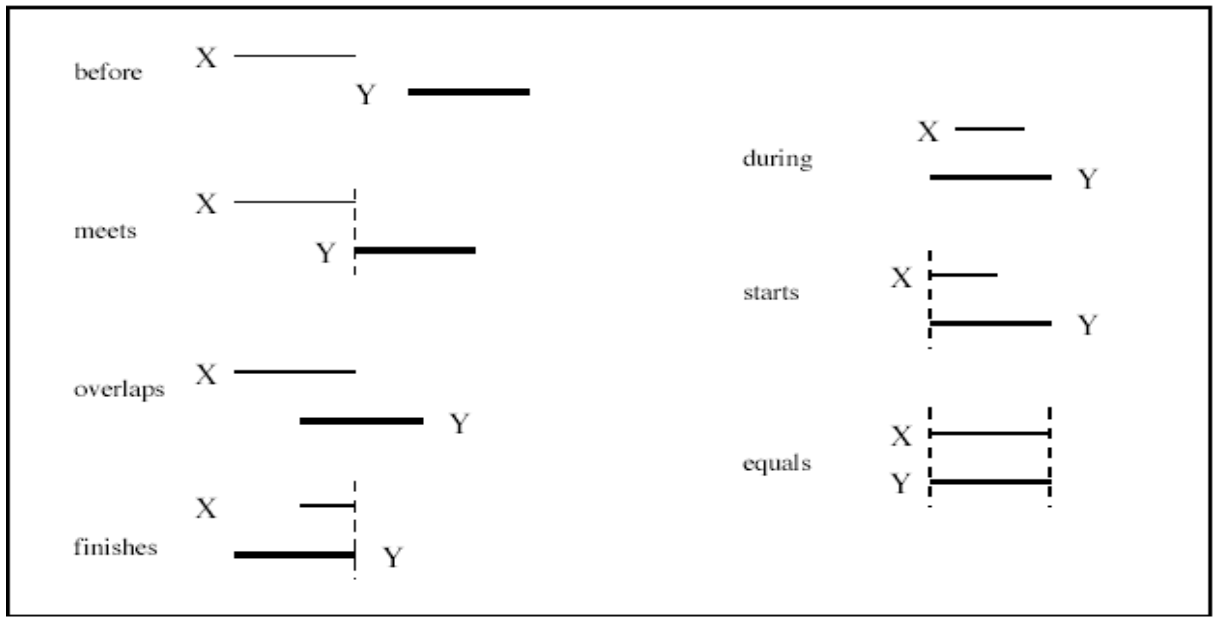

**Figure 2.1: Représentation des relations d'Allen**

Deux intervalles peuvent être reliés par une relation primitive, mais aussi par une relation complexe ; il est ainsi possible de représenter une connaissance incomplète des relations.
La connaissance temporelle sur un ensemble d'intervalles peut être représentée par un réseau de contraintes. Il s'agit d'un graphe orienté dont les nœuds représentent les intervalles et dont les arcs sont étiquetés par la relation entre les intervalles.

L'exemple suivant, très simple, permet d'illustrer rapidement le raisonnement sur les intervalles :
*Paul entra dans la pièce* (1). *Marie regardait la télévision* (2) .*Elle l'éteint* (3).
(1) Et (2) introduisent l'assertion temporelle suivante :
I *entre* during I *regarde_television*
Puis, en examinant (2) et (3), nous obtenons :
 I *regarde_television* meet I *enteindre- television*

# 5) Le temps dans la langue

Qu'est-ce qui distingue le temps linguistique des autres notions de temps? "Ce que le temps linguistique a de particulier c'est qu'il est organiquement lié à l'exercice de la parole, qu'il se définit et s'ordonne comme fonction du discours. Ce temps a son centre – un centre, à la fois, générateur et axial - dans le présent de l'instance de la parole" (Benveniste, 1974). Le discours instaure un *maintenant*, moment de l'énonciation. En opposition au *maintenant*, nous créons un alors. Ce *maintenant* est donc le fondement des oppositions de la langue.

## 5.1) Le modèle de Reichenbach

Reichenbach a proposé, pour modéliser la sémantique des temps grammaticaux, les trois repères suivants :
E le moment de l'événement
S le moment de l'énonciation ou de la parole (Speech Time)
R le moment de référence
Les relations possibles entre repères sont la relation de simultanéité) notée « , » et la relation de précédence notée « _ ».  La nouveauté réside surtout dans l'ajout d'un moment de référence, qui permet de prendre en considération certains temps composés. Ainsi, la représentation de quelques temps grammaticaux à l'aide du modèle de Reichenbach est la suivante :
Passé simple       je vis Paul           E, R_S
Plus-que-parfait   j'avais vu Paul       E_R_S
Futur              je verrai Paul        S_E,R
Futur antérieur    j'aurai vu Paul       S_E_R

## 5.2) Les adverbiaux temporels

Un adverbial est un élément (mot ou groupe de mots) ayant une fonction similaire à celle d'un adverbe ou d'un complément circonstanciel, c'est-à-dire qu'il modifie le verbe auquel il est rattaché (Charolles, 1997). Nous pouvons le supprimer sans rendre la syntaxe ni la sémantique de la phrase incorrecte. Ainsi, les passages soulignés des exemples suivants ont une fonction adverbiale temporelle :
*Paul arrive demain.*
*Marie est revenue à cinq heures.*
Nous pouvons distinguer :
Les adverbiaux de référence temporelle dont le rôle est d'exprimer la localisation d'un événement dans le temps : *demain.*
Les adverbiaux de durée : *pendant une heure, en trois jours*
Les adverbiaux de durée : *pendent une heure.*
Les adverbiaux  de fréquence : *souvent, tous les mois.*
Les adverbiaux itératifs : *trois fois, plusieurs fois.*
Les adverbiaux de quantification : *toujours, quelquefois.*
Les adverbiaux présuppositionnels : *encore, déjà.*

## 6) L'inférence temporelle

Si l'annotation des marqueurs du temps dans le discours sont l'objet de plusieurs sujets de recherches, l'étude de l'inférence temporelle et ses applications ne sont qu'à ses débuts. Ce problème commence à générer des travaux en informatique linguistique liés aux enjeux que représentent les informations temporelles entre autre pour la recherche d'information et les systèmes questions-réponses.

Avec l'exploration de ce nouveau champ d'action  dans le traitement du langage naturel, l'inférence temporelle nous permet d'établir des relations temporelles existantes  entre évènements dans un texte, de détecter les relations existantes entre expressions temporelles  et aussi les relations entre  expressions temporelles et événements.

Dans ce qui suit nous présentons l'une des recherches les plus abouti dans le domaine.

## 6.1) Le travail du groupe Human Language Technology Research Institut (HLTRI) sur l'inférence temporelle :

HLTRI est un groupe de recherche travaillant sur l'inférence temporelle. Il est aussi membre de l'organisation fondatrice du langage (TimeML[8]) qui est un langage de spécification d'événements et d'expressions temporelles dans le langage naturel.
Afin d'étudier l'inférence temporelle dans le langage naturel, le groupe (HLTRI) a établi un grand corpus de questions-réponses qui sont fondées sur la recherche d'information temporelle.
Les questions sont annotées comme ceci:
• Expressions temporelles, annotées par la balise TIMEX3.
• La balise EVENT correspond à un événement.
• LIEN est une balise qui code les relations entre éléments temporels.

Pour découvrir les relations temporelles entre les événements dans un texte, le groupe a utilisé la représentation graphique.
Les nœuds du graphe sont représentés par les événements et les arcs entre les nœuds sont soit des relations TLink, SLink ou ALink.
Pour classer les événements dans un même texte, il utilise les trois relations ALINK, TLINK, SLINK et entre les évènements de deux textes différents il n'utilise que le module TLINK.
- TLink, représentant les relations temporelles entre les événements ou entre un événement et une expression temporelle.
- SLink ou relation de subordination, est utilisée pour introduire des contextes et des relations entre deux événements.
- ALink ou relation aspectuelle, représentant la relation aspectuelle entre un événement et son argument (en général c'est un autre événement).
Afin d'avoir toutes les relations temporelles possibles entre les évènements des deux textes, le groupe a conçu un module d'inférence temporelle.
À partir des différents liens TLINK, ALINK et SLINK existant entre les expressions temporelles et les événements, le module infère de nouveaux liens non détectés auparavant. Pour cela le groupe a définit plusieurs règles d'inférences qui sont citées ci-dessous :

---

[8] TimeML a été développé dans le cadre de trois ateliers AQUAINT et des projets. En 2002, TERQAS atelier vise à renforcer la langue naturelle de répondre à la question des systèmes de réponse temps-fondé des questions sur les événements et les entités dans des articles de journaux. La première version de TimeML a été définie et la TimeBank corpus a été créé comme une illustration. Tango a un atelier de suivi dans lequel un outil graphique d'annotation a été développé. Actuellement, le TARSQI projet développe des algorithmes qui balise les événements et le temps des expressions NL textes dans le temps et l'ancrage et l'ordre des événements. En outre, TimeML a été examinée et encouragée dans: ARTE atelier ACL : Annoter et Raisonnement sur le temps et les événements (Juillet 2006), Séminaire Dagstuhl Annoter, l'extraction et le raisonnement sur le temps et les événements (avril, 2005).

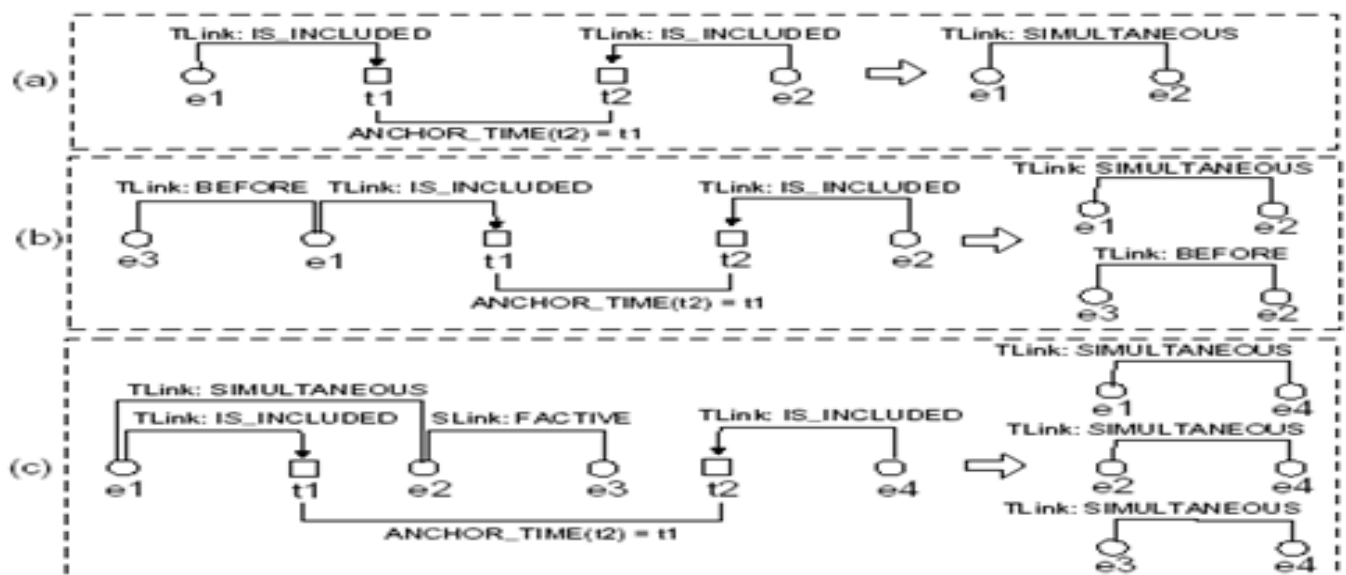

**Figure 2.2: Règles de l'inférences temporelles**

Concrètement le module suit les étapes suivantes :
Etape1:trouver T1 et T2, deux expressions temporelles dans la phrase ou des phrases adjacentes.
Etape2:Rechercher des événements E1 et E2 lié à T1 et T2 respectivement.
Etape3:Trouver une relation CE1 lien entre E1 et d'autres événements.
Etape4:Trouver relation CE2 lien entre E2 et d'autres événements.
Etape5:Utiliser une inférence temporelle reliant CE2 et CE1.

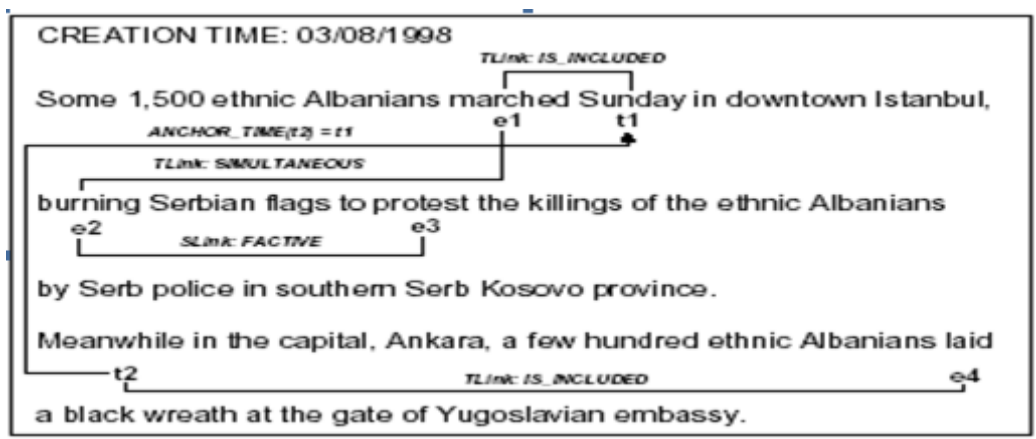

**Figure 2.3: Application des règles de l'inférences sur un exemple du corpus RTE**

Lors de l'application de la procédure à l'exemple illustré dans la figure 2.3 le module suit les étapes suivantes :
Etape 1: trouver les expressions temporelles t1 et t2.
Etape 2: événements e1 et e4 reliées à t1 et t2 avec TLink : is_includes.
Etape 3: détecter la chaîne des événements (e1, e2, e3).
Etape 5: T1 et T2 sont liés (par un ANCHORTIME (t2)=t1), ce qui signifie que e1, e2, e4 et e3 sont simultanés.

## 6.2) Synthése

Ce groupe s'est focalisé sur la déduction de nouvelles relations entre évenements. Cette étude sur l'inférence temporelle a permis d'établir plusieurs régles d'inférences reliant les évenements et les expressions temporelles.

## 7) Conclusion

Dans ce chapitre nous avons exploré la logique temporelle et ses applications dans le traitement du langage Naturel. Nous avons aussi illustré avec les travaux du groupe (HLTRI) les différents types d'inférences temporelles existants, Nous avons remarqué que ce groupe ne se base que sur l'amélioration des détections des relations temporelles existantes entre évenements et expressions temporelles.
Nous nous sommes inspirés de ces travaux dans notre façon de procéder pour élaborer le corpus et concevoir nos inférences.
Nous allons montrer dans les prochains chapitres comment nous avons concrétisé cet objectif.

# Partie 2

# Conception, réalisation et mise en œuvre du système TIMINF

## Résumé :

Cette partie de notre mémoire est composée de trois chapitres qui regroupent la conception et la réalisation de notre projet.

Dans le chapitre trois, nous avons entrepris une démarche expérimentale à base de corpus afin de dégager les différentes classes d'inférence temporelle et à partir de cette analyse, nous avons conçu l'architecture du système de reconnaissance d'inférence textuelle TIMINF présenté dans le chapitre quatre.

Nous nous sommes intéressés dans le dernier chapitre à l'évaluation des sorties de notre système en le confrontant à un corpus de test adapté.

- **Chapitre 3 -**

# L'élaboration et étude du corpus

# Chapitre 3

# L'élaboration et l'étude du corpus

## 1) Introduction

L'importance du RTE dans le TALN a poussé les chercheurs à s'investir dans ce domaine et à explorer différents chemins pour parvenir à détecter et à classifier différents types d'inférences.

Dans les chapitres précedents, nous avons d'abord étudié les groupes travaillant sur la reconnaissance de l'inférence textuelle et nous avons remarqué qu'aucun groupe n'utilisait l'inférence temporelle dans son système. Dans le chapitre précédent nous avons étudié le temps dans la langue et nous avons remarqué que les groupes travaillant sur l'inférence temporelle se base sur l'amélioration des détéctions des relations temporelles existantes entre évènements et expressions temporelles mais ils n'essayaient en aucun cas d'intégrer leurs travaux a un systéme d'inférence textuelle.

Afin de répondre au manque de l'inférence temporelle dans le RTE, notre objectif est d'intégrer le système de détéction d'inférence temporelle dans un systéme d'inférence textuelle. Pour cela, nous avons l'obligation d'étudier les relations temporelles qui peuvent exister entre deux ségments de textes à travers un corpus que nous avons élaboré. Ceci nous a permis de distinguer différents types d'inférences.
Nous allons montrer tout au long de ce chapitre comment nous avons concrétisé ces différents objectifs.

## 2) L'élaboration du corpus

La première étape à entreprendre consiste à créer le corpus constitué de paires de textes et hypothèses (T-H) qui correspond à des informations collectées à travers le web dans des domaines différents. Nous avons choisi d'établir notre corpus en langue anglaise car jusqu'à nos jours les recherches les plus abouties sur l'inférence temporelle et aussi sur le RTE sont en langue anglaise.

Pour cela, nous avons choisi d'utiliser le corpus de questions élaborées pour le test par la compagne d'évaluation des systèmes de recherches d'informations (clef[9]) pour l'année 2006.

---

[9]Le lien du challenge clef : http://www.elda.org/article225.html

Le challenge CLEF est crée en 2000 pour fournir une infrastructure visant à soutenir le développement, d'essai et d'évaluation des systèmes de cross-langue de recherche d'information dans plusieurs langues européennes (Français, Italien, Allemand).

Pour pouvoir développer et évaluer notre système, nous avons sélectionné des questions portant sur des événements temporels et nous avons soumis ces questions au système de question-réponse answerbus [10] disponible sur le web. Nous avons récupéré les réponses correspondantes et nous les avons modifiées pour obtenir l'inférence souhaitée. Nous avons aussi transformé les questions à l'affirmatif.

Nous illustrons ces démarches par l'exemple montré ci-dessous :

La question numéro 13 du corpus de test de challenge clef 2006:

*In what year did the catastrophe in Chernobyl happen?*

La requête va être mise dans le système de question réponse Answerbus. Le résultat est montré ci-dessous :

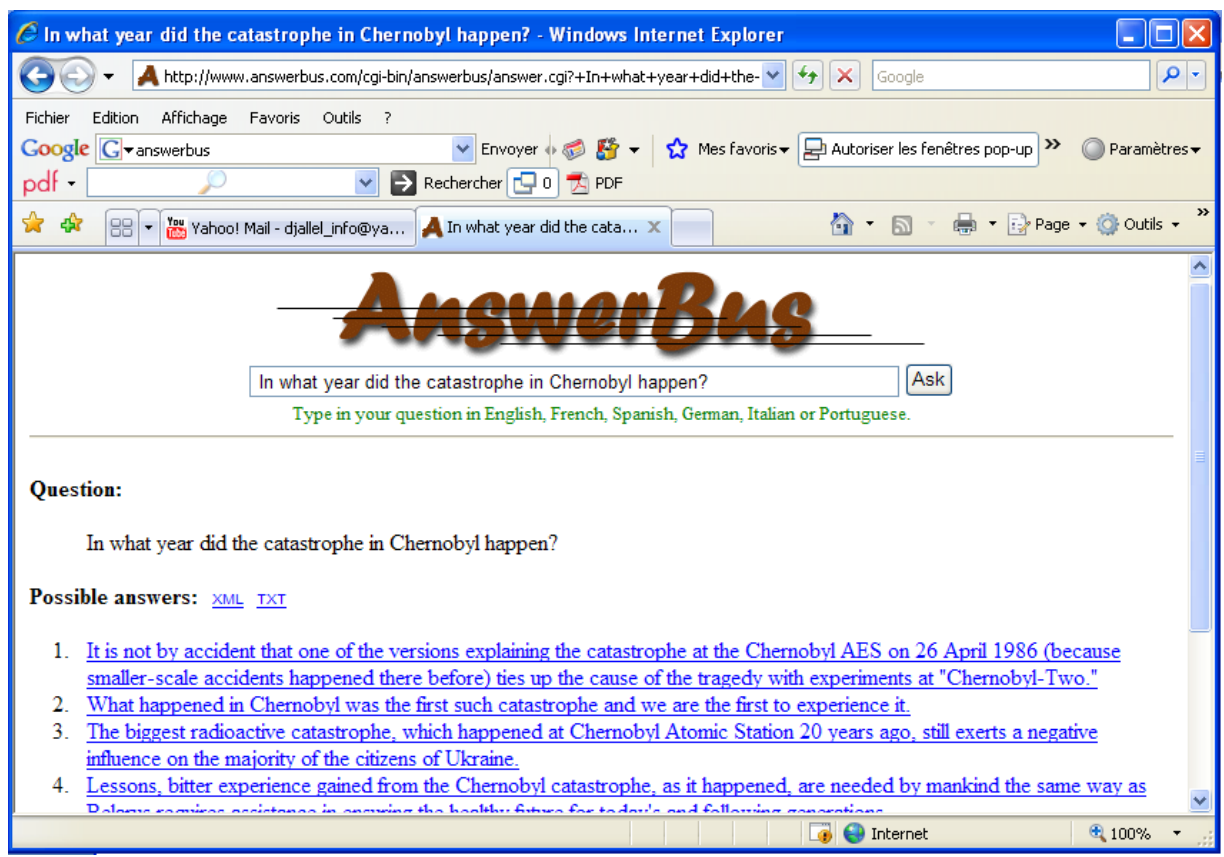

**Figure 3.1 : Représente la réponse du système AnswerBus**

Nous choisissons la première réponse donnée par le système qui est :

---

[10] http://www.answerbus.com

*H: It is not by accident that one of the versions explaining the catastrophe at the Chernobyl AES on 26 April 1986 (because smaller-scale accidents happened there before) ties up the cause of the tragedy with experiments at "Chernobyl-Two" .*

Aussi, nous transformons la question en affirmatif en répondant à la question. Comme résultat nous avons la réponse suivante :

*T: the catastrophe of Chernobyl happens in 1987.*

Finalement nous avons une paire de texte de la forme :

*T: the catastrophe of Chernobyl happens in 1987.*
*H: It is not by accident that one of the versions explaining the catastrophe at the Chernobyl AES on 26 April 1986 (because smaller-scale accidents happened there before) ties up the cause of the tragedy with experiments at "Chernobyl-Two"  .*

Comme dans le challenge RTE, les exemples sont divisés en deux types de corpus (corpus de développement et corpus de test).

Les deux corpus sont constitués de 30 paires de textes et chaque portion du corpus doit inclure 50% d'exemples avec une inférence vrai 50% d'exemples avec une inférence fausse. Pour cela, chaque exemple (T-H) paire est jugé par un annotateur pour voir s'il y a une inférence textuelle dans la paire de texte entre (T-H) ou pas.

La figure suivante montre un exemple du corpus après annotation :

```
<pair id="754" value="TRUE" >

    <t> the catastrophe of Chernobyl happens in 1987</t>

    <h> It is not by accident that one of the versions explaining the catastrophe at the
    Chernobyl AES on 26 April 1986 (because smaller-scale accidents happened there
    before) ties up the cause of the tragedy with experiments at "Chernobyl-Two" . </h>

</pair>

Id : représente le numéro de la pair.

Value : représente la décision de l'annotateur (vrai ou faux).
```

**Figure 3.2 : Exemple du corpus annoté**

L'exemple est évalué par un second juge qui évalue les paires de textes et d'hypothèses, sans avoir pris conscience de leurs contextes.

Les annotateurs étaient d'accord avec le jugement dans 86,66 % des exemples, ce qui correspond à 0.6 Kappa qui est une mesure statistique pour calculer a quel point deux personnes A et B sont d'accord pour classer N éléments dans K catégories mutuellement exclusives, les 13,33% du corpus où il n'y a pas eu d'accord ont été supprimés. Le reste du corpus est considéré comme un «gold standard» ou « BASELINE » pour l'évaluation.

## 3) Classification de l'inférence temporelle

Apres avoir conçu notre corpus, nous avons annoté manuellement les événements, les dates et les différents types d'inférences (lexicales, syntaxiques et temporelles) existant entre les segments de textes. Cela nous a permis de détecter les différents types d'inférences temporelles entre les segments de textes.
Nous détaillons dans ce qui suit les différentes classes que nous avons distingué:

## 3.1) Les inférences entre expressions temporelles

L'inférence permet d'établir des relations temporelles liant date, heure et durée entre elles. Dans le même contexte, nous avons distingué trois types d'inférences temporelles liant des expressions temporelles.

Cette figure représente le nombre de paires de textes pour chaque sous classe d'inférence dans notre corpus de développement.

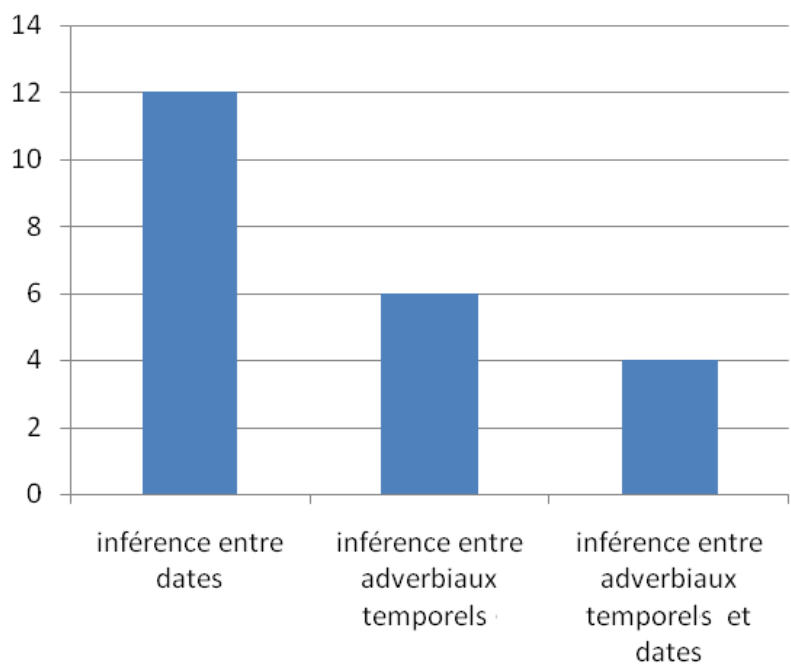

**Figure 3.3 : Pourcentage de paires par types d'inférences**

Dans ce qui suit, nous présentons les trois types d'inférences :

## 3.1.1) Les inférences entre dates

C'est la relation temporelle entre qui peut y avoir entre les dates du texte T et les dates du texte H.

L'exemple suivant permet de montrer la relation qui peut exister entre les dates.

Exemple 1:

<pair id="8" value="TRUE" >
T: the football world cup finished on t1: **july 12 th 2006**.
H: the football world cup finished in t2: **july 2006**.

Dans cet exemple, nous remarquons que l'inclusion entre les deux dates t1 et t2 a permis d'avoir l'inférence temporelle.

Exemple 2:

1) <pair id="1" value="TRUE" >
T: the second world war finished in t1: **1945**.
H: the end of the second world war took part t2: **between 1940 and 1950.**

Dans cet exemple nous remarquons aussi que l'inclusion entre les deux dates t1 et t2 a permis d'avoir l'inférence temporelle.

## 3.1.2) Les inférences entre adverbiaux temporels

L'inférence permet d'établir une relation temporelle entre adverbiaux de référence temporelle qui exprime la localisation d'un événement dans le temps.
L'exemple suivant permet de montrer la relation qui peut exister entre deux adverbiaux temporels.

Exemple 1:

<pair id="15" value="TRUE" >
T: he has worked **during 10 days**.
H: He has worked **for many days**.

Dans cet exemple, nous pouvons remarquer que l'adverbial temporel **« During 10 days »** l'infère l'adverbial **« many days ».**

Exemple 2:

14) <pair id="14" value="TRUE" >
T: **the day before yesterday,** Paul disappeared.
H: **two days ago,** Paul disappeared.

Dans cet exemple nous remarquons que l'adverbial temporel **« the day before yesterday»** infère l'adverbial **« two days ago».**

### 3.1.3) Les inférences entre dates et adverbiaux temporels

L'inférence permet d'établir des relations temporelles entre dates et adverbes.
L'exemple suivant permet de montrer la relation qui peut exister entre un adverbial temporel et une date.

Exemple 1:

18) <pair id="18" value="TRUE" >
T: the building collapsed at **2 o'clock p.m**.
H: in **the afternoon** the building collapsed.

Dans l'exemple précédant nous pouvons remarquer que « **2 o'clock p.m** » infère l'adverbial «**the afternoon ».**

Exemple 2:

19) <pair id="19" value="TRUE" >
T: Mark has arrived **on Monday**, **the day after** Celine has arrived.
H: Celine has arrived **on Tuesday**.

Dans l'exemple précédant nous pouvons remarquer que si nous ajoutons « **the day after** » à « **Monday»** nous arrivons à **«Tuesday».** Ceci implique une inférence entre ces adverbiaux temporels.

### 3.3.2) Les inférences entre évènements

L'inférence permet d'établir des relations temporelles entre événements. Dans ce contexte, nous avons détecté deux types d'inférences, une qui demande la relation entre événements pour détecter l'inférence, et l'autre ne demande que l'inférence lexico sémantique.
Cette figure présente le nombre de paires de textes dans chaque sous classe dans le corpus.

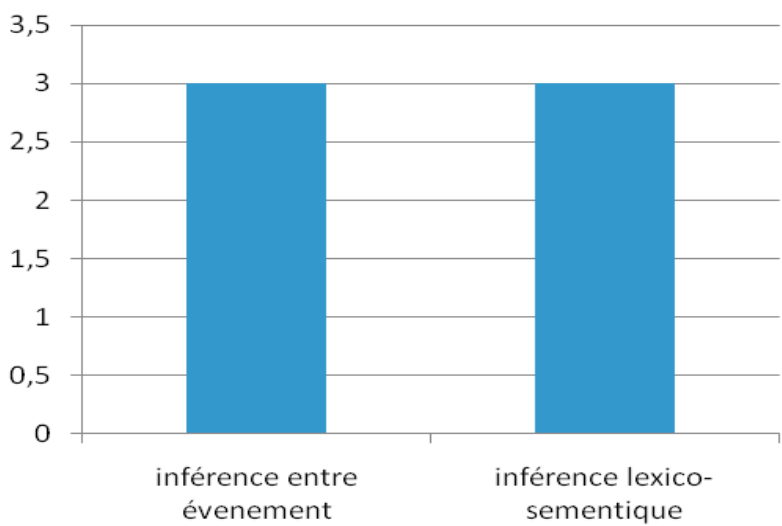

**Figure 3.4 : Nombre de paires par types d'inférences**

### 3.3.2.1) Les relations entre évènements temporels

La relation temporelle entre événements est établie par rapport aux relations qu'elle peut avoir avec d'autres événements dans le texte.

L'exemple suivant permet de montrer la relation qui peut exister entre un adverbial temporel et une date.

Exemple 1:

22) <pair id="22" value="TRUE" >
T: since **the death of Turing**, the scientific community **gives** the Turing prize to researchers who found out discoveries in computer science.
H: The Turing prize was not **given** before **the death of Turing**.

Dans l'exemple précédent, nous pouvons apercevoir que les deux événements « **given** » apparaissant dans les deux segments dépendent d'autres événements « **the death of turing** » pour se situer dans le temps.

Exemple 2:

23) <pair id="23" value="TRUE" >
T: Algeria has become **independent**.
H: before its **independence** Algeria was colonized.

Dans l'exemple précédent, nous pouvons apercevoir que l'événement « **independent** » apparaissant dans le segment H, dépendent de l'événement « **was colonized** » pour se situer dans le temps.

### 3.3.2.2) Les inférences lexico sémantiques

La relation temporelle entre événements est établie par rapport aux relations sémantiques qui peuvent exister entre eux.
L'exemple suivant permet de montrer la relation lexico-sémantique existante entre deux évènements.

Exemple 1:

26) <pair id="26" value="TRUE" >
T: France has **won** the match against Brasil.
H: France has **played** the match against Brasil.

Dans l'exemple précédent, nous pouvons constater que l'évènement « **won** » se produit après l'évènement « **played ».**

Exemple 2:

27) <pair id="27" value="TRUE" >
T : Amine **was dreaming** .
H : Amine **was sleeping deeply**.
Dans l'exemple précédant nous pouvons constater que l'évènement « **was dreaming»** se produit durant l'évènement « **was sleeping deeply».**

### 3.3.4) Les inférences entre évènements et expressions temporelles

L'inférence permet d'établir des relations temporelles entre événements et expressions temporelles.
Cette figure représente le nombre de paires de textes où existe ce type d'inférence.

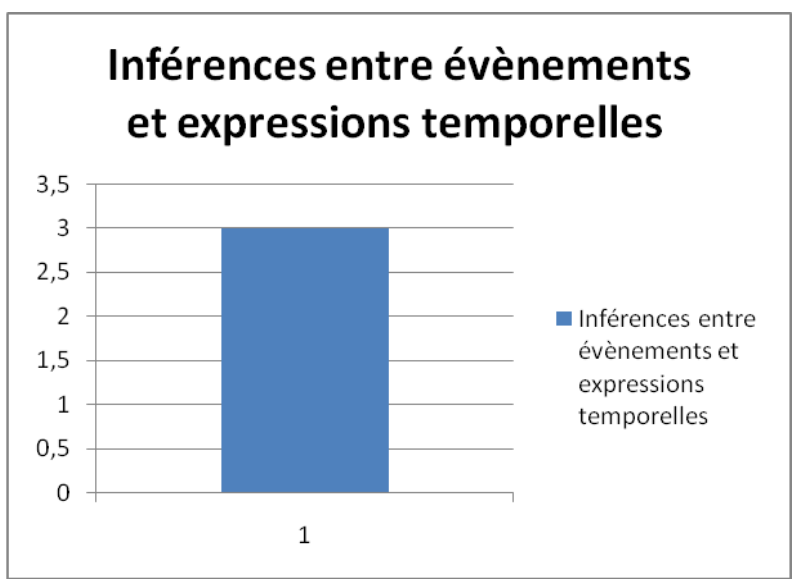

**Figure 3.5 : Nombre de paires par types d'inférences**

L'exemple suivant permet de montrer la relation temporelle existante entre évènements et expressions temporelles.
Exemple 1:

29) <pair id="29" value="TRUE" >
T: Japan gave weapons back after **the explosion of the first atomic bomb**.
H: Japan gave weapons back in **1945**.

Dans l'exemple précédent, nous pouvons remarquer que l'évènement « **the explosion of the first atomic bomb** » est ancré temporellement avec l'expression temporelle «**1945** ».

Exemple 2:

30) <pair id="30" value="TRUE" >
T: Germany has become unified since **the fall down of the Berlin wall.**
H: Germany unified **19 years ago**.

Dans l'exemple précédent, nous pouvons remarquer que l'évènement «**the fall down of the Berlin wall**» est ancré temporellement avec l'expression temporelle «**19 years ago**».

# 4) Le bilan de l'étude du corpus

Dans notre élaboration du corpus, nous nous sommes limités à des segments de textes relativement brefs et concrets. Nous retrouvons dans ce corpus des inférences temporelles sous des formes variées.
Le tableau suivant représente le pourcentage de paires du corpus de développement par type d'inférence temporelle existante, sachant qu'il existe 30 paires dans notre corpus.

| Types d'inférences temporelles | Nombres de paires |
|---|---|
| Inférences entre expressions temporelles | 21/30 |
| Inférences entre évènements | 6/30 |
| Inférences entre évènements et expressions temporelles | 3/30 |

**Tableau 3.1 : Nombre de paire dans le corpus**

Cette figure représente le pourcentage de paires de chaque type d'inférence dans le corpus de développement :

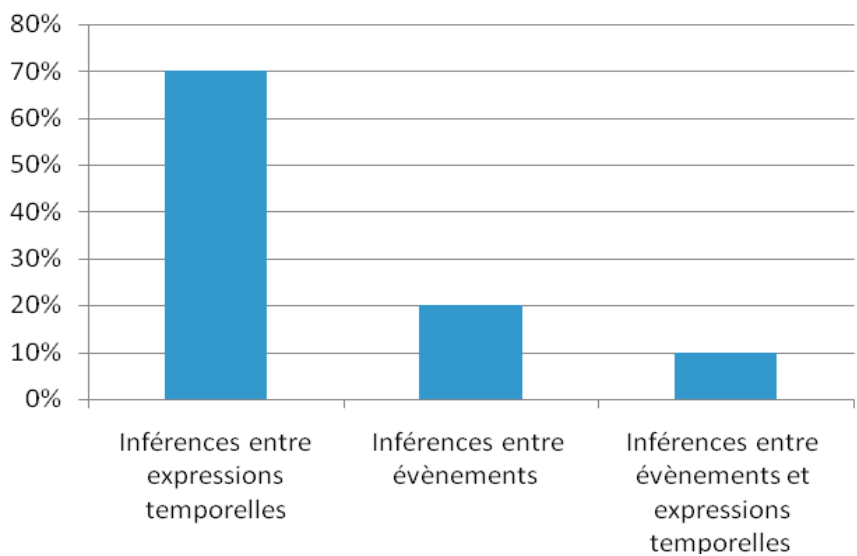

**Figure 3.6 : Pourcentage de paires par types d'inférences**

Nous constatons que notre corpus de développement a un pourcentage élevé de paires contenant une inférence temporelle entre expression temporelle et cela est dû à une forte présence de questions d'ordres temporelles extraites du corpus de test du challenge clé.
Les détails des corpus de test et de développement sont disponibles en annexe.

# 5) Conclusion

Dans ce chapitre nous avons expliqué, comment nous avons élaboré un corpus contenant des paires de segments de textes integrant des relations temporelles, ensuite nous avons fait une classification des différents types d'inférences temporelles existantes dans le corpus.
La suite logique de ce travail consiste à déduire des régles d'inférences temporelles et à les intégrer à un systéme d'inférence texuelle. Ces démarches sont l'objet du chapitre suivant que nous allons exposer.

- **Chapitre 4 -**

# La présentation du système TIMINF

**Chapitre 4**

# La présentation du système TIMINF

## 1) Introduction

Nous présentons, dans ce chapitre notre projet d'inférence temporelle, nommé TIMINF. Ce projet a pour but de développer et d'évaluer l'apport de l'inférence temporelle dans la reconnaissance de l'inférence textuelle.

L'un des principaux défis de ce type de système est de permettre aux systèmes d'inférences textuelles, d'ouvrir un voile sur l'inférence temporelle et d'explorer cette nouvelle approche. Dans ce cadre, l'objectif de TIMINF est de définir ce que devrait être un système d'inférence textuelle intégrant l'aspect temporel dans son fonctionnement, qui tient en compte la relation entre expression temporelle et relation entre les évènements dans la déduction de l'inférence textuelle.

Nous allons montrer tout au long de ce document comment nous avons concrétisé cet objectif. Nous décrivons alors les principaux modules constituant le système.

## 2) Architecture informatique de TIMINF

L'architecture générale de TIMINF, telle que déduite de l'analyse du corpus présenté au chapitre précédent, est illustrée dans la figure 4.1 suivante. Cette dernière s'articule autour de trois étapes essentielles qui sont :

- Le prétraitement qui permet de repérer les données temporelles et les composants syntaxiques de la paire de texte (T, H).

- L'inférence textuelle qui contient les modules de test d'inférence textuelle et du balisage des expressions temporelles.

- L'inférence temporelle qui contient les moteurs d'inférence et les règles d'inférences.

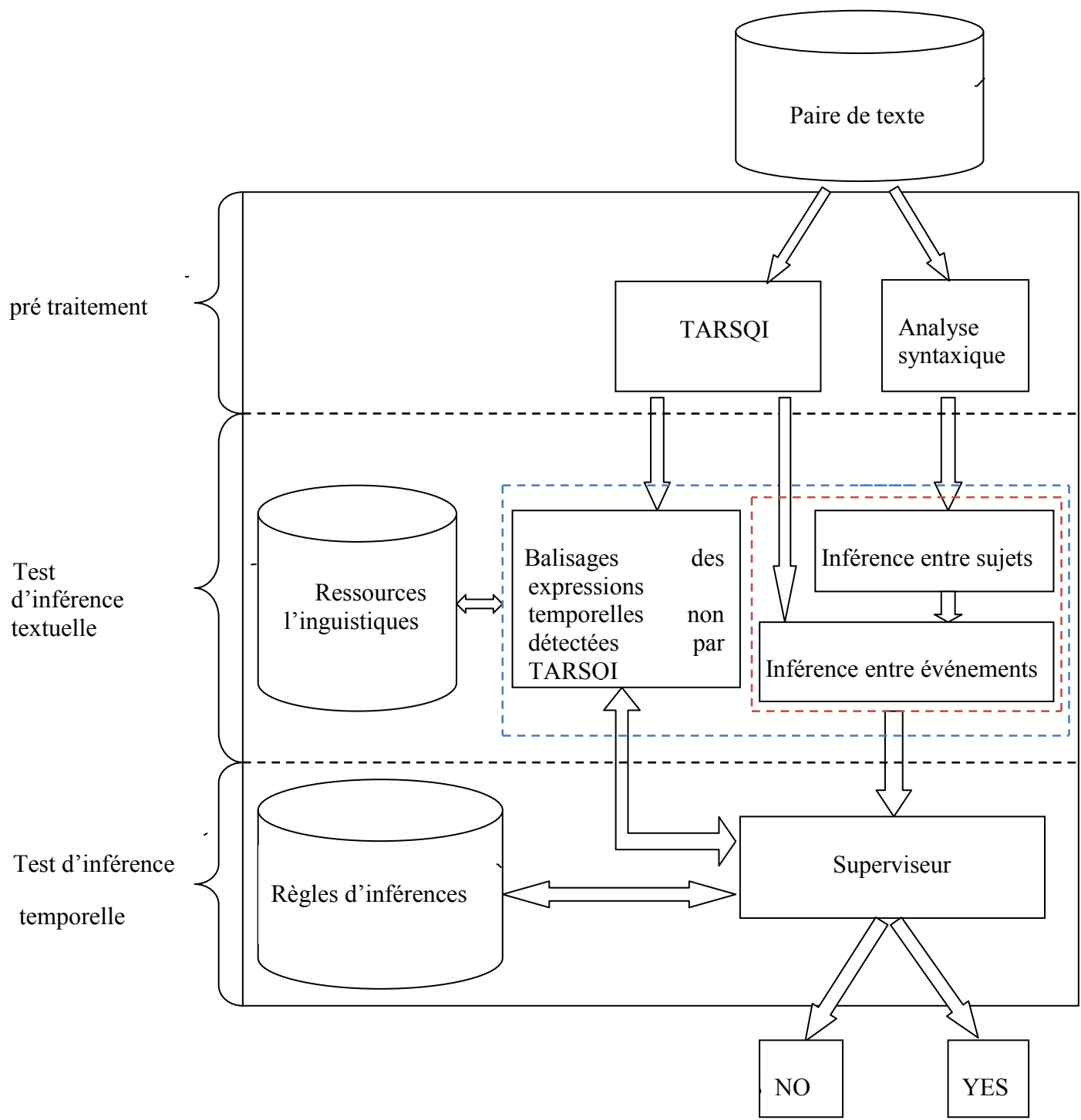

**Figure 4.1 : Architecture du système TIMINF**

Dans ce qui suit nous présentons les différents modules constituants le système TIMINF.

## 2.1) Le prétraitement

Le prétraitement est effectué par les deux modules TARSQI et LINK parseur. Ces deux modules s'exécutent en parallèle et nous permettent respectivement de repérer les données temporelles et les composants syntaxiques de la paire de texte (T, H). Nous détaillerons dans ce qui suit les deux modules et leurs utilisations dans notre système.

### 2.1.1) Le projet TARSQI

TARSQI est un outil permettant d'organiser des textes en langages naturels en fonction de leurs caractéristiques temporelles (Pustejovsky et al., 2003). Son objectif est d'annoter les données temporelles dans un texte en langage naturel, d'extraire des données temporelles à partir de textes et d'effectuer des raisonnements sur les données temporelles (http://www.timeml.org). Afin de répondre à ces différents objectifs, le module TARSQI utilise les balises TimeML pour marquer les expressions temporelles, les événements, les relations temporelles et les Subordinations syntaxiques des événements. Le système TARSQI est mis en place comme une cascade de modules successivement ajoutés.

L'architecture du système est définie dans le schéma ci-dessous.

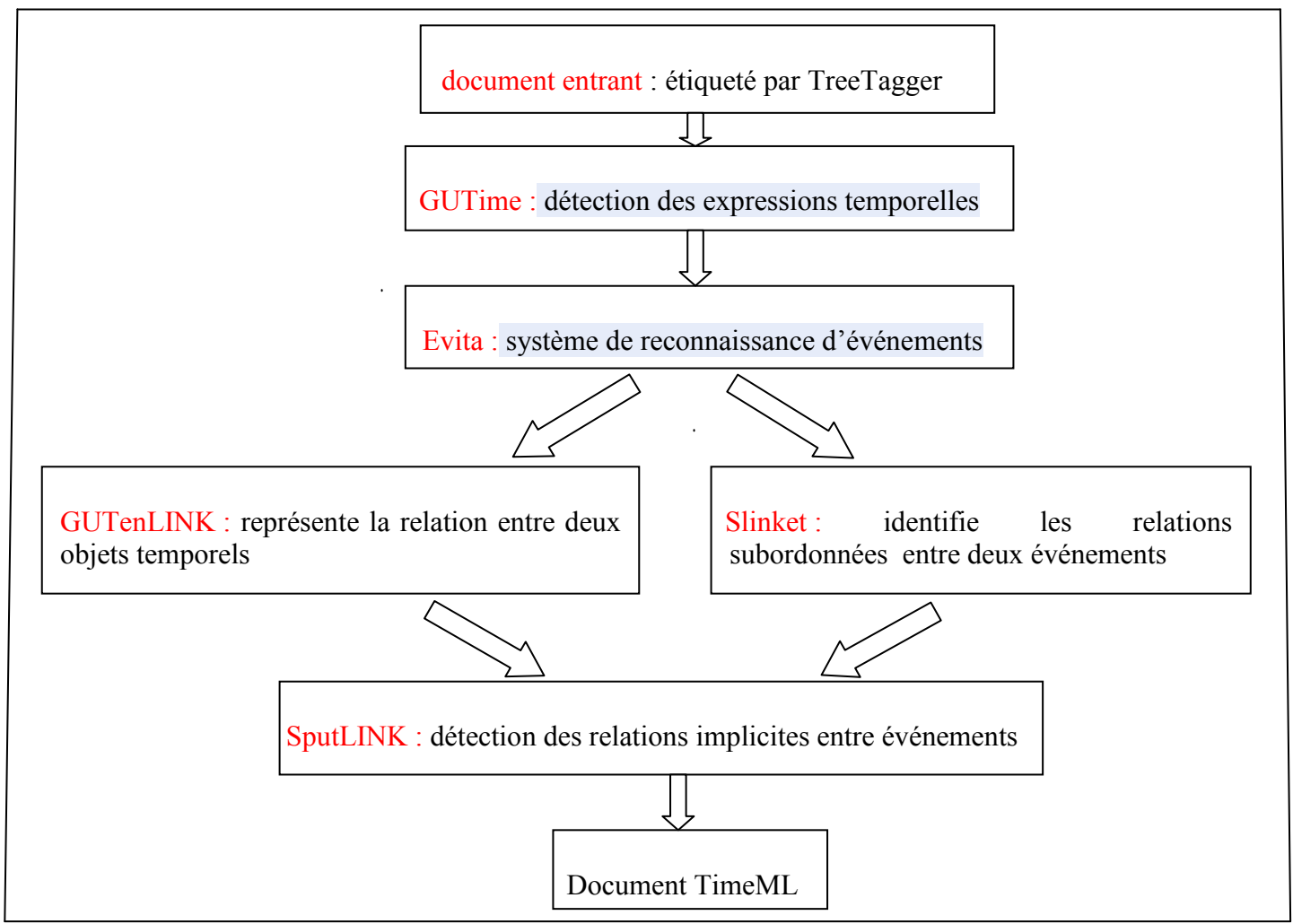

**Figure 4.2 : Architecture du module TARSQI**

Le module **TARSQI** doit avoir comme entrée des documents prétraités syntaxiquement. Pour cela, les concepteurs de **TARSQI** ont choisi d'utiliser une analyse morphosyntaxique avec le module **TreeTagger**.

Dans ce qui suit nous allons décrire le module **TreeTagger.**

### 2.1.1.1) TreeTagger

C'est un système d'étiquetage automatique des catégories grammaticales des mots avec lemmatisation et tokenisation (Helmut Schmid, 1994) (www.ims.uni-stuttgart.de/projekte/corplex/**TreeTagger**/).

Le module Treetagger a comme entrée un texte brut et il admet deux types de sorties :

### A) Une sortie en forme de tableau

Comme il est montré dans l'exemple suivant (figure 4.3), le mode de sortie est un tableau représentant l'étiquetage des mots dans la phrase.

---

Entrée :

```
Le TreeTagger est facile à utiliser.
```

Sortie:

| Mot | POS | Lemme |
|---|---|---|
| Le | DT | La |
| TreeTagger | NP | TreeTagger |
| Est | VBZ | Être |
| Facile | JJ | Facile |
| À | D' | À |
| Utiliser | VB | Utiliser |

---

**Figure 4.3 : Sortie en format tableau de TreeTagger**

Sachant que **:**

**Mot :** représente le mot étiqueté.

**POS :** représente la catégorie grammaticale du mot par exemple (VB pour verbe, DT pour un déterminant…..).

**Lemme :** représente la lemmatisation du mot.

### B) Sortie format XML

Avec La sortie format XML, chaque mot est tagué avec les balises de TreeTagger.

Exemple d'entrée, sortie TreeTagger :

```
Entrée:
He also slept on Friday night.
Sortie:
<BODY>
<TEXT>
<s> <NG><lex pos="PP">He</lex></NG> <lex pos="RB">also</lex> <VG <lex pos="VBD">slept</lex></VG> <lex pos="IN">on</lex> <NG> <lex pos="NNP">Friday</lex> <lex pos="NN">night</lex> </NG> <lex pos=".">.</lex> </s>
</TEXT>
</BODY>
```

**Figure 4.4 : Sortie en format XML de TreeTagger**

Les balises utilisées par TreeTagger sont :

- <BODY> contient le corps du document.
- <TEXT> contient le texte.
- Les phrases doivent être marquées d'un <s>.
- Le groupe nominal est balisé avec <NG> et le groupe verbal avec <VG>.
- chaque mot dans la phrase est balisé par <LEX>.

Les attributs utilisés par TreeTagger sont :

- **Stem :** représente la lemmatisation du mot qui est balisé.
- **Pos :** donne la catégorie grammaticale du mot balisé. (**DT** pour déterminant-nom, **PP** pour une préposition…). pour en savoir plus sur les différents symboles utilisés par Treetagger pour étiqueter les différentes catégories grammaticales, toutes les définitions des symboles sont disponibles sur le site (www.ims.uni-stuttgart.de/projekte/corplex/**TreeTagger**/).

### 2.1.1.2) GUTime

L'étiqueteur GUTime, développé à l'Université de Georgetown, utilise TIMEX3 tag pour représenter les expressions temporelles, telles que : les dates, les heures, les durées, etc (Mani et Wilson, 2000).

Il existe 3 types d'informations temporelles détectées par TIMEX3.

**DATE :** c'est-à-dire les années, les mois et les jours.

Exemple:

*USA were touched by terrorism in **September 11, 2001**.*

**TIME** : c'est-à-dire les heures de la journée.

Exemple:

*The building collapsed at **2 o'clock p.m.***

**DURATION :** représente un intervalle de temps entre deux dates.

Exemple:

*The end of the second world war happened **between 1940 and 1950**.*

Un exemple de sortie du module GUTime est montré ci-dessous :

```
In Washington <TIMEX3 tid="t1" TYPE="DATE" temporalFunction="true"
valueFromFunction="tf1"   anchorTimeID="t0">today</TIMEX3>,   the
Federal Aviation Administration released air traffic control tapes
from the night the TWA Flight eight hundred went down.
```

**Figure 4.5: Sortie du module GUTime**

Les attributs de TIMEX3 dans l'exemple sont :

**Tid :** donne l'identifiant de l'expression temporelle, pour chaque expression tagger par TIMEX a son propre identifiant.
**Type :** chaque TIMEX est assigné à ces différents types {DATE, TIME, DURATION}.
**TemporalFunction :** c'est un attribut qui retourne si la date est précise dans le temps ou pas.

Exemple:
*Next Tuesday* → TemporalFunction= true.
*September 11,2001* → TemporalFunction= false.

**AnchorTimeID :** **s**'il y a un ancrage temporel de l'expression temporelle identifiée par **Tid** avec une autre expression temporelle, **AnchorTimeID** donne son identifiant.

### 2.1.1.3) Evita

Evita est un système de reconnaissance d'événements, pour cela le module utilise deux balises de TIMEML (EVENT et MAKEINTANCE) qui sont décrites ci-dessous :

#### A) EVENT

EVENT est utilisé pour annoter les événements dans un texte, syntaxiquement, les évènements sont généralement des verbes, mais un nom peut aussi être utilisé pour dénoter un événement.
Les différentes classes d'événements qui sont détectées sont représentées ci-dessous.

- **occurence :** la plupart des événements font partie de cette classe. Ils décrivent ce qui se produit dans le monde.
- **state :** les états décrivant les circonstances dans lesquelles un événement a lieu et dont l'état peut être modifié ; et les états introduits par les i-action, i-state et reporting.
- **Reporting :** description de l'action d'une personne par un acte narratif.
- **i-action :** une action intentionnelle introduisant un autre événement, comme un essai, une enquête, un rapport, un ordre, une demande, une promesse, une nomination.
- **i-state :** similaire à i-action mais pour identifier un état tel que penser, ressentir. suspecter, douter, vouloir, désirer, détester, être prêt, être capable.
- **aspectual :** un événement débutant, terminant ou continuant une action.
- **Perception :** constatation physique d'un événement telle qu'entendre ou voir l'action.

## B) MAKEINSTANCE

MAKEINSTANCE est une réalisation de lien, il indique les différentes instances d'un événement donné.

Dans l'annotation, les <EVENT> ne participe jamais à une relation, c'est la réalisation (<MAKEINSTANCE>) de l'événement qui y participe et chaque EVENT introduit au moins un correspondant MAKEINSTANCE.

Un exemple de sortie du module Evita est montré ci-dessous:

```
In Washington today, the Federal Aviation Administration <EVENT eid="e1"
class="OCCURRENCE">released</EVENT> air traffic control tapes from the
night the TWA Flight eight hundred <EVENT eid="e2"
class="OCCURRENCE">went</EVENT> down.

<MAKEINSTANCE    eventID="e1"    eiid="ei1"    pos="VERB"    tense="PAST"
aspect="NONE"/>
<MAKEINSTANCE    eventID="e2"    eiid="ei2"    pos="VERB"    tense="PAST"
aspect="NONE"/>
```

**Figure 4.6 : Sortie du module Evita**

Les attributs de EVENT dans l'exemple sont :
**Eid :** donne l'identifiant de l'évènement, pour chaque évènement tagger par EVENT a son propre identifiant.
**Class :** détermine la classe auquel appartient l'évènement.
Les attributs de MAKEINSTANCE dans l'exemple sont :
**eventID :** donne l'identifiant de l'évènement, pour chaque évènement tagger par EVENT a son propre identifiant.
**Eiid :** instance de l'événement trouvé dans le texte.
**Pos :** donne la catégorie grammaticale du mot balisé.
**Tense :** donne le temps de l'évènement si l'évènement est un verbe.

## 2.1.1.4) GutenLink

GutenLink est un module de TARSQI qui utilise les balises TLINK de TIMEML pour représenter la relation entre deux objets temporels, que ce soit deux événements, deux marqueurs temporels ou un marqueur temporel et un événement. Il y a quatorze types de relations identifiées par le module, bien que certaines soient simplement l'inverse d'autre :

- **before** et **after** spécifient qu'un objet temporel précède ou suit l'autre objet temporel de la relation ;
- **ibefore** et **iafter** spécifient qu'un objet temporel est immédiatement avant ou après un autre.
- **includes** et **is-included** spécifient qu'un objet temporel inclut ou est inclus dans un autre, p. *ex. John arrived in Montreal yesterday.*
- **during** spécifie que l'état ou l'événement se poursuit durant une période de temps, *p. ex. John taught for 90 minutes.*
- **during-inv** est l'inverse de la relation précédente.
- **simultaneous** spécifie que deux instances d'événements semblent coïncider dans le Temps.
- **identity** indique que deux objets temporels représentent le même événement.
- **begins** spécifie qu'un événement débute par l'objet temporel avec lequel il est lié.
- **begun-by** est l'inverse de begin, elle relie un objet temporel à un événement débutant par l'objet temporel.
- **ends** et **ended-by** sont similaires aux deux relations précédentes sauf qu'elles Spécifient la fin de l'événement.

Un exemple de sortie du module **GutenLink** est montré ci-dessous :

```
In     Washington     <TIMEX3  tid="t1"   TYPE="DATE"   VAL="PRESENT_REF"
temporalFunction="true"                        valueFromFunction="tf1"
anchorTimeID="t0">today</TIMEX3>, the Federal Aviation Administration
<EVENT eid="e1" class="OCCURRENCE">released</EVENT> air traffic control
tapes from the night the TWA Flight eight hundred <EVENT eid="e2"
class="OCCURRENCE">went</EVENT> down. There's nothing new on why the plane
<EVENT eid="e3" class="OCCURRENCE">exploded</EVENT>, but you <EVENT
eid="e4"    class="OCCURRENCE">cannot</EVENT>        <EVENT     eid="e5"
class="OCCURRENCE">miss</EVENT> the moment. ABC's Lisa Stark <EVENT
eid="e6" class="OCCURRENCE">has</EVENT> more.

<MAKEINSTANCE    eventID="e1"    pos="VERB"    eiid="ei1"    tense="PAST"
aspect="NONE"/>
<MAKEINSTANCE    eventID="e2"    pos="VERB"    eiid="ei2"    tense="PAST"
aspect="NONE"/>
<MAKEINSTANCE    eventID="e3"    pos="VERB"    eiid="ei3"    tense="PAST"
aspect="NONE"/>
<MAKEINSTANCE   eventID="e4"   pos="VERB"   eiid="ei4"   tense="PRESENT"
aspect="NONE"/>
<MAKEINSTANCE  eventID="e5"  pos="VERB"  eiid="ei5"  tense="INFINITIVE"
aspect="NONE"/>
<MAKEINSTANCE   eventID="e6"   pos="NONE"   eiid="ei6"   tense="PRESENT"
aspect="NONE"/>

<TLINK   eventInstanceID="ei1"   relatedToTime="t1"   relType="IS_INCLUDED"
rule="2-1"/>
<TLINK   eventInstanceID="ei2"   relatedToTime="t1"   relType="IS_INCLUDED"
rule="2-1"/>
<TLINK eventInstanceID="ei1" relatedToEventInstance="ei3" relType="BEFORE"
rule="3-19"/>
<TLINK eventInstanceID="ei3" relatedToEventInstance="ei4" relType="BEFORE"
rule="6-1"/>
<TLINK eventInstanceID="ei3" relatedToEventInstance="ei6" relType="BEFORE"
rule="3-23"/>
```

**Figure 4.7 : Sortie du module GutenLink**

Les attributs de TLINK dans l'exemple sont :

**eventInstanceID :** donne l'identifiant de l'évènement.
**relatedToTime :** donne l'identifiant de l'expression temporelle.
**relType :** donne la relation temporelle existant entre les l'expressions temporelles, ils utilisent pour cela les relations d'Allen.

## 2.1.1.5) Slinket

Les liens subordonnants <SLINK> identifient les relations entre deux événements. Ils sont habituellement introduits par des verbes modaux qui impliquent une confirmation.

Les liens subordonnants sont définis selon six types de relations qui interagissent avec les classes d'événements reporting, i-state et i-action (modal introduit la possibilité d'un événement, *p. ex. John promised Mary to buy some beer)*.

Les différentes classes d'événements qui sont détectées sont représentées ci-dessous.

- **evidential** introduit la perception ou le compte-rendu de l'événement, *p. ex. Johnsaid he bought a pack of beer.*
- **neg-evidential** introduit la perception ou rapporte que l'événement ne s'est pas réalisé, *p. ex. John denied he bought beers*
- factive est une action qui implique ou présuppose qu'un événement a déjà eu lieu, *p. ex. John managed to leave the party.*
- **counter-factive** est la négation de la relation précédente *p. ex. John forgot to buy beers.*
- **conditional** indique que la réalisation de l'action entraînera l'événement en relation.

Un exemple de sortie du module SLINKET est montré ci-dessous:

```
The    Soviet    Union    <EVENT    eid="e12"    class="REPORTING">said</EVENT>
today it had <EVENT eid="e13" class="OCCURRENCE">sent</EVENT> an envoy to
the Middle East.

<MAKEINSTANCE    eventID="e12"    eiid="ei12"    tense="PAST"    aspect="NONE"
pos="VERB"/>
<MAKEINSTANCE    eventID="e13"    eiid="ei13"    tense="PAST"    aspect="PERFECTIVE"
pos="VERB"/>

<SLINK            relType="EVIDENTIAL"                    eventInstanceID="ei12"
subordinatedEventInstance="ei13" "/>
```

**Figure 4.8 : Sortie du module SLINKET**

Les attributs de SLINK dans l'exemple sont :

**eventInstanceID :** c'est l'identifiant de l'évènement concerné par la relation de subordination.
**subordinatedEventInstance** : c'est l'identifiant de l'évènement subordonné.
**relType :** donne la relation temporelle existante entre entités.

### 2.1.1.6) SputLink

Le module SputLink effectue des inférences temporelles en tenant compte des relations temporelles déjà générées par les modules qui le précèdent, c'est-à-dire (GUTenLINK et Slinket) et génère de nouvelles relations temporelles.

SputLink est fondé sur l'algèbre d'intervalle fondé par James Allen's en 1983.

Allen réduit tous les évènements et expressions de temps à 13 intervalles de bases et identifie les relations entre les intervalles. Les informations temporelles dans un document sont représentées comme un graphe où les événements et les expressions temporelles forment les nœuds, les relations temporelles forment les arcs.

Exemple

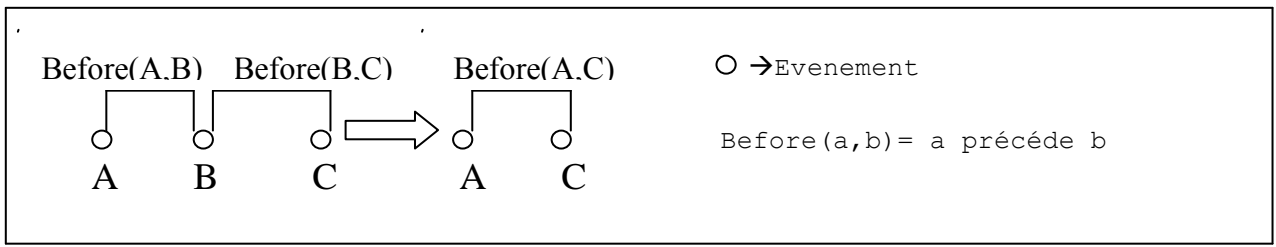

**Figure 4.9: Inférence effectué par le module SputLINK**

Ainsi, si A précède B et B précède C. des deux relations, on déduit que A précède C.

### 2.1.1.7) L'utilisation de TARSQI

Afin de permettre la portabilité du module TARSQI, les concepteurs ont proposé deux formats d'entrée possible à TARSQI qui sont décrits ci-dessous:

#### A) Format simple–xml

Avec ce format, l'analyse morphosyntaxique est incluse dans le module TARSQI. L'entrée est représentée par le format suivant:

Exemple d'entrée simple-xml.

```
<DOC>
<DOCID> Simple Test </DOCID>
<TEXT>
 In the afternoon, the building collapsed.
</TEXT>
</DOC>
```

**Figure 4.10 : Entrée format simple-xml**

Les balises de simpe_xml :

- <DOC> pour annoter le début et la fin du document.
- <DOCID> contient le type du document.
- <TEXT> contient le texte.

### B) Format RTE3

Avec ce format, l'analyse morphosyntaxique n'est pas incluse dans le module TARSQI et nous avons comme entrée le format RTE3, qui est le résultat d'un prétraitement effectué par le groupe COGEX qui travaille sur le RTE. Le groupe a choisi de développer son propre prétraitement.

Nous présentons dans ce qui suit un exemple de sortie du format RTE3 :

Exemple :

```
<XML version="1.0" ?>

<pair length="short" task="IE" id="1">

<t><s>text1</s></t>

<br/><h><s><NG>

<HEAD><lex start="0" end="12" pos="NNP" stem="Le Beau Serge">Le Beau Serge</lex>   </HEAD></NG><VG><lex start="14" end="16" pos="VBD" stem="be">was</lex>

<HEAD><lex start="18" end="25" pos="VBN" stem="direct">directed</lex> </HEAD></VG>

<HEAD><lex start="27" end="28" pos="IN" stem="by">by</lex> </HEAD><NG>

<HEAD><lex start="30" end="36" pos="NNP" stem="Chabrol">Chabrol</lex> </HEAD></NG><lex start="37" end="37" pos="." stem=".">.</lex> </s></h>

</pair>
```

**Figure 4.11 : Sortie du module GutenLink**

Les balises du format RTE3 sont :

- Les phrases doivent être marquées d'<s>.
- Les groupes nominaux sont balisés avec <NG> et les groupes verbaux avec <VG>.
- Les débuts de phrases sont marqués par des balises <HEAD>.
- <t> représente la premier phrase et <s> représente la deuxième phrase.
- <pair> représente la paire de phrases.

Les attributs :

- **Start :** représente la position du caractère de début de la chaine dans le texte.
- **End :** représente la position du caractère de fin de la chaine dans le texte.
- **Stem :** représente la lemmatisation du mot qui est balisé.

- **Pos :** donne la catégorie grammaticale du mot balisé.

### 2.1.1.8) L'intégration de TARSQI au système TIMINF

Dans notre système d'inférence, nous avons utilisé le format simple-xml au lieu de RTE3 car nous avons choisi d'utiliser le module TreeTagger pour l'analyse morphosyntaxique qui est intégré dans le module TARSQI dans le format simple-xml.

En plus de la détection des expressions temporelles, la phase de prétraitement intègre l'analyse syntaxique pour détecter la relation grammaticale entre les mots dans une phrase. Dans ce qui suit, nous allons présenter l'outil que nous avons choisi pour effectuer l'analyse syntaxique.

## 2.1.2) L'analyse syntaxique

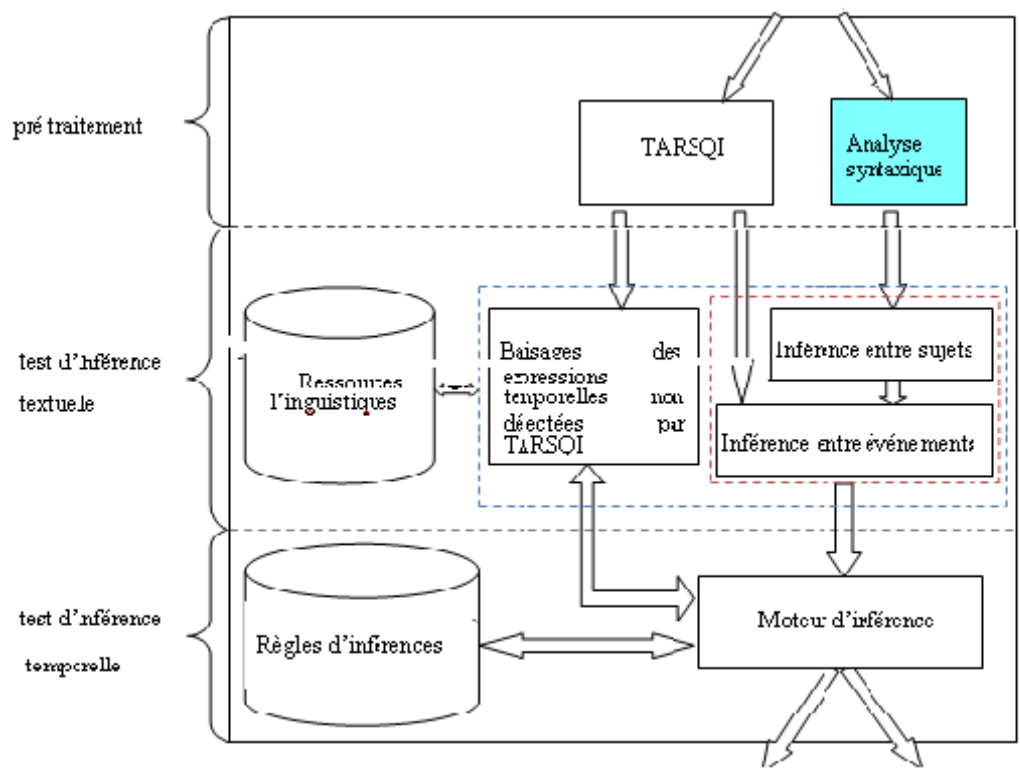

**Figure 4.12 : L'analyse syntaxique**

### 2.1.2.1) Présentation de link grammar parser

Nous avons utilisé le *Link Grammar Parser* (Sleator et Temperley, 1991) qui est un analyseur syntaxique de la langue anglaise, basé sur la dépendance syntaxique.
Partant d'une phrase fournie en entrée, cet analyseur produit un ou plusieurs graphes de dépendances, qui consistent en un ensemble de liens reliant des paires de mots.
Les **nœuds** du graphe sont les mots de la phrase. Certains d'entre eux ont un suffixe qui indique la partie du discours (nom, verbe, adjectif,adverbe, préposition, etc.).

Les **arcs étiquetés** relient les nœuds du graphe. Chaque étiquette précise un rôle grammatical (**D** pour déterminant-nom, **S** pour sujet-verbe…). Dans ce qui suit, nous montrons un exemple de sortie du parseur *Link Grammar Parser*.
Exemple :

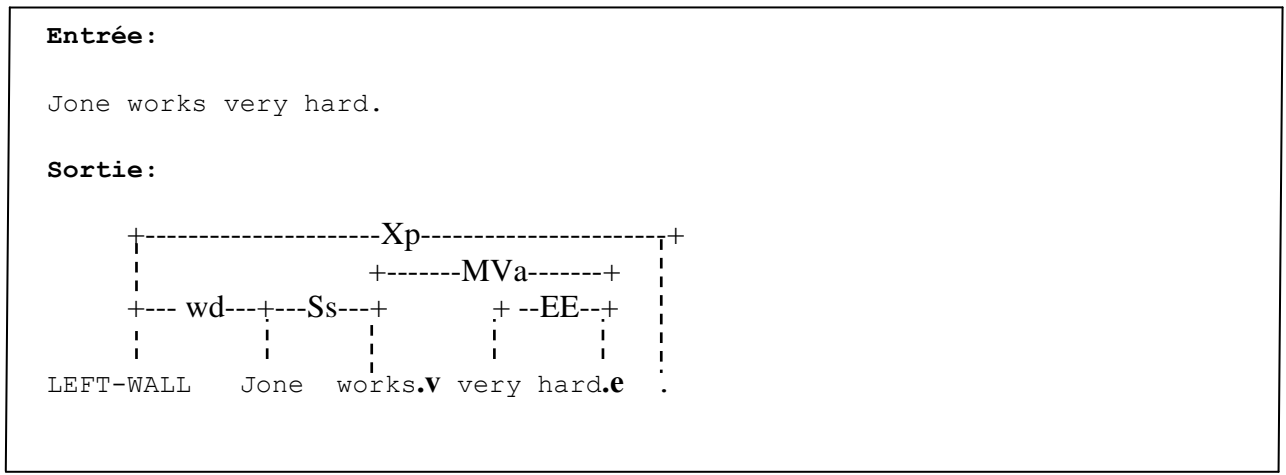

**Figure 4.13 : Sortie du module Link Grammar Parser**

Les définitions des différents liens représentés dans le graphe sont les suivantes :

EE→ adverbe se connecte à un autre adverbe.
Ss→connecte le sujet au verbe.
Xp→connecte le début et la fin de la phrase.
MVa→connecte le verbe à l'adverbe.
Wd→ le premier mot est un sujet.
.v→ verbe.
.e→ adverbe.
LEFT-WALL → détermine le début de la phrase.

Pour en savoir plus sur les différents symboles utilisés par *Link Grammar Parser* pour étiqueter les différents liens grammaticaux, toutes les définitions des symboles sont disponibles sur le lien suivant (www.**link**.cs.cmu.edu/**link**/).

## 2.1.2.2) L'intégration du link parser à notre système

Concrètement l'analyseur syntaxique nous a permis de détecter les sujets dans les deux segments de textes (T, H) et de les baliser avec nos propres balises comme il figure dans l'exemple suivant.

Exemple:

<pair id="28" value="TRUE" >
*<s> **<syntax type: sujet>**Poland**</syntax>** became a communistic state in 1945.</s><s>*
***<syntax type: sujet>**Poland**</syntax>** has become a communistic state since the invasion of Russians.</s>*

La balise **<syntaxe type: sujet>***sujet***</syntaxe>** est choisi pour baliser les sujets dans la paire(T, H).

Apres l'analyse syntaxique et le traitement par TARSQI, la paire de texte est prête à être soumise au test d'inférence textuelle qui a besoin des prétraitements effectués précédemment pour tester l'inférence textuelle. Dans ce qui suit nous décrivons les différents constituants de la phase de test d'inférence textuelle.

## 2.2) Les tests d'inférences textuelles

La deuxième phase s'articule autour de deux modules. Le premier permet de tester l'inférence textuelle pour savoir s'il y a une inférence textuelle ou pas et le deuxième module permet de détecter les expressions temporelles non détectées par TARSQI. Les deux modules exploitent des ressources linguistiques.

Dans ce qui suit nous allons présenter les deux modules et les ressources linguistiques utilisées :

### 2.2.1) Les tests d'inférences entre événements et entre sujets

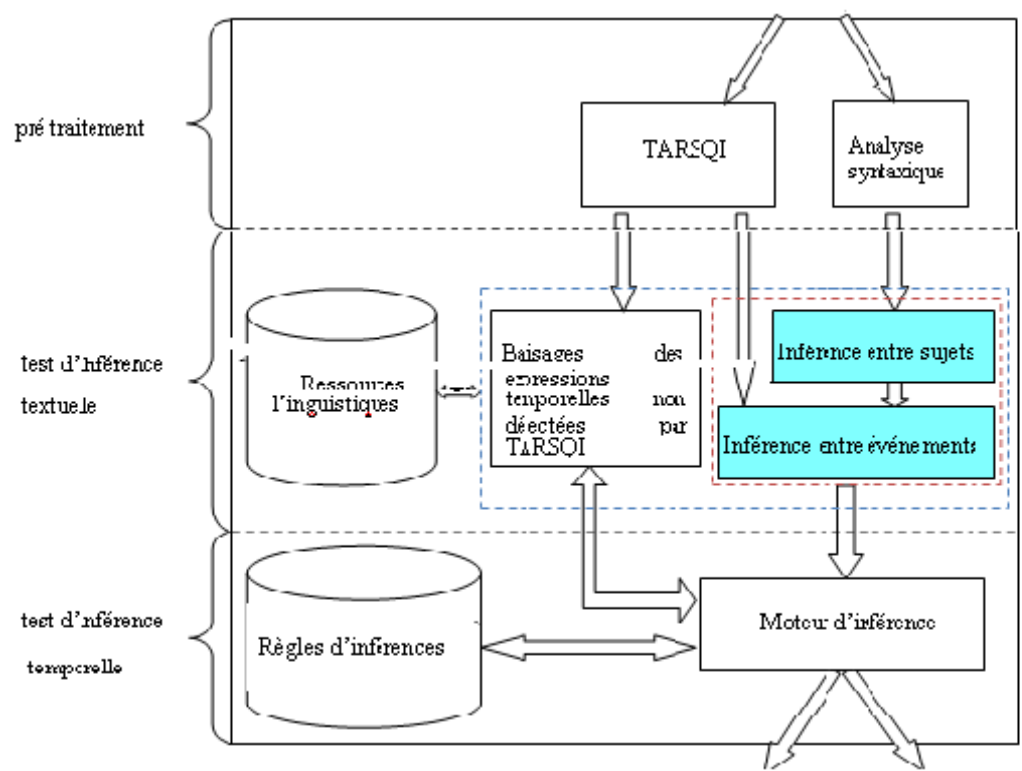

**Figure 4.14 : L'inférence entre évènements et sujets**

Le but de ce module est de détecter s'il y a une inférence textuelle entre les deux paires de textes (T, H).

Le module est mis en place comme une cascade de sous modules successives. Le premier module détecte les inférences textuelles entre les sujets des deux segments de textes (T, H) et le deuxième détecte les inférences textuelles entre évènements des deux segments de textes (T, H).

Nous décrivons ci-dessous les deux modules d'inférences :

## 2.2.1.1) L'inférence entre sujets

Le module d'**inférence entre sujets** détecte s'il y a une inférence textuelle entre les sujets du texte H avec les sujets du texte T. Pour cela, le module utilise les sorties du module *LINK parser* c'est-à-dire que pour chaque sujet détecté, dans le texte H nous recherchons s'il y a une relation de synonymie avec un des sujets du texte T. Pour cela, le module emploie WordNet pour retrouver toutes les relations ontologiques qui lient les deux entités (l'utilisation de wordNet dans notre système est détaillée dans le chapitre cinq). Aussi nous utilisons le comptage de mots pour comparer des groupes de mots (l'algorithme de comptage de mots est expliqué dans l'exemple (1)).

Ce module accepte comme entrée au module le résultat de l'analyse syntaxique des paires de texte T et H et en sortie il existe deux possibilités :

- Si le module trouve une équivalence entre deux sujets, il déclenche le module d'inférence entre événements en lui envoyant les événements correspondants aux deux sujets.

- Si le module ne trouve pas d'équivalence entre sujets, le module envoie le message « pas d'inférence » au module du test d'inférence.

Exemple (1) : paire numéro 3 du corpus de développement.

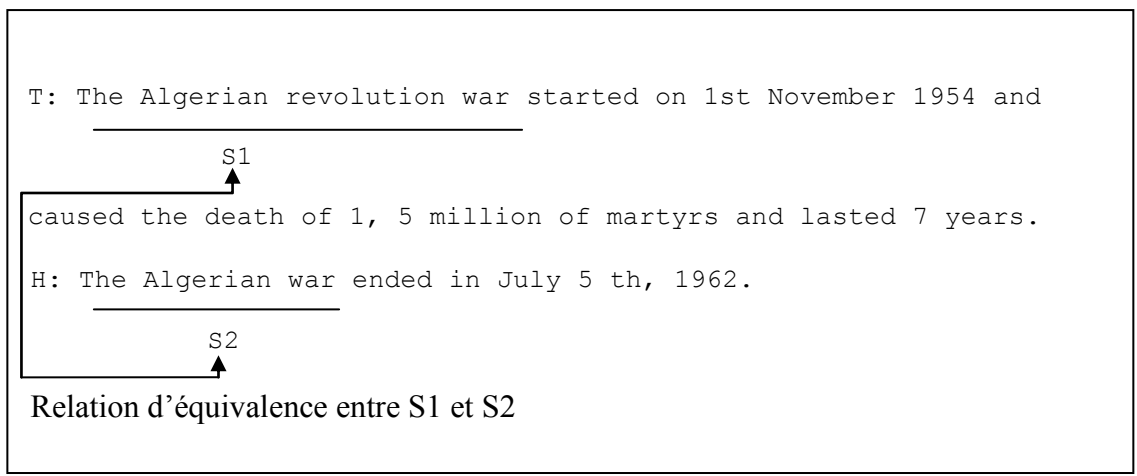

**Figure 3.15 : Exemple d'inférence entre sujets**

Dans l'exemple suivant le module détecte tous les sujets contenus dans le segment T et le segment H et les met dans deux listes différentes, ensuit il effectue la comparaison entre les évènements des deux listes.

Dans notre exemple la première liste ne contient qu'un seul sujet {S1} et la deuxième liste contient le sujet {S2}. Une relation d'équivalence est détectée entre les évènements S1 et S2.

Pour détecter l'équivalence le module utilise l'algorithme de comptage de mots pour déduire l'inférence entre «`the Algerian war`» et «`the Algerian revolution war`».

**Le comptage de mots :** L'algorithme récupère les deux groupes deux mots dans deux listes différents et compare chaque mot d'une liste avec les mots contenu dans la deuxième liste et s'il y a un seul mot qui est semblable ou sous mot d'un mot de la deuxième liste, il considère qu'il y a une inférence entre sujets.

### 2.2.1.2) L'inférence entre évènements

Le module d'inférence entre évènements détecte s'il y a une relation ontologique entre les deux évènements reçus du module d'inférence entre sujets. Pour cela, le module emploie WORDNET pour retrouver toutes les relations qui lient les deux entités.

Le module a comme entrée les évènements reçus du module d'inférence entre sujets et le balisage de TARSQI et comme sortie les résultats suivants :

- S'il trouve une équivalence entre deux évènements, il envoie le message « oui» au module de test d'inférence.

- S'il trouve deux évènements contraires, il envoie le message « non » au module de test d'inférence.

- S'il ne trouve pas de relation entre événements, il envoie le message « pas d'inférence » au module de test d'inférence.

Exemple :

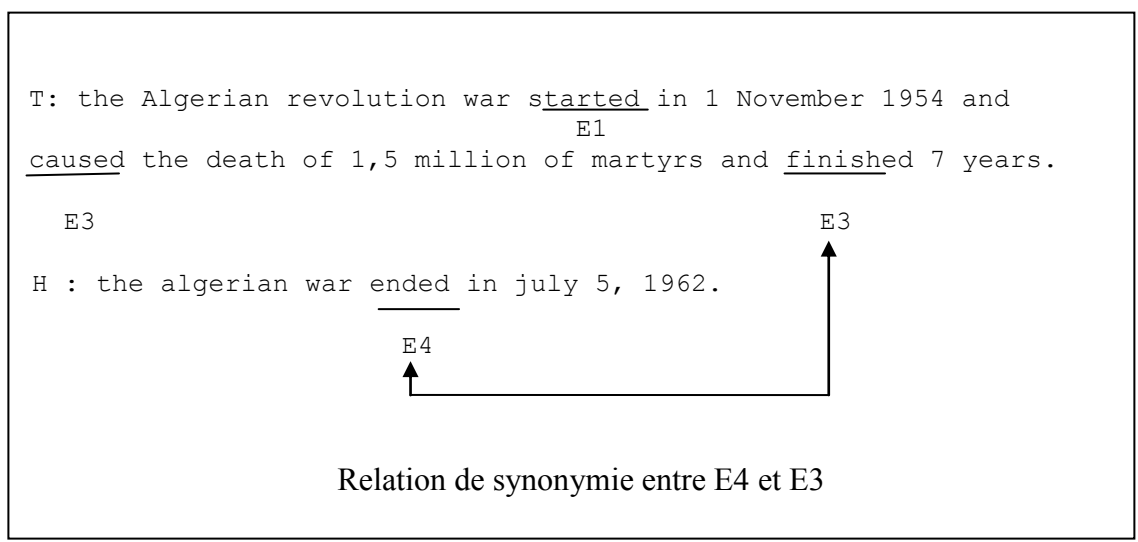

**Figure 4.16 : Exemple d'inférence entre évènements**

Dans l'exemple ci-dessus le module détecte une liste d'évènements dans le texte T {E1, E2, E3} et une autre liste d'évènements dans le texte H {E1} et effectue la comparaison entre les évènements des deux listes. Une relation ontologique (synonymie) est détectée entre les évènements E4 et E3.

## 2.2.2) Le balisage des expressions temporelles non détectées par TARSQI

Nous avons remarqué qu'au niveau de la détection des expressions temporelles, les modules de balisages existant ont un manque au niveau de la détection des entités nommées et des adverbiaux temporels.

Dans ce qui suit nous montrons les différentes balises utilisées :

**Entités nommées :** elles sont balisées par **<NE TYPE=" " Val=" ''>entité nommée</NE>.**

- TYPE contient le type d'expression temporelle {date, durée}.
- Val contient la date ou la durée correspondante.

Exemple :

```
T: Germany has become unified since t2: the fall down of the Berlin Wall.

H: Germany unified t1: 19 years ago.

T2 est balisé ainsi:

  <NE TYPE=" date "    Val=" 1989 ''> the fall down of the Berlin Wall</NE>.
```

**Figure 4.17 : Exemple de balisages d'expressions temporelles**

Notre objectif avec le balisage de « the fall down of the Berlin Wall» est de repérer l'évènement dans le temps.

Dans l'exemple précédent le balisage avec le module TARSQI ne détecte que « fall » comme événement et ne le relie pas à une date.

**Adverbiaux temporels** : ils sont balisés par <TIMEX3 tid="t" TYPE=" " VAL="" >

- **TYPE** contient le type d'expression temporelle {date, durée}.
- **Val** contient les entités que nous avons mises pour représenter les expressions temporelles.

Dans ce qui suit nous relions à chaque expression les symboles correspondants.

- Les jours de la semaine c'est à dire **{Monday, Tuesday, Wednesday, Thursday, Friday Saturday, Sunday}** sont représentés respectivement par des nombres de 1 à 7,

- **{Day before yesterday, two days ago, Yesterday},** sont représentés respectivement par {-2, -2, -1}

- **{everyday often}** sont représentés avec {often}.

- **{Someday, Many days, morning**, **evening, Afternoon}** sont représentés respectivement par PSD, PMD, aMORNING, aNIGHT et AFTERNOON.

## 2.3) Les Ressources linguistiques

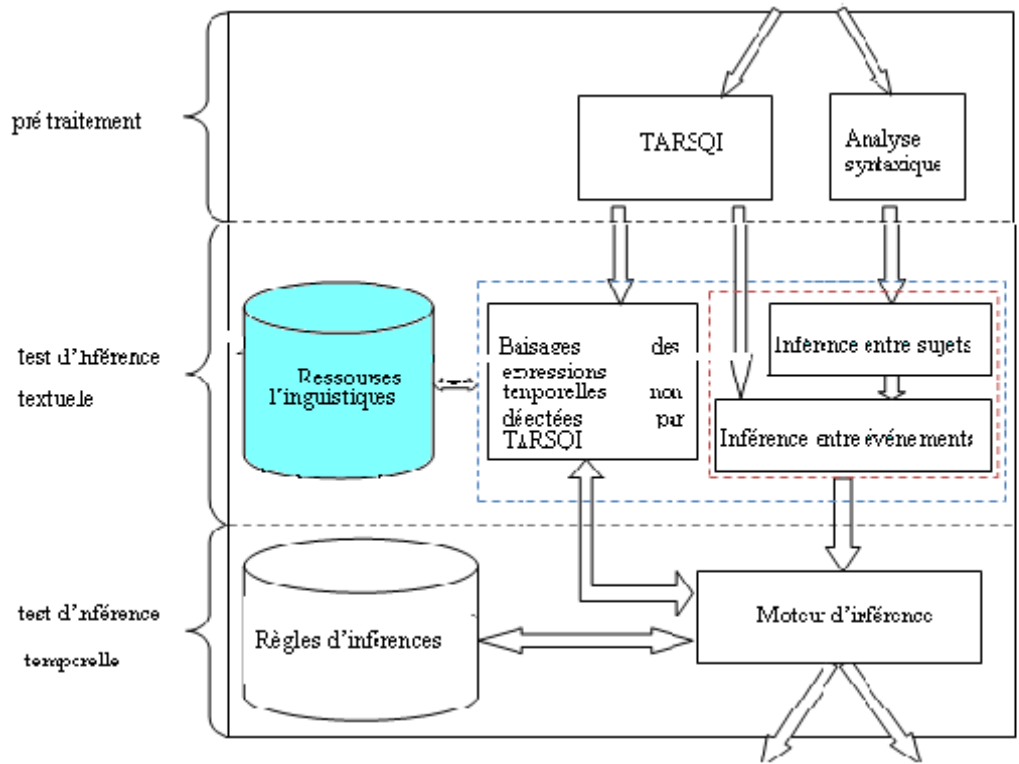

**Figure 4.18 : Ressources linguistiques**

Deux types de ressources sont utilisés :

### 2.3.1) Les ressources externes

Dans la conception de notre module d'inférence, l'utilisation d'une ressource lexicale est indispensable au bon fonctionnement des deux modules (inférence entre sujets et inférence entre événements). Pour cela, nous avons choisi d'utiliser **Wordnet** qui est la **base de données lexicale** qui correspond le plus à notre besoin en termes de relations ontologiques entre mots.

### 2.3.1)  Les ressources internes

Ce module est en fait une base de données lexicale contenant les différentes entités nommées qui sont utilisées par le module de balisage pour annoter les expressions temporelles non détectées par le module TARSQI. Puisque notre objectif est de se focaliser sur l'inférence entre expressions temporelles, non pas sur leur détection, nous avons effectué une annotation manuelle de ces expressions temporelles sachant qu'il existe des logiciels payant qui peuvent effectuer la détection.

## 2.4) Les tests d'inférences temporelles

Cette phase permet de détecter s'il y a une inférence temporelle et aussi textuelle entre les deux segments de textes T et H. Pour cela, nous utilisons un superviseur qui communique avec une base de règles d'inférences et d'après les résultats de la phase précédente (phase de test d'inférence textuelle), il décide de la règle à utiliser.

Dans ce qui suit nous décrivons les modules constituants cette phase.

### 2.4.1)  Les règles d'inférences

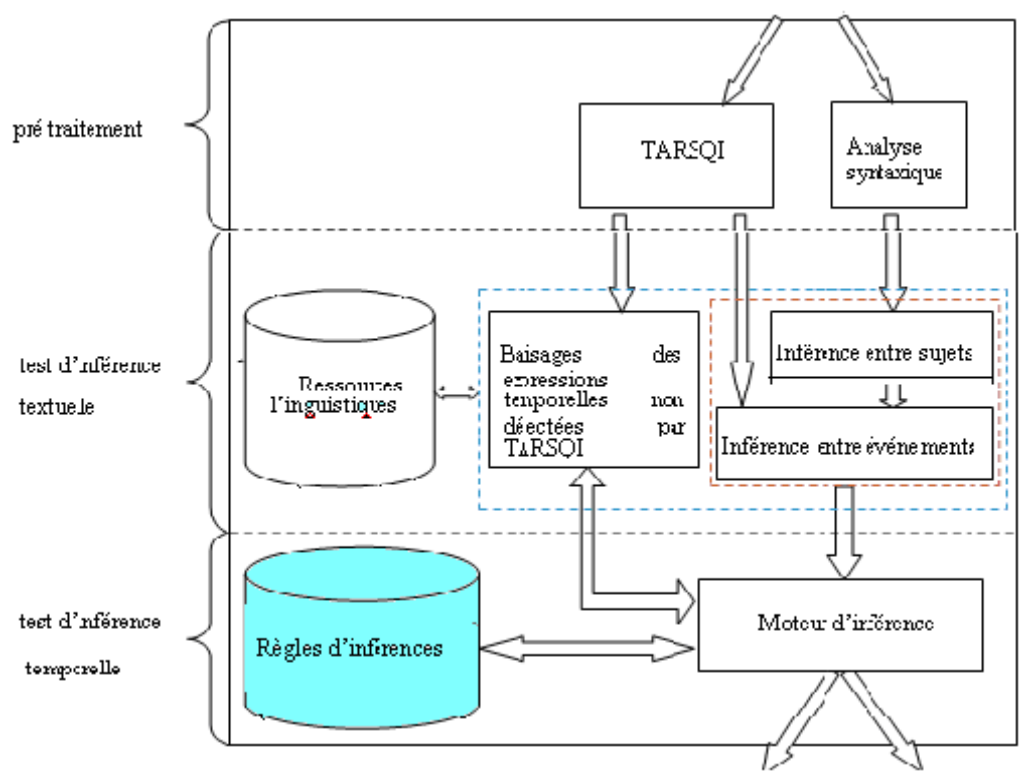

**Figure 4.19 : Règles d'inférences**

Les règles d'inférences sont divisées en deux groupes.

**Groupe 1 :** contient les fonctions qui testent si les événements ont un ancrage temporel identique.

**Groupe 2 :** contient les fonctions qui testent si les événements ont un ancrage temporel différent**.**

## 2.4.1.1) Définition des fonctions utilisées dans l'abstraction des règles d'inférences

Dans ce qui suit, nous allons définir toutes les fonctions que nous avons utilisées dans l'abstraction de nos règles d'inférences.

Sachant que **S** représente un des textes **T** ou **H** de la paire, **E** représente un événement dans le texte et **t** représente une expression temporelle**.**

- **<S, E>:** indique que l'événement **E** est dans le segment de texte **S**.

- **Subj** (**<S, E>**): la fonction retourne le **sujet** de l'évènement **E** dans le segment **S**.

- **Equivalent** (**<S,E1>**,**<S,E2>**): la fonction retourne s'il y a une équivalence entre **E1** et **E2** ou pas.

- **Contraire** (**<S,E1>**,**<S,E2>**): la fonction retourne s'il y a une antonymie entre **E1** et **E2** ou pas.

- **Inclut** (**<S,t>**,**<S,t>**): la fonction retourne si **t** est inclut dans l'intervalle de **t'** ou pas.

- **Egale** (**<S, t1>**,**<S, t2>**): la fonction retourne s'il y a une équivalence entre **t1** et **t2** ou pas.

- **Début** (**<S,t>**,**<S,E>**,type): la fonction et booléen et renvoi vrai ou faux, si **t** est la date de debut de l'événement **E** ou pas.

- **After**(**<S,E1>**,**<S,E2>**,type): la fonction retourne s'il y a une équivalence entre **t1** et **t2** ou pas.

- **before**(**<S,E1>**,**<S,E2>**,type): la fonction retourne si **E1** est avant **E2** ou pas .

- **Fin** (**<S,E>**,**<S, t>**,type): la fonction retourne si **t** est la date de fin de l'événement **E** ou pas.

- **Relation** (**<S, E>**,**<S,t>**,type): indique s'il y a une relation TLINK entre l'événement **E** et la date **t**. Tel que l'argument **type** indique le type d'expression temporelle {date, durée}.

- **Inf** (**T**, **H**): indique, en sortie s'il y à une inférence entre les segments de textes T et H ou pas.

- **Somme**(**<S, t1>**,**<S,t2>**): renvoi en sortie la somme des deux dates.

- **Différence**(**<S, t1>**,**<S,t2>**): renvoi en sortie la différence entre les deux dates.

Les symboles utilisés dans les schémas de représentation de nos règles d'inférences sont présentés ci-dessous :

──── **:** représente le lien entre les deux évènements.

------ **:** représente le lien entre les deux expressions temporelles.

------ **:** représente les éléments du texte T.

------ : représente les éléments du texte H.

──→ : représente le lien entre les événements et expressions temporelles.

◯     : représente l'événement.

☐     : représente l'expression temporelle.

.t     : représente une date.

.e     : représente un évènement.

.d     : représente une durée.

Ainsi, les différentes règles d'inférences conçues sont reparties comme suit :

## 4.1.1.2) Les règles du groupe 1

Ces règles permettent de savoir s'il y a un ancrage temporel entre évènements.

Si **équivalent** (<T, e1>, <H, e2>) ^ **équivalent** (<T, **Subj** (<T, e1>)>, <H, **Subj** (<H,e2>)>) alors :

### A) Règle R1

Si les différentes conditions se réunissent c'est-à-dire :

- détecter une équivalence entre les deux évènements (**e1**, **e2**)

- chaque événement est relie avec la même relation TLINK avec une date **e1→t1** et **e2→t2**

- les dates sont égales.

Nous aurons une inférence temporelle et textuelle entre les segments **T** et **H.**

L'abstraction de la règle R1 est représentée dans ce qui suit :

- o Si **relation** (<T,e1>,< T,t> ,date) ^ **relation** (<H,e2>, <H,t'>,date)

  alors **Inf**(T,H)= Vraie SSi **inclus**(<T,t> ,<H, t'>) v **égale**(<T,t> <H, t'>)

  sinon **inf**(T,H)=Faux

Les numéros des exemples dans le corpus de développement où les règles peuvent s'appliquer : 8, 7, 15, 14, 18, 28, 29, 30.

Cette figure représente la règle d'inférence R1:

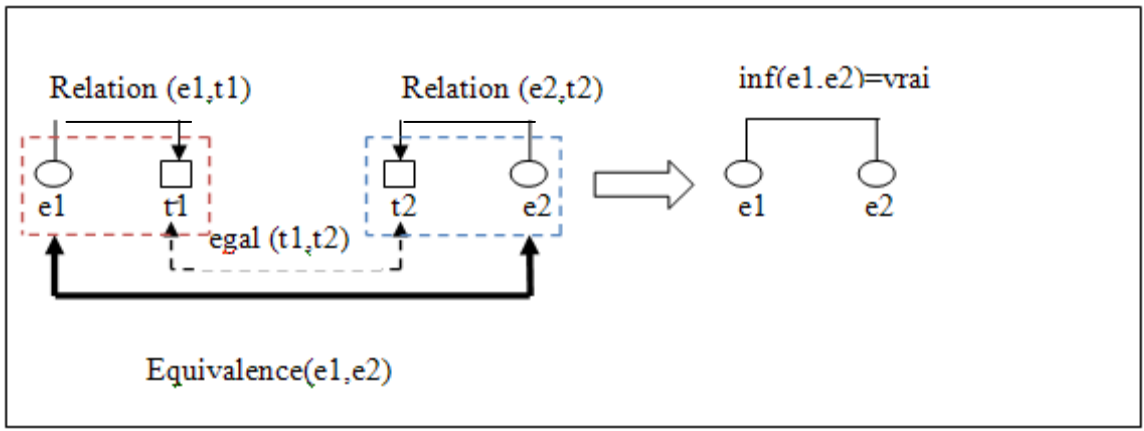

**Figure 4.20 : Règle R1 de l'inférence temporelle**

Dans l'exemple qui suit, nous allons appliquer la règle R1 sur la paire numéro 8 du corpus.

8) <pair id="9" value="TRUE" >
T: since its e1: **creation** in 1948, Israel had faced a lot of conflict with the Arabic countries.
H: Israel e2: **was conceived** in 1948.

Puisque l'événement e1 se déroule en 1948 et l'événement e2 se déroule entre 1948 et puisque e2 est le synonyme de e1 alors il y a une inférence temporelle entre e1 et e2.

### B) Règle R2

Si les différentes conditions se réunissent c'est-à-dire :

- détecter une équivalence entre les deux évènements (**e1**, **e2**)

- **t1** est la date de début de l'évènement **e1**, **t2** est la date de fin de l'évènement **e1** et l'événement **e2** est relié a une durée **e2→d**.

- la différence entre les dates **t1** et **t2** est égale à la durée **d.**

Nous aurons une inférence temporelle et textuelle entre les segments **T** et **H.**

L'abstraction de la règle R2 est représentée dans ce qui suit :

o Si     **debut**(<T,e1>,<T,t>,date)     ^     **fin**(<T,e1>,<T,t'>,date)     ^     **relation**(<H,e2>, <H,t''>,durée)

    Alors **Inf**(T,H)= égale(**Différence** (<T,t>,<T,t'>) , <H,t''>)

    sinon **Inf**(T,H)=Faux

Les numéros des exemples dans le corpus de développement où les règles peuvent s'appliquer: 4, 12.

Cette figure représente la règle d'inférence R2 :

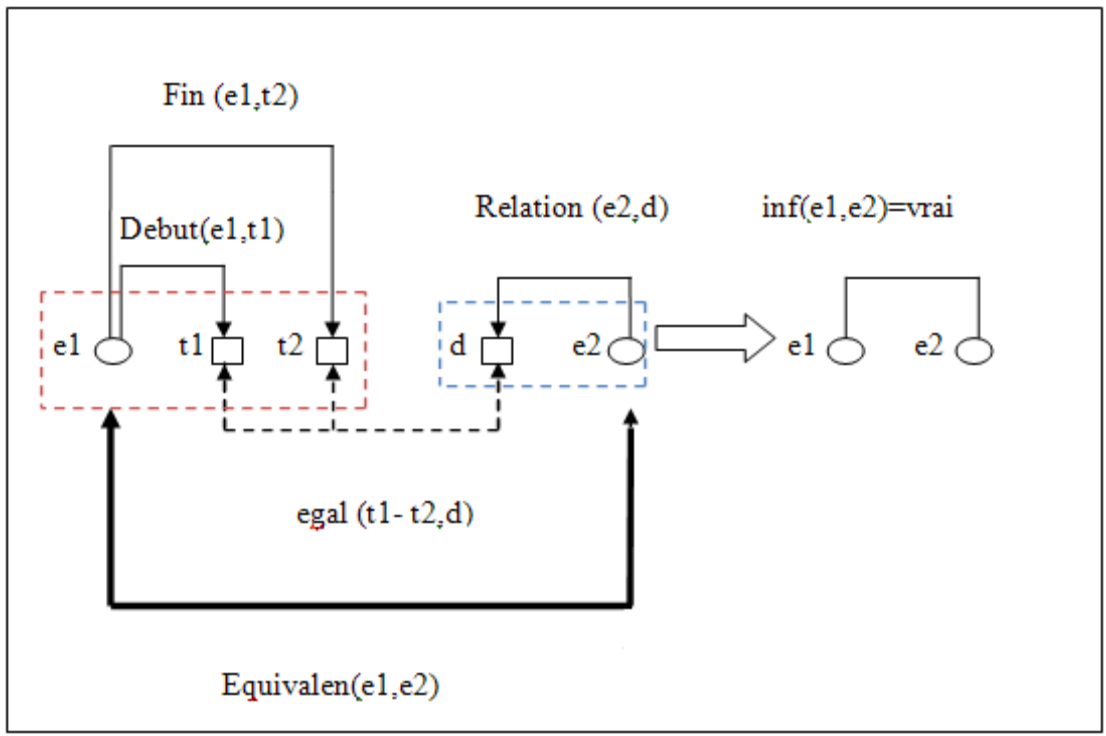

**Figure 4.21 : Règle R2 de l'inférence temporelle**

Dans l'exemple qui suit, nous allons appliquer la règle R2 sur la paire numéro 4 du corpus.

4) <pair id="4" value="TRUE" >
T: Pasteur began looking for the germ that causes rabies in 1880, and in july 1885 he **found** the efficient vaccine against the illness.
H: **to find** the vaccine, Pasteur's researches took five years.

Puisque l'événement e1 se déroule au même moment que l'évènement e2 car si nous ajoutons « **7 years**» à « **november the first 1954**» nous serions en 1962 qui est la date où s'est déroulé l'évènement e2 et puisque e2 est le synonyme de e1 alors il y a une inférence temporelle entre e1 et e2.

### C) Règle R3

Si les différentes conditions se réunissent c'est-à-dire :

- détecter une équivalence entre les deux évènements (**e1**, **e2**)

- **t1** est la date de début de l'évènement **e1**, événement **e1** est relié a une durée **e1→d** et **t2** est la date de fin de l'évènement **e1**.

- la somme entre la date **t1** et la durée **d** est égale à **t2.**

Nous aurons une inférence temporelle et textuelle entre les segments **T** et **H.**

L'abstraction de la règle R3 est représentée dans ce qui suit :

- Si **debut**(<T,e1>,<T,t>,date) ^ **relation**(<T,e1>,<T,t'>,durée) ^ **fin**(<H,e2>, <H,t''>,date)

    Alors **Inf**(T,H)= **egal**(**Somme**(<T,t>, <T,t'>) , <H,t''>)

    sinon **Inf**(T,H)= Faux

Les numéros des exemples dans le corpus de développement où les règles peuvent s'appliquer : 3, 11.

Cette figure représente la règle d'inférence R3:

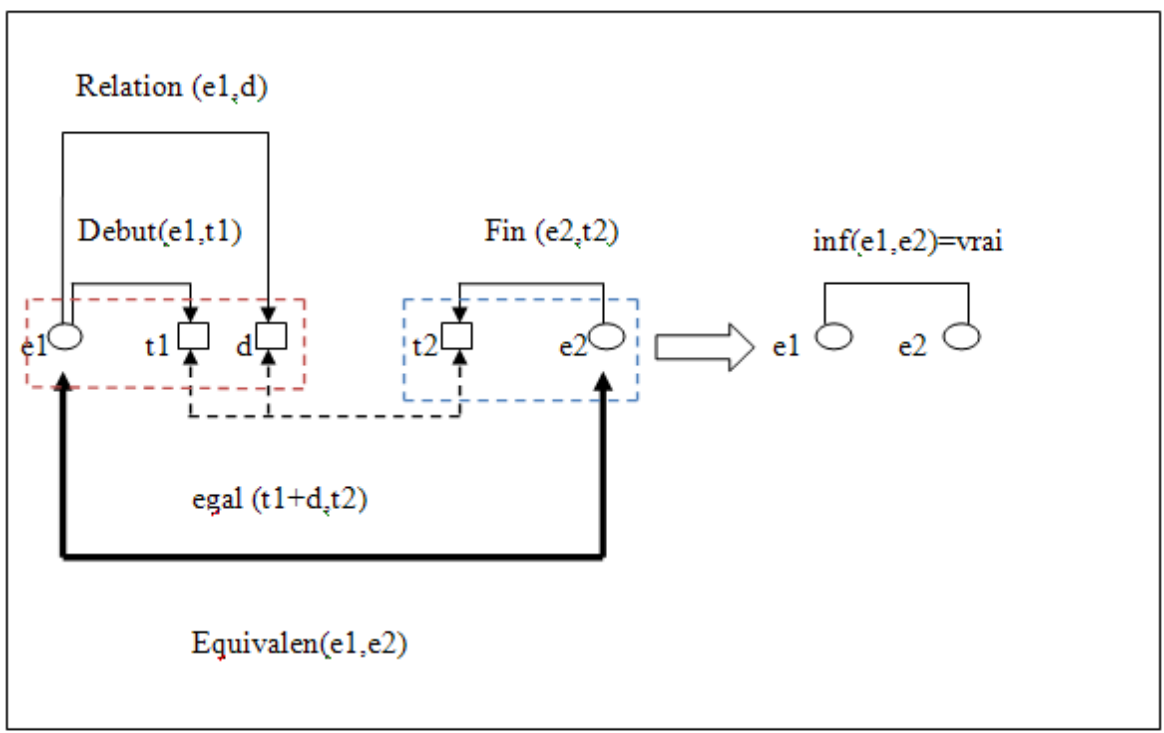

**Figure 4.22 : Règle R3 de l'inférence temporelle**

Dans l'exemple qui suit, nous allons appliquer la règle R3 sur la paire numéro 3 du corpus de développement.

3) <pair id="3" value="TRUE" >

T: the Algerian revolution war started on **november the first 1954**, it caused the death of 1,5 million of martyrs and it e1: **lasted 7 years**.
H: the Algerian revolution war e2: **ended** on **july the fifth 1962**.

L'événement e1 se déroule au même moment que l'événement e2 car si on ajoute « **7 years»** à « **november the first 1954»** nous serions en 1962 qui est la date où se déroule l'évènement e2 et puisque e2 est le synonyme de e1 alors il y a une inférence temporelle entre e1 et e2.

### D) Règle R4

Si les différentes conditions se réunissent c'est-à-dire :

- détecter une équivalence entre les deux évènements (**e1**, **e2**)

- **t1** est la date de fin de l'évènement **e1**, événement **e1** est relié à une durée **e1→d** et **t2** est la date de début de l'évènement **e1**.

- la somme entre la date **t1** et la durée **d** est égale a **t2.**

Nous aurons une inférence temporelle et textuelle entre les segments **T** et **H.**

L'abstraction de la règle R4 est représentée dans ce qui suit :

o Si **relation**(<T,e1>,<T,t>,durée) ^ **fin**(<T,e1>,<T, t'>,date) ^ **debut**(<H,e2>, <H, t''>,date)

Alors **Inf**(T, H)= **égale**(**Différence** (<T, t'>,<T, t>),<H, t''>)

sinon **Inf**(T, H)=Faux

Les numéros des exemples dans le corpus de développement où les règles peuvent s'appliquer : 10, 20.

Cette figure représente la règle d'inférence R4:

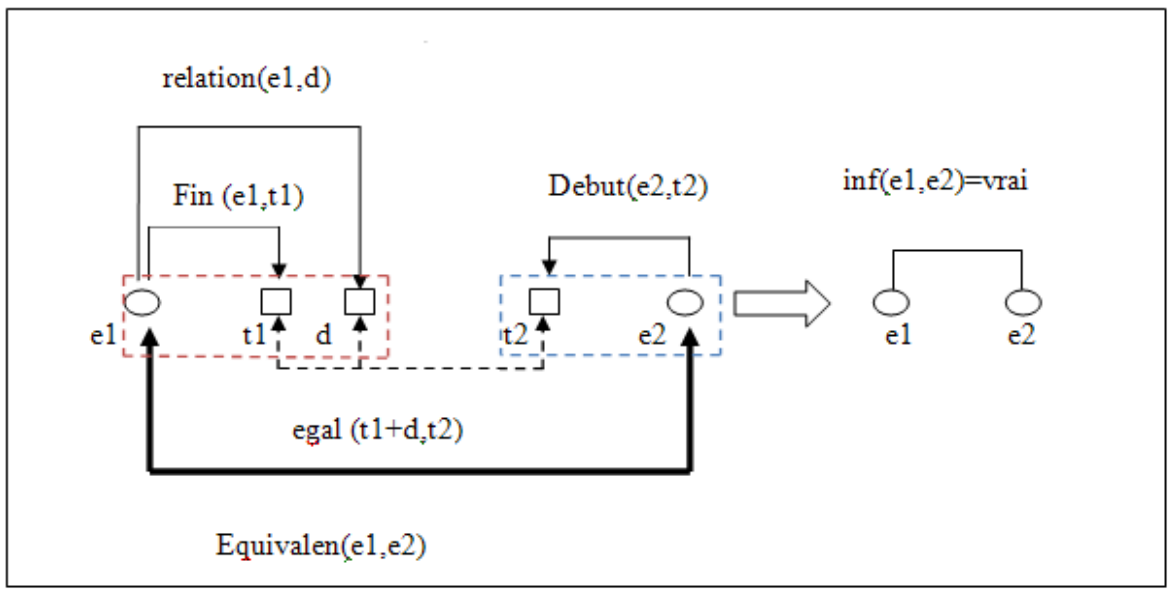

**Figure 4.23: Règle R4 de l'inférence temporelle**

Dans l'exemple qui suit, nous allons appliquer la règle R4 sur la paire numéro 10 du corpus.

10) <pair id="11" value="TRUE" >
T: on **t1:** december 2$_{nd}$ 1804, Napoleon Bonaparte became the emperor of the French, before **d1:** one year exactly, he **e1: won** the battle of Austerlitz.
H: in t2: 1803, Napoleon **e2: won** the battle of Austerlitz.

L'événement e1 se déroule au même moment que l'événement e2 car si on réduit « **one year exactly** » à « **december 2$_{nd}$ 1804** » nous serions en 1803 qui est la date où se déroule l'évènement e2 et puisque e2 est le synonyme de e1 alors il y a une inférence temporelle entre e1 et e2.

### E) Règle R5

Si les différentes conditions se réunissent c'est-à-dire :

- détecter une équivalence entre les deux évènements (**e1**, **e2**)

- l'événement **e1** est relié avec une relation TLINK **e1→t2** et l'événement **e1** est relié avec la même relation TLINK à une durée **e2→d.**

- inclusion entre la date **t1** et la durée **d.**

Nous aurons une inférence temporelle et textuelle entre les segments **T** et **H.**

L'abstraction de la règle R5 est représentée dans ce qui suit :

- **Relation** (<T, e1>, < T,t> ,date) ^ **relation**(<H,e2>, <H,t'>,durée)

    alors **Inf**(T, H)= Vraie SSi **inclus**(<T,t> ,<H, t'>)

    sinon **inf**(T, H)=Faux

Le numéro de l'exemple dans le corpus de développement où cette règle peut être appliquée: 1.
Cette figure représente la règle d'inférence R5:

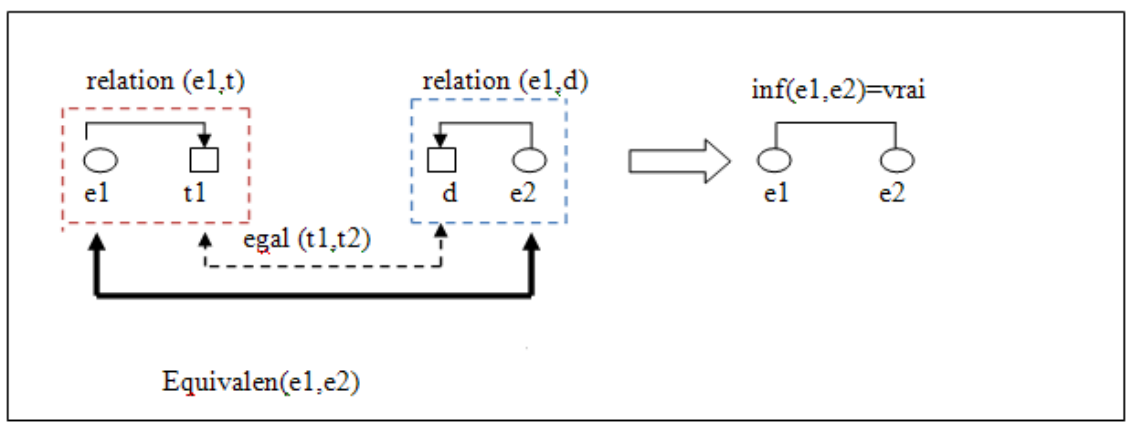

**Figure 4.24 : Règle R5 de l'inférence temporelle**

Application de la règle 5 sur la paire numéro 1 du corpus.

1) <pair id="1" value="TRUE" >
T: the second world war e1: **finished** in 1945.
H: the end of the second world war e2: **took part** between 1940 and 1950.

Puisque l'événement e1 ce déroule en 1945 et l'événement e2 se déroule entre 1940 et 1950 et puisque e2 est le synonyme de e1 alors il y a une inférence temporelle entre e1 et e2.

### 4.1.1.3) Groupe 2

Cette règle permet de savoir s'il n'y a pas d'ancrage temporel entre évènements.

Si **contraire**(<T,e1> ,<H,e2>) ^ **équivalent**(<T ,**Subj**(e1)> , <T,**subj**(e2)>) =<H,**Subj**(e2)>)

#### A) Règle R6

Si les différentes conditions se réunissent c'est-à-dire :

- détecter que l'évènement **e1** est le contraire de l'évènement **e2**.
- l'événement **e1** se produit soit avant ou après **e2**.

Nous aurons une inférence temporelle et textuelle entre les segments **T** et **H.**

L'abstraction de la règle R6 est représentée dans ce qui suit :

Si **relation** (<T, e1>, <T, t>, date) ^ **relation**(<H, e2>, <H, t'>, date)

    Alors

    **Inf**(T, H) = Vraie SSi **before**(**relation** (<T,e1>,<T,t>, date), **relation**(<H, e2>, <H, t'>, date)

    v

    **after**(**relation** (<T, e1>,<T, t>, date) , **relation**(<H, e2>, <H, t'>, date)

    sinon **inf**(T, H)=Faux

Les numéros des exemples dans le corpus de développement où les règles peuvent s'appliquer : 2, 5.

Cette figure représente la règle d'inférence R6:

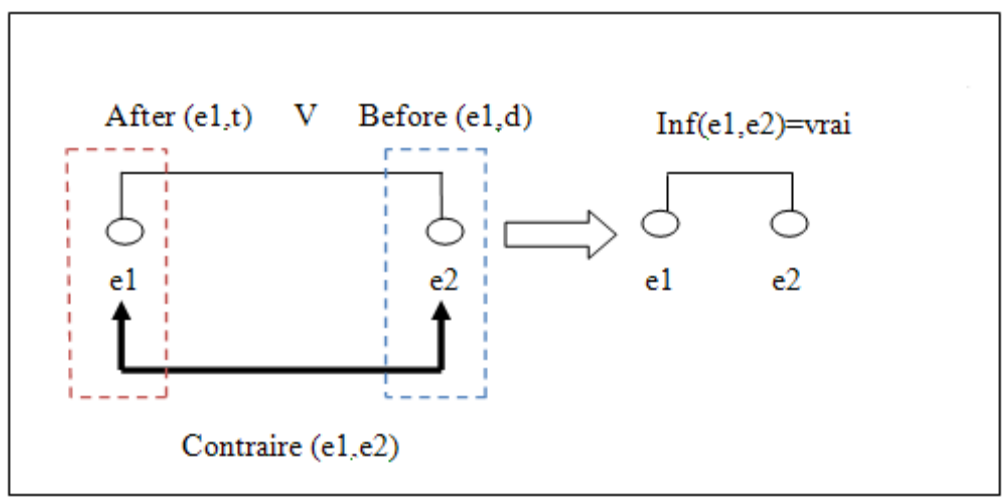

**Figure 4.25 : Règle R6 de l'inférence temporelle**

Application de la règle 6 sur la paire 2 du corpus.

2) <pair id="2" value="TRUE" >

T: Algeria got its e1: **independence** in **1962**.

H: **Before 1962** Algeria was e2: **colonized**.

Dans cet exemple ci-dessus, puisque l'événement e1 se déroule en 1962 et l'événement e2 se déroule avant l'évènement e1 et puisque e2 est l'antonyme de e1 alors il y a une inférence temporelle entre e1 et e2.

## 4.2.2) Le superviseur

Ce module, accepte en entrée, les résultats du module « inférence entre événements », le résultat de « TARSQI », « les ressources » à ajouter et « les règles d'inférences » et en sortie, il indique s'il y une inférence textuelle ou pas.

Le superviseur permet de choisir les règles d'inférences temporelles à appliquer et de décider de l'existence ou pas de l'inférence textuelle.

Ainsi, le superviseur applique la procédure suivante:

- Si le module a comme message « pas d'inférence » de la phase précédente c'est-à-dire du module de test d'inférence le superviseur va afficher, « pas d'inférence textuelle ».

- Si le module a comme message « non » qui veut dire qu'il y a une relation d'antonymie entre les évènements le module va exécuter les règles d'inférences temporelles qui détectent si les deux évènements ne sont pas ancrés temporellement.

- Si le module a comme message « oui » qui veut dire qu'il y a une relation de synonymie entre les évènements, le module va exécuter les règles d'inférences temporelles qui détectent si les événements sont ancrés temporellement.

Nous représentons dans la figure suivante l'architecture du superviseur :

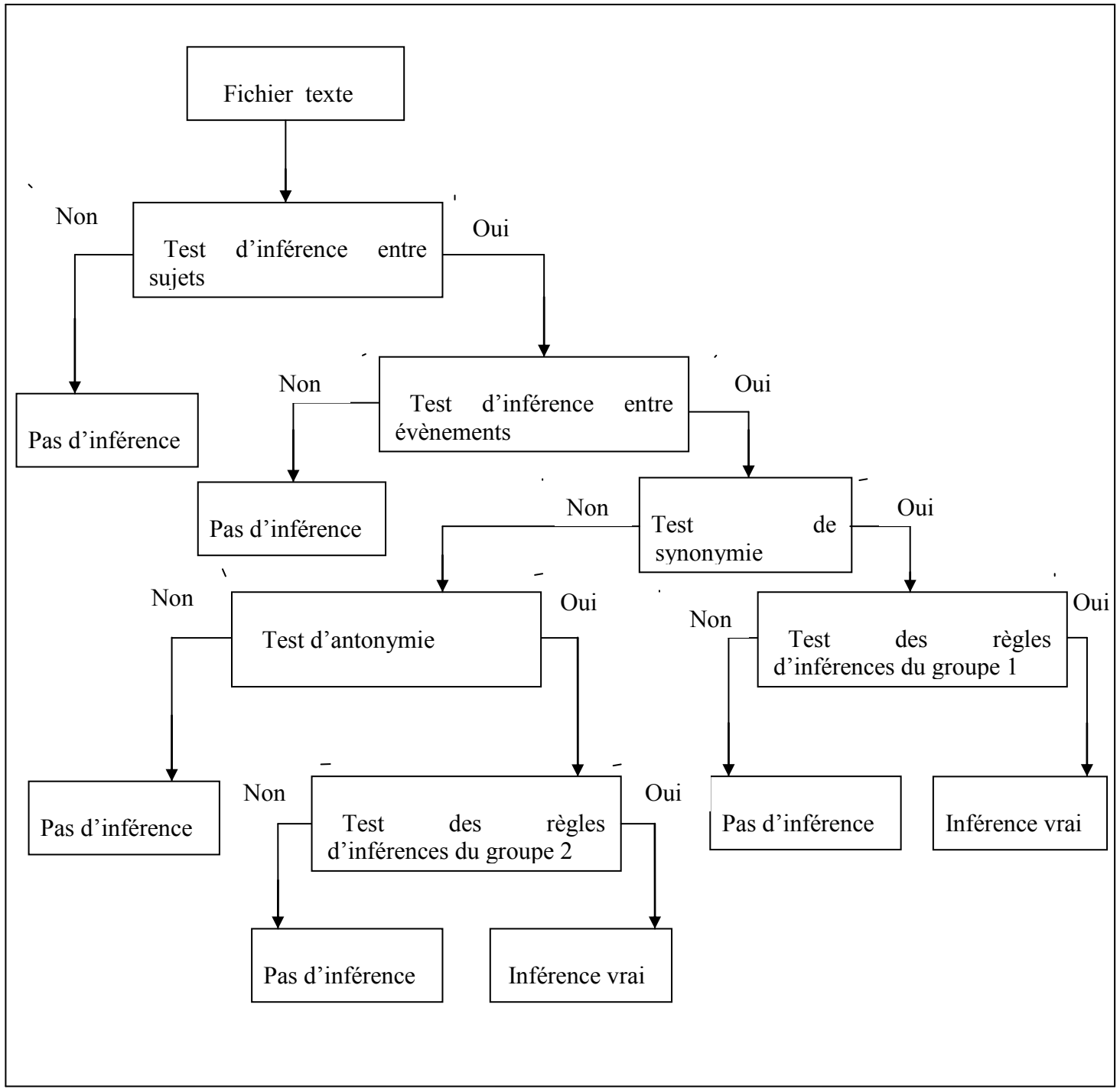

**Figure 4.26 : Architecture du superviseur**

Il existe des cas où plusieurs règles peuvent s'appliquer. Pour cela, le superviseur prend les mesures suivantes :

- S'il existe une fonction qui retourne une fausse inférence temporelle, cela implique qu'il n'y a pas d'inférence textuelle entre les segments T et H.

- Si toutes les fonctions retournent une inférence temporelle, cela implique qu'il y a une inférence textuelle entre les segments T et H.

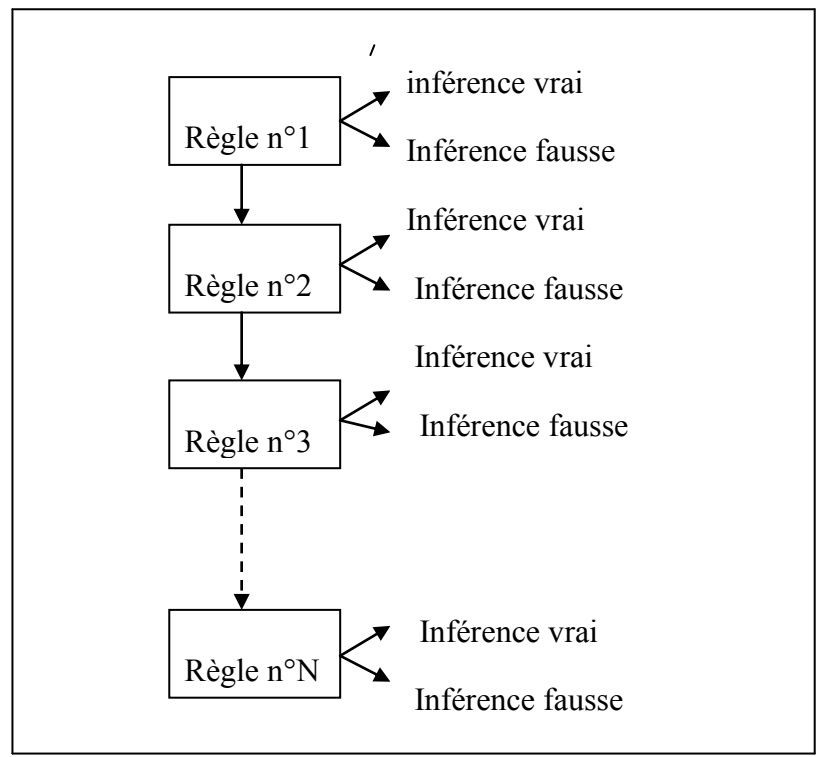

**Figure 4.27: Test des règles d'inférences**

Comme il est montré dans la Figure **4.27** le superviseur exécute les règles du même groupe une par une.

## 4.4) Conclusion

Nous avons présenté dans ce chapitre, notre projet TIMINF. Son architecture informatique se base sur cinq modules principaux. Les deux modules TARSQI et *Link Grammar Parser* constituent la phase de prétraitement indispensable à la phase de test d'inférence, qui nous permet de détecter l'inférence entre évènements et sujets. Le module de balisage qui est inclue dans la deuxième phase est utilisé pour baliser les expressions temporelles non détectées par TARSQI.

Le superviseur est le dernier module de notre système. Celui ci communique avec une base de règle et décide des choix des règles d'inférences temporelles a appliqué. Il a aussi le rôle de tester l'inférence textuelle entre les phrases T et H d'après les données reçues de tous les composants du système. Nous allons présenter dans le chapitre qui suit les différentes étapes de la mise en œuvre du système TIMINF ainsi qu'une étude expérimentale.

- **Chapitre 5 -**

# La mise en œuvre et l'évaluation du système TIMINF

**Chapitre 5**

# La mise en œuvre et l'évaluation du système TIMINF

## 1) Introduction

Dans ce chapitre nous allons expliquer l'installation des différents outils utilisés pour aboutir à notre objectif. Nous donnons aussi un exemple de déroulement de notre système qui résume les principales spécifications de notre projet et montre comment les différents modules peuvent être mis en œuvre dans un système de test d'inférence textuelle intégrant l'aspect temporel dans ses décisions. Nous finissons ce chapitre avec l'évaluation de notre système.

## 2) Environnement et outils utilisés

### 2.1) Python

Pour concevoir notre système nous avons choisi d'utiliser le langage de programmation Python qui a fait ses preuves dans la programmation de nombres applications du TALN.
Python est un langage portable, dynamique, extensible, gratuit, qui permet une approche modulaire et orientée objet de la programmation. Python est développé depuis 1989 par Guido van Rossum et de nombreux contributeurs bénévoles (Swinnen, 2005).

L'interpréteur peut être lancé directement depuis la ligne de commande (dans un « shell » *Linux*, ou bien dans une fenêtre *DOS* sous *Windows*) : il suffit d'y taper la commande **"python"** (en supposant que le logiciel lui-même ait été correctement installé).
Nous utilisons une interface graphique telle que *Windows*. Pour cela nous avons préféré travailler dans un environnement de travail spécialisé tel que *IDLE*.
Avec *IDLE* sous *Windows*, notre environnement de travail ressemblera à celui-ci :
Les trois caractères « supérieur à » constituent le signal d'invité, ou *prompt principal*, lequel indique que Python est prêt à exécuter une commande.

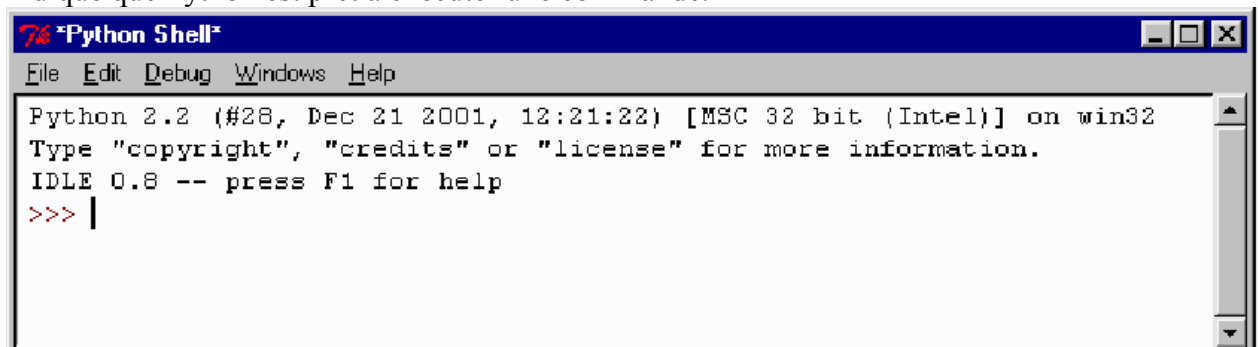

**Figure 5.1 : Shell python**

Pour rédiger nos séquences d'instructions nous avons utilisé l'éditeur incorporé dans une interface de développement telle que *IDLE)*. Il serait parfaitement possible d'utiliser un système de traitement de textes, à la condition d'effectuer la sauvegarde sous un format "texte pur" (sans balises de mise en page). Il est cependant préférable d'utiliser un véritable éditeur ANSI "intelligent" tel que *nedit* ou *IDLE*, muni d'une fonction de coloration syntaxique pour Python, qui aide à éviter les fautes de syntaxe.

La figure ci-dessous illustre l'utilisation de l'éditeur *IDLE)*. Sous (*windows*) :

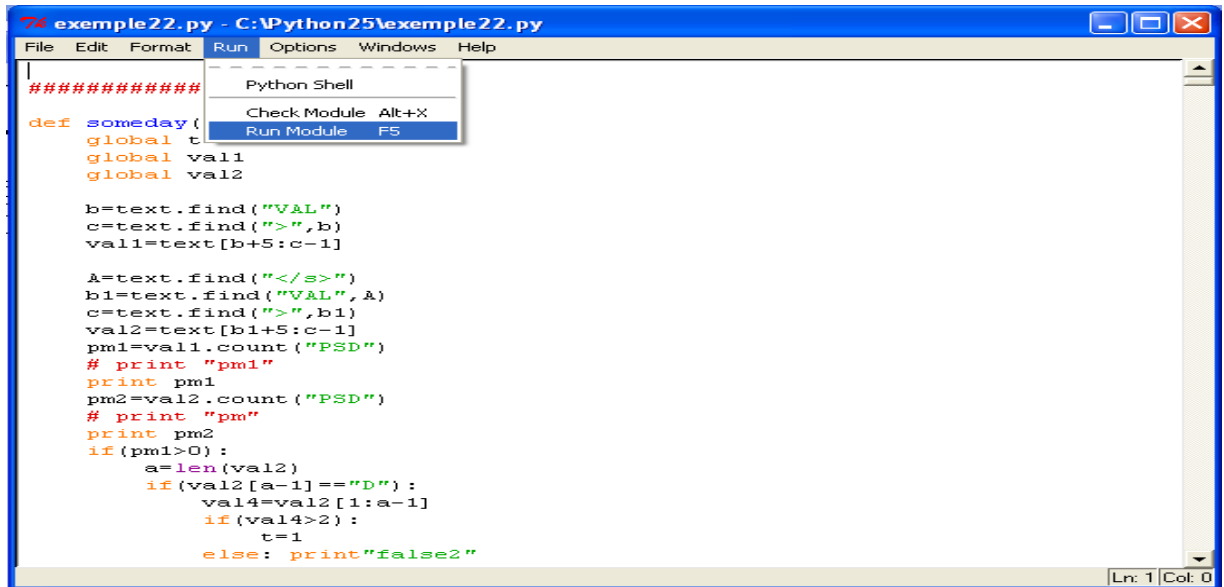

**Figure 5.2 : Comment exécuter un programme**

Par la suite, pour tester l'exécution de notre programme, il nous suffit de lancer l'interpréteur Python en lui fournissant (comme argument) le nom du fichier qui contient le script.

Par exemple, si nous avons placé un script dans un fichier nommé « MonScript », il suffira d'entrer la commande suivante dans une fenêtre de terminal pour que ce script s'exécute :

**python MonScript**

Dans l'*explorateur Windows,* nous pouvons lancer l'exécution de notre script en effectuant un simple clic de souris sur l'icône correspondante ou dans *IDLE*, en lançant l'exécution du script en cours d'édition, directement à l'aide de la combinaison de touches <Ctrl-F5>.

## 2.2) TARSQI

Nous décrivons dans ce qui suit le processus d'installation de TARSQI dans un environnement Linux puisqu'il n'existe pas actuellement une version Windows de TARSQI. Toutefois, le code est écrit pour être multiplateforme. Le groupe TIMEML travaille actuellement sur une version de TARSQI adapté pour Windows qui sera publiée dès que possible.

## 2.2.1) L'installation

La boîte à outils requiert au moins la version 2,3 de Python et la version 5,8 de Perl. La boîte à outils a été testée sur les plates-formes suivantes:
Red Hat Linux 5, avec Python 2.4.3 et Perl 5.8.8
Mac OS X, avec Python 2.3.5 et Perl 5.8.8

Pour installer TARSQI, nous avons d'abord téléchargé et décompresser l'archive dans un répertoire et, taper dans l'invité de commande ce qui suit :

```
% Gunzip-c TTK-1.0.tar.gz | tar xp
```

Cette commande permet de décompresser le contenu dans un répertoire nommé `TTK-1,0`, qui est un répertoire choisi par nous.

La boîte à outils TARSQI est conçue pour fonctionner de façon transparente avec le SGI TreeTagger. Le TreeTagger doit être installé dans `ttk-1.0/code/components/preprocessing/treetagger/`
Ce répertoire doit avoir des sous-répertoires `bin` et `lib`.

## 2.2.2) L'utilisation de la boite à outils TARSQI

Pour exécuter l'outil TARSQI, nous devons ouvrir un terminal, aller au répertoire où se trouve le fichier tarsqi.py et taper :

```
python tarsqi.py <input_type> [drapeaux] <infile> <outfile>
```

`<input_type>`: Il existe deux formats d'entrée de TARSQI : `simple-xml` et `rte3`.

`[drapeaux]`: Avec les drapeaux nous pouvons exécuter un seul ou plusieurs module de TARSQI où l'ordre des modules est important. En voici un exemple:

`[drapeaux]=`          préprocesseur,          GUTIME,          EVITA

L'exemple montre une demande d'exécution des 3 premiers modules de TARSQI.

## 2.2.3) L'utilisation de la boite à outils d'interface graphique

La Boîte à outils d'interface graphique peut être utilisée en tapant :

```
% Pythonw gui.py
```

L'interface graphique a trois avantages sur l'utilisation de la version en ligne de commande:
- Il est plus rapide lors de l'utilisation sur un fichier par fichier, parce que toutes les bibliothèques sont chargées soit au démarrage ou lorsque le premier fichier est traité.
- Il est plus facile à utiliser.
- Il permet à l'utilisateur de taper certains points d'entrée et voir ce qui se passe.

Le principal inconvénient est qu'il n'est pas possible de traiter tous les fichiers dans un répertoire. Voici une capture d'écran:

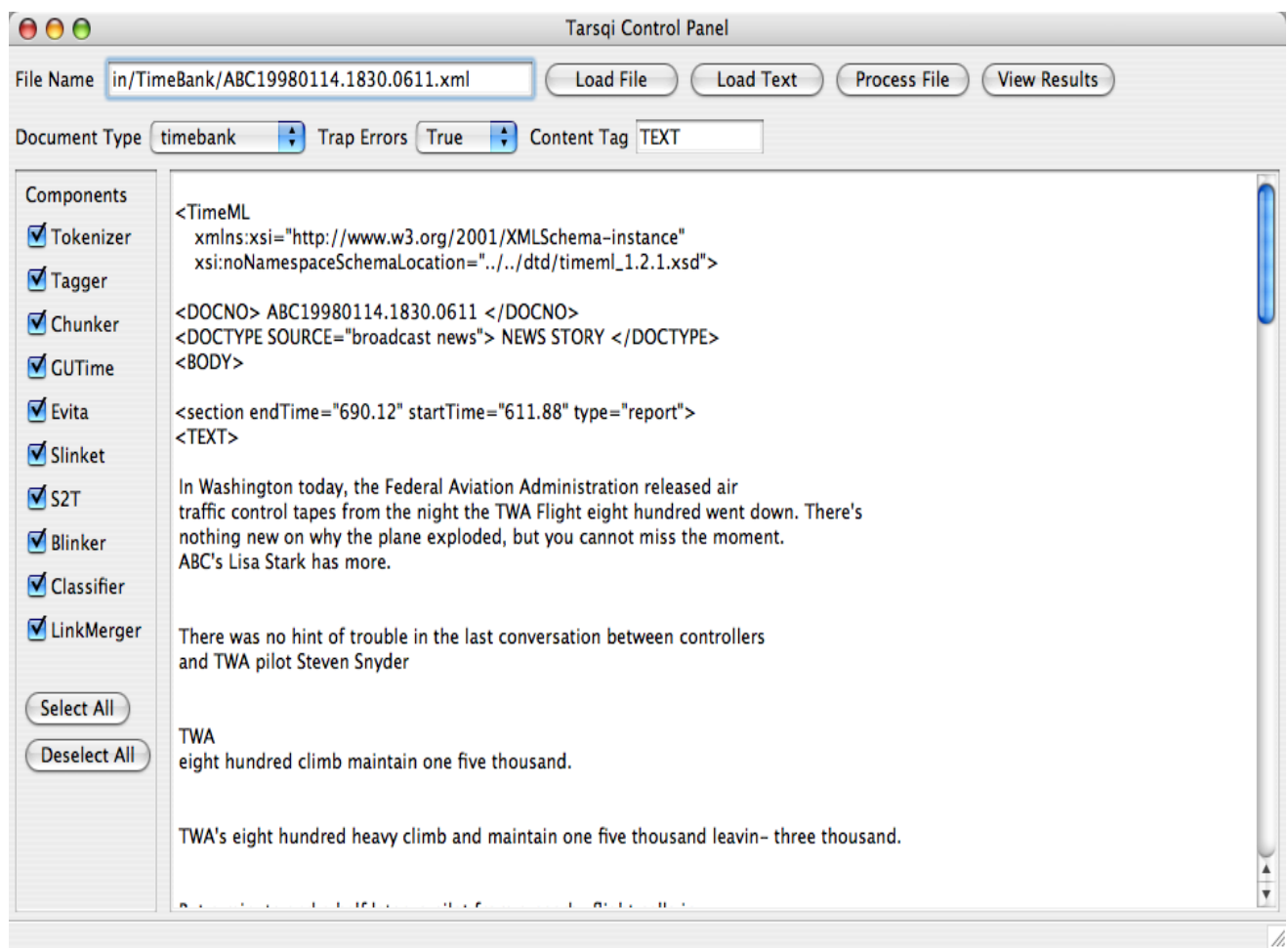

**Figure 5.3 : Capture d'écran de l'interface TARSQI**

Les fonctionnalités peuvent être résumées comme suit:

- Utilisez "Chargez le fichier" pour sélectionner un fichier à traiter.
- Utilisez "Texte de charge" à saisir du texte. Cette opération va créer un fichier dans `le dossier data / en /` répertoire `utilisateur,` qui est ensuite sélectionné comme fichier d'entrée.
- Utilisez « Processus de dossier » pour traiter le fichier d'entrée conformes aux paramètres sélectionnés.

## 2.3) Link Parseur

L'installation de ce module n'est pas difficile, puisque après avoir téléchargé le système de puis le lien suivant (http://www.abisource.org/projects/link-grammar/), nous avons décompressé le contenu dans un répertoire de notre choix. Il suffit d'un click sur l'exécutable contenu dans le répertoire.

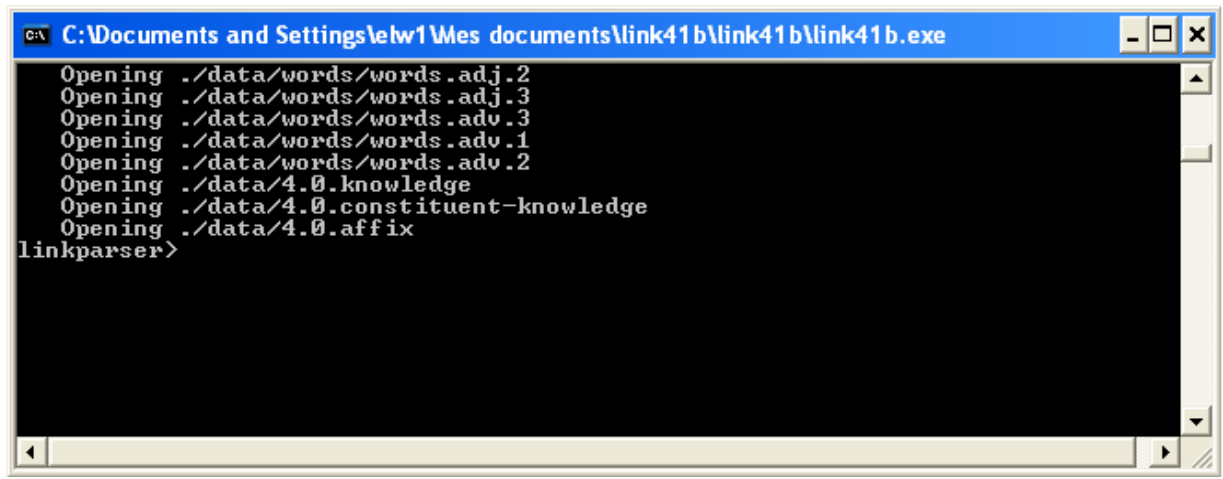

**Figure 5.4 : Capture d'écran de l'interface de link parser**

Il suffit d'écrire le texte que nous voulons analyser et nous aurons l'analyse syntaxique.

# 2.4) PyWordNet

Dans notre module nous avons choisi d'utiliser une version de WordNet qui correspond au choix de notre langage de programmation.
En effet, PyWordNet est une interface Python pour la base de données WordNet qui permet avec des fonctions du langage python de consulter la base de données WordNet.
Exemple :
Si nous tapons l'expression suivante dans l'invité de commande python :

```
>>> N['dog'] dog(n.)
>>> N['dog'].getSenses()
```

Nous interrogeons la base de données sur les différents sens du mot « dog ».
```
{'dog' in {noun: dog, domestic dog, Canis familiaris}
```

## 2.4.1) L'installation
Pour installer **Pywordnet**, il nous a fallu d'abord Télécharger et installer WordNet de 2.0 qui est disponible sur le site http://www.cogsci.princeton.edu/ wn ~ /. Aussi nous avons téléchargé PyWordNet de http://sourceforge.net/projects/pywordnet et décompresser dans le répertoire. Ensuite avec l'invité de commande nous accédons au répertoire contenant les fichiers décompressés et nous tapons `python setup.py`.

Cette commande va permettre concrètement d'installer les deux bibliothèques nécessaires au bon fonctionnement du système. Les deux bibliothèques sont respectivement wordnet.py contient la base de données et wntools.py contient les fonctions qui permettent de consulter la base de données.

## 2.4.2) L'utilisation de PyWordNet dans notre système

Pour savoir si les mots sont antonyme ou synonymie, qui est l'objet du module inférence entre sujet et événement, nous avons utilisé la fonction meet (mot1, mot2, Synonymie) de la bibliothèque PyWordNet qui permet de donner vrai s'il y a une synonymie entre les deux.

La même chose pour l'antonyme mot et meet (mot1, mot2, antonymie) qui permet de donner vrai si 'il y a une antonymie entre les deux mots.

Cette figure représente la fonction qui détecte s'il y a une antonymie entre deux mots programmés en Python :

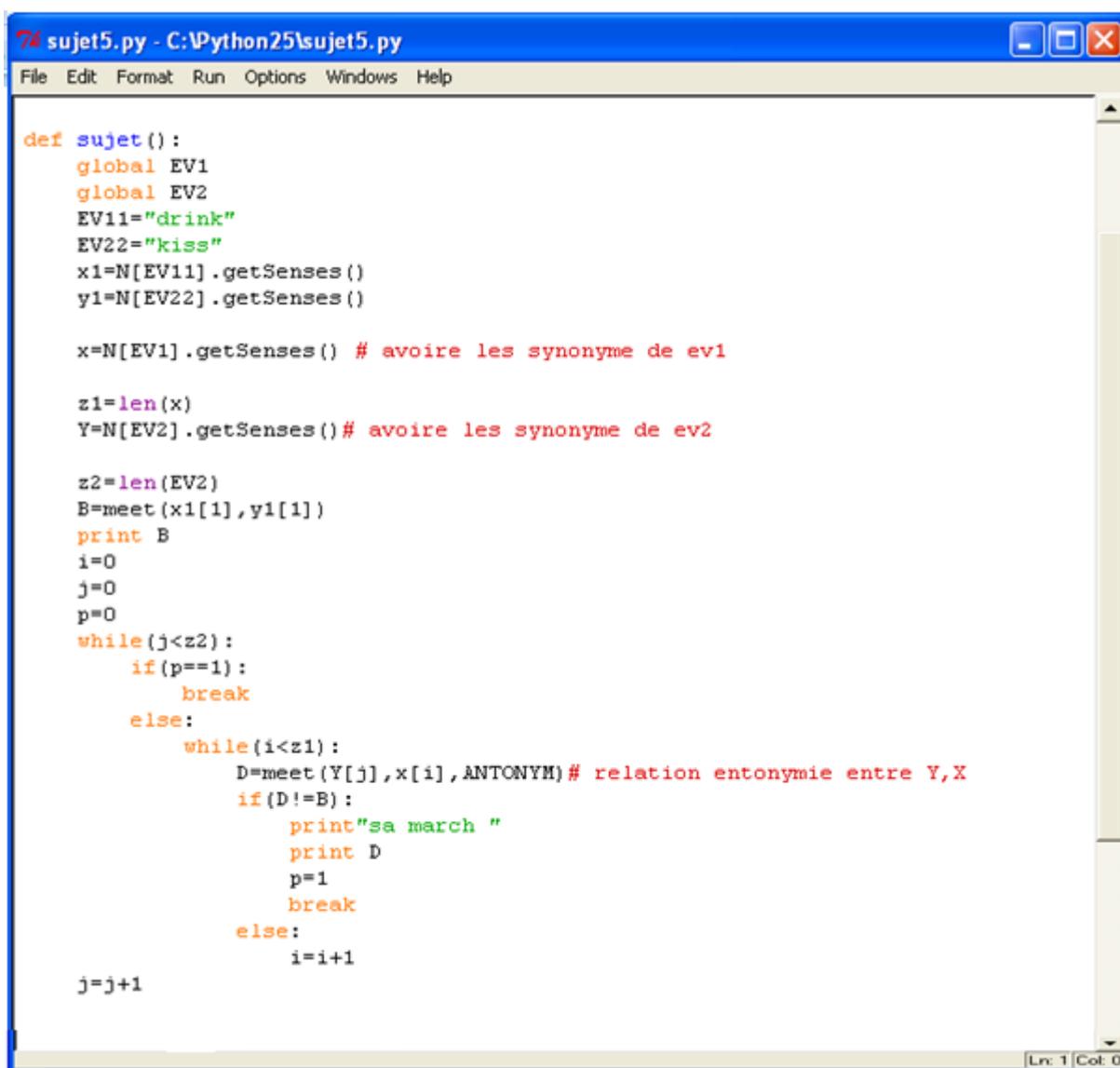

**Figure 5.5 : La fonction d'interfaçage avec WordNet**

Dans ce qui suit nous allons illustrer nos travaux avec le déroulement d'un exemple du corpus sur notre système. Nous allons citer les différentes phases de traitement de la paire de textes.

# 3) Exemple d'exécution du TIMINF sur un exemple du corpus

Nous avons choisi pour l'exemple, la paire numéro 8 du corpus de développement.

Comme il est montré dans l'exemple ci-dessous la première étape est de transformer le texte brut en format simple-xml, pour cela nous avons balisé manuellement les paires du corpus.

```
Entré du système TIMINF

Exemple d'entrée simple-xml.
<DOC>
<DOCID> Simple Test </DOCID>
<TEXT>
the building collapsed at 2 o'clock p.m.
In the afternoon, the building collapsed.
</TEXT>
</DOC>
```

**Figure 4.6 : Entré simple-xml**

## 3.1) TARSQI

Le module TARSQI va permettre de détecter les deux événements de la paire de deux textes (T,H) qui correspondent dans l'exemple au verbe **collapsed** qui est l'évènement des deux segments de textes.

La commande qui permet d'enclencher le module TARSQI avec le format simpel-xml dans un environnement UNIX c'est :

```
python tarsqi.py simple-xml (le nom du fichier contenant les deux
segments de textes)  (le nom du fichier de sortie).
```

Ci-dessous nous montrons la sortie TARSQI correspondant à l'exemple :

**Figure 5.7 : Sortie TARSQI**

## 3.2) L'analyse syntaxique

L'analyse syntaxique se fait en parallèle avec TARSQI et elle va permettre de détecter les sujets des deux segments de textes, qui correspond dans l'exemple à **the building.**

Ci-dessous nous montrons la sortie de LINK Parseur correspondante à l'exemple :

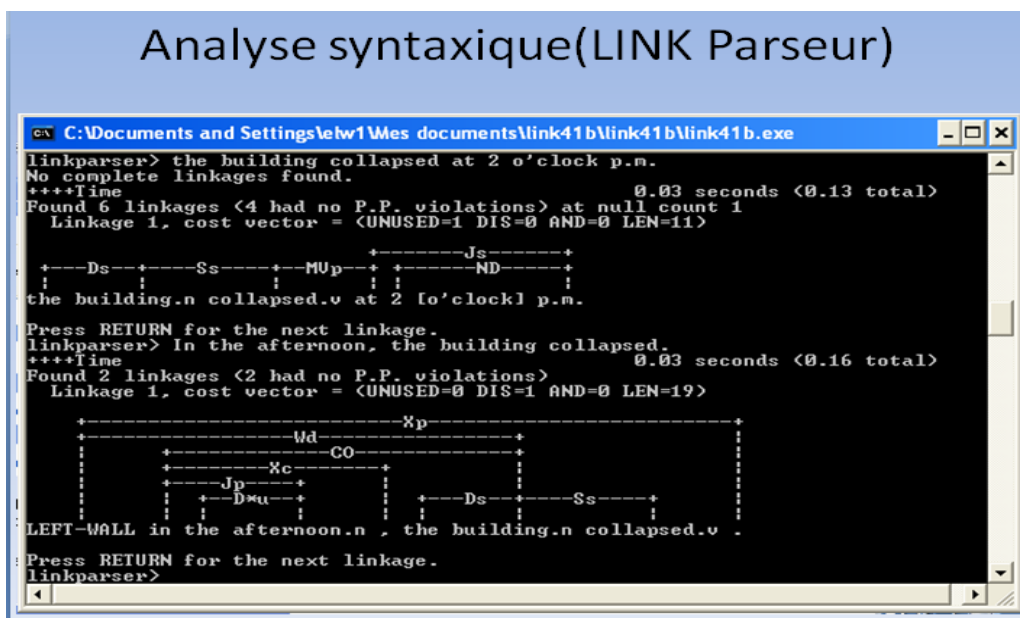

**Figure 5.8: Sortie de l'analyseur syntaxique**

## 3.3) L'inférence entre sujets et événements

Comme il est montré dans l'exemple ci-dessous le module de test d'inférence entre sujets et évènements va permettre de détecter l'équivalence entre les deux sujets et les deux évènements des segments T et H.

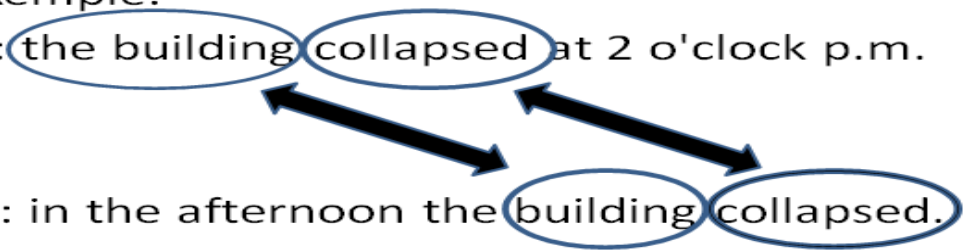

**Figure 5.9 : Inférence entre sujets et évènements**

## 3.4) Le balisages des expressions temporelles non détectées par TARSQI

Comme il est montré dans l'exemple ci-dessous, le balisage des expressions temporelles va permettre de positionner l'expression temporelle dans le temps. Dans l'exemple il détermine que **the afternoon** c'est l'intervalle temporel entre midi et 18 heures.

**Figure 5.10: Balisages des expressions temporelles**

## 3.5) Le superviseur

L'équivalence entre les sujets et les évènements est détectée par la phase d'inférence textuelle et d'ancrage entre les expressions temporelles (2 o'clock et the afternoon) est détecté par l'application de la régle R5 de la base de règles d'inférences qui stipule que si les différentes conditions se réunissent c'est-à-dire :

- détecter une équivalence entre les deux évènements (**e1**, **e2**)

- l'événement **e1** est relié avec une relation TLINK **e1→t2** et l'événement **e1** est relié avec la même relation TLINK à une durée **e2→d.**

- inclusion entre la date **t1** et la durée **d.**

Nous aurons une inférence temporelle et textuelle entre les segments **T** et **H.**

Des deux résultats précédents le superviseur décide qu'il y a une inférence textuelle entre les segments T et H.

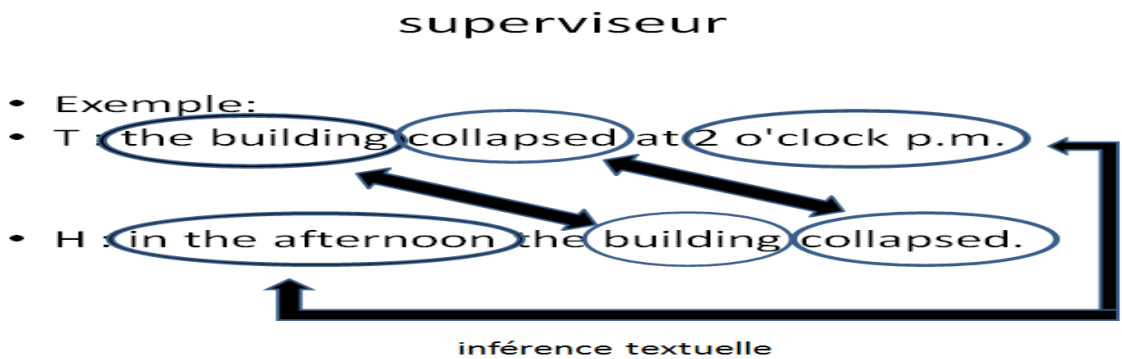

**Figure 5.11 : Test d'inférences**

Cette figure représente les différentes conditions nécessaires à une inférence textuelle.

# 4) L'évaluation de notre système

Notre objectif consiste à améliorer les systèmes d'inférences textuelles. Dans ce cadre, nous avons choisi d'évaluer notre système d'inférence avec le système d'évaluation adopté par le challenge RTE. Pour cela, nous devons évaluer le système par rapport au corpus de développement et aussi par rapport au corpus de test.

Chaque paire du corpus est lancée dans notre système qui donne en sortie s'il y a une inférence textuelle ou pas. Les résultats sont comparés au « **GOLD standard »** que nous avons établi dans notre étape de conception du corpus. Le pourcentage donnant le nombre de fois où il y a similitude entre notre système et le « **gold standard »** donne « **l'accuracy »** du système. l'accuracy est une mesure standard fréquemment utilisée dans les systèmes de traitements du langage naturel.

Dans ce qui suit, nous allons présenter les résultats préliminaires des évaluations des deux corpus.

## 4.1) L'évaluation du système sur le corpus de développement

Nous avons élaboré notre système d'après l'étude des inférences existantes dans le corpus de développement. Ce corpus nous a permis de tester notre système plusieurs fois en effectuant à chaque fois des modifications jusqu'à ce qu'on arrive à concevoir un système qui a donné 100% d'accuracy par rapport à ce corpus.

## 4.2) L'évaluation du système avec le corpus de test

Le corpus de test est constitué de 30 paires de textes, 15 d'entres elles sont évaluées comme contenant une inférence textuelle fausse et les autres sont évalués comme vrai.

Nous avons soumis ce corpus a notre système qui nous a permis de calculer l'accuracy.

Les résultats d'accuracy sont montrés dans le tableau suivant :

| Les systémes | L'accuracy |
|---|---|
| **Système** | 58 % |

**Tableau 2.1 : Le tableau représente l'accuracy du système**

Les résultats de l'évaluation sont encouragent puisque nos résultats sont plus élevés que la moyenne de l'accuracy des systèmes participants au RTE 2 qui sont de 56.6 %.

Dans ce qui suit nous allons étudier les causes de défaillance de notre système.

## 4.4) L'analyse des erreurs causées par le système

D'après notre étude des résultats donnés par notre système nous avons pu élaborer un tableau contenant des statistiques concernant les causes d'échecs de notre système.

| Problème | Pourcentage d'erreur dans le corpus |
|---|---|
| Analyse syntaxique | 38 % |
| TARSQI | 62 % |

**Tableau 3.2 : les causes d'erreurs du système**

Nous remarquons dans ce tableau que les majeures parties des erreurs commises par notre système sont en générale causé par la déficience de l'outil TARSQI.

En effet, TARSQI ne détecte pas plusieurs choses. Par exemple, au niveau de la détection des évènements où nous avons remarqué que TARSQI ne détecte pas les verbes composés comme un événement mais plutôt comme deux évènements indépendants.

Exemple: paire numéro 1 du corpus de test.
T: the First World War spent 7 years.

H: World War I, also known as the First World War, the Great War and the War To End All Wars, was a global military conflict which **took place** primarily in Europe from 1914 to 1918.

Dans l'exemple la détection de l'événement **took place** par TARSQI n'a pas pu se faire car **took place** est un verbe composé.

Aussi les erreurs de notre système viennent de l'analyse syntaxique effectuer en pré traitement par *link parser* où les sujets des verbes ne sont pas détectés.

Exemple: paire numéro 10 du corpus de test.

T: Protracted military S1: conflict between Iran and Iraq. **It** officially began on t1: Sept. 22, 1980, finally, in July, 1988, Iran was forced to accept a United Nations–mandated cease-fire.
H: With more than 100000 Iranian victims of Iraq**'s** chemical weapons during the ten-year war, Iran is one of the countries most severely afflicted by weapons.

Dans l'exemple précédant, la relation entre la date t1 et le sujet S1 n'est pas détecté par TARSQI puisque l'analyse syntaxique n'a pas pu auparavant relier entre **conflict** et **it.**

## 5) Conclusion

Nous avons présenté dans ce chapitre le processus d'installation de nos différents outils nécessaires au bon fonctionnement de notre système. Nous avons également présenté le déroulement de notre système sur un exemple du corpus qui a permis de montrer comment les différents modules étaient mis en œuvre dans notre système.

Enfin, nous avons donné les performances de notre système qui étaient encourageantes et nous avons étudié les différentes failles de notre système. Cela a permis de monter que l'inférence temporelle à un besoin inéluctable aux d'autres modules d'inférences.

# Conclusion générale et perspective

Nous avons présenté, tout au long de ce manuscrit, notre démarche pour la conception d'un système d'inférence textuelle considérant l'inférence temporelle dans sa décision. Pour cela nous avons d'abord exploré l'apport du RTE dans les différentes applications du TAL (RI, QR, EI et RA) et étudié les différentes approches utilisées pour détecter l'inférence (lexical, lexico syntaxique, sémantique et logique). Puis nous avons analysé les approches des différents groupes de recherches qui ont participé aux trois challenges Pascal RTE. Cette étape nous a permis de découvrir les chemins qui n'ont pas encore été étudiés pour détecter l'inférence textuelle.

Ensuite, nous avons exploré la logique temporelle, ses applications dans le traitement du langage nature et les différents types d'inférences temporelles existantes. Cette étude nous a permis de constater qu'il n'y a pas de travail à nos jours liant l'inférence temporelle et la reconnaissance de l'inférence temporelle.

Nous avons élaboré un corpus contenant des paires de segments de textes integrant des relations temporelles et nous avons fait une classification des différents types d'inférences temporelles existants dans le corpus.

La suite logique de ce travail est de déduire des régles d'inférences temporelles et les intégrer à un systéme de reconnaissance d'inférence texuelle.

Une fois le systéme concu, nous avons évalué ses performances avec la méme stategie d'evaluation adoptée dans le challenge pascal RTE. Cette evaluation nous a donné des résultats encourageants.

Enfin, nous avons étudié les différentes failles de notre système. Cela a permis de prévoir plusieurs perspectives de recherches.

## Contribution

Etant donné les objectifs que nous nous sommes fixés pour ce projet, les principales contributions de TIMINF peuvent être résumées comme suit :

L'élaboration d'un corpus à base d'inférence temporelle permettra d'évaluer les recherches futures dans ce Domaine.

L'étude du corpus nous a permis de classifier différents types d'inférence temporelle et de développer différentes règles d'inférences temporelles.

Aussi l'évaluation de notre système a permis de voir concrètement quel est l'apport de l'aspect temporel dans le RTE.

# Perspectives et travaux futurs

Nous envisageons de poursuivre nos recherches futures dans trois directions principales.

Notre système ne permet pas de détecter les entités nommées et de gérer les anaphores. Pour cela, nous envisageons d'introduire un module permettant de détecter et de dater les entités nommées automatiquement. Aussi nous pensons à intégrer un module pour gérer les anaphores et étudier l'impacte de celui-ci sur la performance de notre système.

La seconde direction scientifique est d'évaluer le système prédicat argument comme prétraitement au lieu d'une simple analyse syntaxique.

Enfin, nous envisageons également de développer un système pouvant tester l'inférence textuelle dans des segments de textes plus grandes et utiliser un système qui utilise comme réponse trois sorties possibles (inférence vrai, inférence Fausse ou on ne c'est pas s'il y a une inférence) et nous associons chaque inférence vraie à une application du TALN (QR, RI, IE, PP…).

# Références